\documentclass{article}

\usepackage[utf8]{inputenc}
\usepackage[T1]{fontenc}
\usepackage[english]{babel}
\usepackage{arxiv}
\usepackage{amsmath, amssymb, amsfonts, amsthm}
\usepackage{bm}
\usepackage{microtype}
\usepackage{graphicx}
\usepackage{booktabs}
\usepackage{multirow}
\usepackage{makecell}
\usepackage{threeparttable}
\usepackage{tabularx}
\usepackage{xcolor}
\usepackage[colorlinks=true, allcolors=blue]{hyperref}
\usepackage{url}
\usepackage{doi}
\usepackage{natbib}
\bibpunct{[}{]}{;}{a}{,}{,}
\usepackage{cleveref}

\newtheorem{definition}{Definition}
\newtheorem{proposition}{Proposition}

\theoremstyle{remark}
\newtheorem{remark}{Remark}

\newcommand{\simplex}{\Delta^{k-1}}
\newcommand{\E}{\mathbb{E}}
\newcommand{\Prob}{\mathbb{P}}
\newcommand{\1}{\mathbf{1}}
\DeclareMathOperator*{\argmax}{arg\,max}

\title{Confidence Is Not a Promise: A Reference Audit of Typed Decision Models}

\author{\href{https://orcid.org/0000-0002-1288-7695}{\includegraphics[scale=0.06]{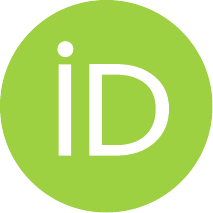}\hspace{1mm}Pedro Carvalho Brom} \\
	L-PRISMA, Department of Mathematics\\
	Federal Institute of Brasília\\
	Brasília, Brazil \\
	\texttt{pedro.brom@ifb.edu.br} \And
\href{https://orcid.org/0000-0003-1826-1850}{\includegraphics[scale=0.06]{orcid.pdf}\hspace{1mm}Li Weigang} \\
	TransLab, Department of Computer Science\\
	University of Brasília\\
	Brasília, Brazil \\
	\texttt{weigang@unb.br}
}

\renewcommand{\shorttitle}{Confidence Is Not a Promise}
\hypersetup{
pdftitle={Confidence Is Not a Promise: A Reference Audit of Typed Decision Models},
pdfsubject={cs.LG},
pdfauthor={Pedro Carvalho Brom, Li Weigang},
pdfkeywords={typed decision models, calibration, selective classification, explanation fidelity, auditing},
}

\usepackage[section]{placeins}

\let\oldsubsection\subsection
\renewcommand{\subsection}{\FloatBarrier\oldsubsection}

\begin{document}
\maketitle

\begin{abstract}
Typed decision models return a probability distribution over options supplied by the caller, and software tends to act on that number as if it were a guarantee. We measure what it licenses, comparing the open-weight Laya, in five checkpoints, with Jev~1.13, available only through an API. A probability licenses a statement about groups of decisions, not about one decision, and a gate that promises an accuracy among the cases it decides alone keeps that promise only with a lower bound checked on data like its traffic. Across 190 exchangeable evaluations, a threshold chosen by point estimate breaks the promise in 147, Learn then Test keeps it in all of them, and a Beta lower-bound rule decides more items on average than Learn then Test in 128 and fewer in none, without its guarantee. Calibrated in English, Jev keeps a promise of 95\% in 8 of 51 locales. Ordinary typing noise breaks a promise calibrated on clean text, and, in exploratory runs, training with noise raises the noisy accuracy with no loss on clean text. In four further domains, a fine-tuned Laya needs from 250 to 8,000 labels to be non-inferior to Jev, and does not get there on LIAR2. Under drift, the preregistered claim for a Bayesian retraining loop is not supported: the loop avoids needless retraining where the change does not degrade the model, at a cost equivalent to recalibration alone, and re-qualifying the threshold in every window keeps the promise more often where it does. On the monthly series of M3, neither decider predicts the direction of the next value better than a constant answer. Confidence separates right from wrong answers with an AUROC from 0.80 to 0.96, but it does not select faithful explanations, and exact coalition games of single decisions show what an explanation can still say through an API that rounds to two decimals. We release the frozen answers with a datasheet, a promise check and the code.
\end{abstract}


%% ============================================================================
\section{Introduction}

Much of the decision logic in software consists of choices among options that change from call to call and arrive written in natural language: which department should handle a message, whether a contract contains a termination clause, how irritated a customer is. A classifier whose output layer has one column per class cannot serve these choices, because the number and the meaning of the classes are fixed in its weights. A large language model can serve them by generating an answer that code then parses, at the cost of one forward pass per generated token and of a probability for each option that has to be reconstructed from token probabilities.

A recent class of models, which its first vendor calls \emph{System One} \citep{typesafe2026docs}, takes a third route. A \emph{typed decision model} receives a state, a question and a set of options described in text, and returns a probability distribution over those options in a single pass. Laya \citep{laya2026}, with open weights, and Jev, available only through an API \citep{typesafe2026docs}, share this interface. The output is directly usable by code: a decision, a probability and a confidence. That convenience raises the question this paper studies. When such a model returns ``billing, 0.97'', what does the number license, and what does it not?

A talk from a security perspective, used here as an example, phrases the question as two statements \citep{ibm2026jev}: confidence is not proof, and a probability is not an explanation. We turn both statements into measurements. The contributions are the following.

\begin{enumerate}
  \item A formal account of the measures that bear on trusting a typed decider, with their calibration, the error types of the gate, its Bayes threshold and its multi-objective formulation, and the number of labels a Beta gate needs before it can promise anything, which is the Bayesian success-run bound (\Cref{sec:measures}).
  \item A paired comparison of five Laya checkpoints and Jev~1.13 on 2,000 MASSIVE utterances per language in Portuguese and English, with fine-tuning and its measured training cost, a test of the vendor's statement about probability one, and a measurement of the relation between a decision's probability and the fidelity of its explanation (\Cref{sec:setup,sec:results}).
  \item A replication, for typed deciders, of the gap between point-estimate and certified thresholds, measured against a plug-in population with Monte Carlo and item-level uncertainty separated, and an evaluation of the Beta lower-bound gate \citep{barkan2023automatic}: its exact relation to Learn then Test, a proposition on when its ascending scan inherits the guarantee of Learn then Test and why it fails otherwise, the coverage that scan buys and the risk it costs, and an ablation over 60 configurations that attributes the effect to the bound rather than to the prior (\Cref{sec:gate,sec:ablation}).
  \item An exact coalition-game analysis of single decisions that states, for each kind of explanation, when it helps a caller decide whether a decision could be overturned by a small change to the input. Words and pairs suffice for short inputs, interaction indices detect redundant evidence even through an API that rounds, no tested method gives calibrated intervals for the effect of a word under that rounding, confidence does not select faithful explanations, a contrastive explanation holds only for the options it was computed on, and an additive head is not a useful route to exactness at a cost of about 30 points of accuracy (\Cref{sec:explain}).
  \item A measurement of natural noise and adversarial typos, showing that a gate calibrated on clean text breaks its promise on noisy text, that a probability of 0.9999 does not protect a decision from a single typo, and that training with noise raises the noisy accuracy with no loss on clean text (\Cref{sec:noise}).
  \item A measurement of how far a promise travels: across 52 locales, where a gate calibrated on Jev in English keeps a promise of 95\% in 8 of 51 locales, and across four further domains, where the labels a fine-tuned Laya needs to be non-inferior to Jev range from 250 to 8,000 and its accuracy on LIAR2 is bounded by what the input carries (\Cref{sec:language,sec:battery}).
  \item A preregistered drift experiment that states when each repair is useful: a Bayesian retraining loop where the change does not degrade the model, re-qualifying the threshold in every window where it does, forgetting faster in per-bin gates, which a dynamic-model gate with its conditions (\Cref{prop:dyngate}) reproduces without adding to it, and, on the monthly series of M3, a model of the series rather than a typed decider (\Cref{sec:drift,sec:drift-pattern,sec:m3}).
  \item A reference package for auditing typed deciders: the frozen answers of both deciders with a datasheet, a check of a selective-accuracy promise across populations, and the code that produces every number of the article.
  \item A validation protocol that separates what a vendor states, what code does, what was measured and what is not published, and maps seven operational controls to measurements (\Cref{sec:validation}).
\end{enumerate}

\Cref{tab:claims} lists each claim of the abstract with its status under the rule of \Cref{sec:compare}, the evidence behind it and the section that reports it, so that a reader can cite a claim with its strength.

\begin{table}[!htbp]
\centering
\caption{The claims of the abstract, their status and their evidence. Confirmatory: fixed in a preregistration. Registered: analysis written before the runs and exploratory. Exploratory: decided after seeing data or not registered. Descriptive: a measurement without a test.}
\label{tab:claims}
\small
\begin{tabularx}{\linewidth}{@{}p{0.36\linewidth}lXl@{}}
\toprule
Claim & Status & Evidence & Where \\
\midrule
A point-estimate threshold breaks the promise in 147 of 190 evaluations, Learn then Test in none & Exploratory & two-sided Clopper--Pearson interval per evaluation, 2,000 to 10,000 draws & \Cref{sec:gate} \\
The ascending Beta rule decides more items on average than Learn then Test in 128 of 190 & Exploratory & means over draws, and on the static front no gain remains after Holm over 24 comparisons & \Cref{sec:gate} \\
Calibrated in English, Jev keeps a promise of 95\% in 8 of 51 locales & Registered & per-locale violation rate on report items & \Cref{sec:language} \\
Typing noise breaks a promise calibrated on clean text, and training with noise helps & Registered & paired accuracy, 95\% bootstrap intervals, margin of 2 points & \Cref{sec:noise} \\
A fine-tuned Laya needs 250 to 8,000 labels to be non-inferior to Jev, and no tested count suffices on LIAR2 & Registered & paired non-inferiority over a grid of label counts & \Cref{sec:battery} \\
The retraining loop does not meet the preregistered criterion & Confirmatory & paired cost against the best comparator in three drifts & \Cref{sec:drift} \\
Re-qualifying the threshold every window keeps the promise most often under degradation & Exploratory & 64 efficiency comparisons, 19 of 27 gains after Holm & \Cref{sec:drift} \\
The gain of the dynamic gate comes from faster forgetting & Registered & paired against recalibration with $\lambda=0.80$ & \Cref{sec:drift-pattern} \\
On M3, neither decider beats a constant answer & Registered & accuracy with 95\% bootstrap intervals over series & \Cref{sec:m3} \\
Confidence detects correct answers, AUROC 0.80 to 0.96 & Descriptive & bootstrap intervals & \Cref{sec:conf} \\
Confidence does not select faithful explanations & Descriptive & Spearman correlation, at most 0.38 in absolute value & \Cref{sec:explain} \\
Exact coalition games state what an explanation can say through rounding & Confirmatory & criteria G1 to G5 fixed before the report items & \Cref{sec:explain} \\
On new items, the bounded rules keep the promise in four of six cases and break it in two, Learn then Test included & Registered & 2,989 and 3,003 external items, Jev called again & \Cref{sec:gate} \\
The advantage of the fine-tuned checkpoints over Jev comes from the learned frequency of the intents & Registered & hard distractors, paired by item and split by the answer to an empty utterance & \Cref{sec:results} \\
\bottomrule
\end{tabularx}
\end{table}

\Cref{sec:related} places these contributions among recent independent evaluations of typed deciders and the literature on selective prediction with guarantees, which already contains the frequentist counterpart of the rule studied here.

%% ============================================================================
\section{Related work}\label{sec:related}

\paragraph{Typed decision models under evaluation.} Since the first typed deciders became available in 2026, an independent literature has asked three questions about them. The first is whether their probabilities are calibrated. Jev~1.13 is found well calibrated and useful for selective prediction on 37 datasets \citep{deusser2026evaluating}, but wording-sensitive and systematically underconfident on a synthetic benchmark built from policy-labelled generators, to the point that under asymmetric costs its probabilities can be worse than its top answer \citep{sharma2026typedbench}. The Laya typed-decisions checkpoint is uniformly underconfident and repaired by one temperature, while a frozen gate misses its error target \citep{nandakishore2026laya}. Calibration measured on familiar tasks does not signal missing knowledge \citep{shankaranarayana2026beyond}, and on Indonesian regional languages Jev's calibration error does not grow while its ranking of its own errors degrades \citep{fadli2026does}. The second question is how they compare with other deciders: Laya and Jev in pairs at eleven decision points of agent harnesses, including a MASSIVE subset of ten items per locale \citep{li2026fast}, eight decision checkpoints against trained classifiers and generative models, with the fraction of intent decisions left to Jev under a guaranteed 5\% risk \citep{rafe2026benchmarking}, and Jev against a frontier LLM on claim verification at a 187-fold cost difference \citep{dorosario2026jev}. The third is whether a decision holds together. Jev and the English Laya checkpoint disagree with themselves when a decision is decomposed \citep{joy2026system}, and calibration with allow, block and defer policies has been evaluated for prompt-injection detection under distribution shift \citep{lan2026detection}. Jeva \citep{li2026jeva} is designed to build the structure of the decision into the architecture, with an explicit null outcome when no option answers the request, invariance to the order of the options, candidate sets of variable size and option representations reused across requests. An audit of the first 28 papers derives a 14-item evaluation checklist \citep{tang2026typed}. Relative to this literature, the present study measures what a promise of accuracy among accepted decisions is worth across the conditions a deployment meets, from the calibration sample to noise, 52 languages, five domains and drift, with one open and one closed decider, fine-tuning and its cost, the explanation of single decisions and a literal test of the vendor's statement about probability one, and it releases the answers it measured.

\paragraph{Selective prediction with a promise.} Abstaining when confidence is low goes back to Chow's rule \citep{chow1970optimum}, and selective classification makes the trade between coverage and risk explicit \citep{elyaniv2010foundations,geifman2017selective}. Selection with guaranteed risk \citep{geifman2017selective} bounds the selective risk by a binomial tail with a union bound over a binary search. The violation rate over calibration draws that a gate controls is the training-conditional, or PAC, guarantee of conformal prediction \citep{vovk2012conditional,park2020pac}. Risk-controlling prediction sets \citep{bates2021distribution} and Learn then Test \citep{angelopoulos2025learn} turn the choice of a threshold into multiple testing with finite-sample guarantees, conformal risk control extends the idea to monotone losses \citep{angelopoulos2024conformal}, selection by conformal $p$-values controls the false discovery rate among selected items \citep{jin2023selection}, and linear expectation constraints control the error probability conditional on selection \citep{wang2026lec}. On the Bayesian side, conformal prediction has been recast as Bayesian quadrature \citep{snell2025conformal}. Pareto testing combines the multi-objective search with these guarantees \citep{laufer2022efficiently}. The cost of ignoring the guarantee has been measured repeatedly: naive thresholds exceed their budget in about half of the settings for binary System One decisions \citep{messoudi2026calibration} and in 49\% to 73\% of trials for signal detectors \citep{zhou2026false}, and \citet{basu2026cross} ablate nine families of bounds, including the Clopper--Pearson bound in the fixed-sequence procedure of Learn then Test, on MASSIVE among other datasets. The Clopper--Pearson lower bound is the equal-tailed bound of an improper $\mathrm{Beta}(0,1)$ posterior \citep{hofer2022comparing,brown2001interval}, which makes the relation between the Bayesian rule studied here and Learn then Test exact rather than approximate (\Cref{sec:ablation}). The contribution here is not a new bound. It is a replication of the gap between point-estimate and certified thresholds for typed deciders, measured against a plug-in population with its uncertainty separated, and an evaluation of the Bayesian lower-bound rule with a uniform prior, described by \citet{barkan2023automatic} for classification thresholds, against Learn then Test and the point estimate: what it buys, what it costs, and where its behaviour comes from. Threshold-based auto-labelling picks the lowest threshold whose estimated error stays below a target with a frequentist bound \citep{vishwakarma2023promises}, and approximate selection with a precision target does the same with a correction for multiplicity \citep{kang2020approximate}. Beta posteriors for model evaluation with few labels are advocated by \citet{bowyer2025position} and \citet{benavoli2017time}, and the optimism of a threshold selected on the same data is an instance of the selection bias documented by \citet{cawley2010overfitting} and \citet{tibshirani2009bias}.

\paragraph{Calibration.} Temperature scaling \citep{guo2017calibration} remains the standard recalibration for neural classifiers and pre-trained transformers \citep{desai2020calibration,minderer2021revisiting}. Top-label calibration is the notion that a thresholded decision needs \citep{gupta2022toplabel}, the binned ECE can underestimate the calibration error of a continuous predictor \citep{kumar2019verified}, and a single temperature is known to fail unevenly across the confidence range and under shift \citep{joy2023sample,ovadia2019can}. The tail failure reported below is an instance of the last point for a typed decider, measured where an extreme cost ratio places the threshold.

\paragraph{Explanation fidelity.} Fidelity is the degree to which an explanation reflects the reasoning of the model \citep{jacovi2020towards}. Deletion and comprehensiveness metrics \citep{samek2017evaluating,petsiuk2018rise,deyoung2020eraser} remove what the explanation points to, and sanity checks and retraining benchmarks warn that such measures can reward explanations unrelated to the model \citep{adebayo2018sanity,hooker2019benchmark}. For decision models, \citet{ahmed2026toward} use leave-one-symptom-out perturbations, and \citet{tropeano2026dont} report that faithfulness and calibration move independently under pruning. We are not aware of a measurement of the relation between a typed decider's probability and the fidelity of its explanation item by item. Attributions that average over coalitions have uncertainty-aware estimators \citep{slack2021,covertlee2021}, interactions between words can be detected hierarchically \citep{chen2020hedge}, and planted shortcuts give a ground truth for faithfulness in text \citep{bastings2022}. Their behaviour when the decider returns rounded probabilities, as a closed API does, has not been measured to our knowledge.

%% ============================================================================
\section{Typed decision models}\label{sec:models}

\subsection{Definition}

Let $s$ be a state (text, a JSON object or a list of messages), $q$ an instruction and $\mathcal{O}=(o_1,\dots,o_k)$ a list of options, each with a textual description. A typed decision model is a map
\begin{equation}
  f_\theta : (s, q, \mathcal{O}) \longmapsto \mathbf{p} \in \simplex, \qquad \simplex = \Big\{\mathbf{p}\in\mathbb{R}^k : p_j \ge 0,\ \textstyle\sum_{j=1}^k p_j = 1\Big\},
  \label{eq:decider}
\end{equation}
where $k$ is chosen by the caller. Since no parameter may depend on $k$, the score of each option is produced by a function shared across options,
\begin{equation}
  z_j = g_\theta(s, q, o_j \mid \mathcal{O}), \qquad p_j = \frac{\exp(z_j/T)}{\sum_{l=1}^{k}\exp(z_l/T)},
  \label{eq:shared}
\end{equation}
with temperature $T>0$. The conditioning on $\mathcal{O}$ allows each option to be scored in the light of the others. Three question types are exposed: \texttt{choice} returns $\mathbf{p}$ and its mode. \texttt{noul} is the case $k=2$ and returns $p_{\text{yes}}$. \texttt{score} takes $L$ ordered levels and returns the expected level $\hat{s}=\sum_{i=0}^{L-1} i\,p_i$, not the most probable one.

\subsection{Two ways of returning a probability}

An autoregressive language model factorises the probability of a token sequence as $\prod_t p_\theta(x_t\mid x_{<t})$. Used as a decider, the probability of an option written with tokens $(w_1,\dots,w_m)$ is $\prod_i p_\theta(w_i\mid\text{prompt},w_{<i})$, and the sum of these over $\mathcal{O}$ is at most one, the remainder being mass on any other text. A typed decider returns a distribution on $\mathcal{O}$ itself, at a cost of one pass per question. Neither pre-training by likelihood of the next token nor preference optimisation \citep{ouyang2022training} requires the stated probability of an answer to match its frequency of being right, and language models are known to be imperfectly calibrated about their own answers \citep{kadavath2022language}. The typed deciders studied here state calibration as a training goal: Jev through a reinforcement-learning stage the vendor calls RLCD \citep{typesafe2026docs}, and Laya through a reward built from strictly proper scoring rules (logarithmic, spherical and ranked probability score) optimised by policy gradient together with cross-entropy, according to its code \citep{laya2026}.

\subsection{The two models}

Laya places one \texttt{[MASK]} token before each option in the input sequence $[\texttt{CLS}]\ \tau(t,q)\ [\texttt{SEP}]\ [\texttt{MASK}]\,\tau(o_1)\cdots[\texttt{MASK}]\,\tau(o_k)\ [\texttt{SEP}]\ \tau(s)\ [\texttt{SEP}]$, where $\tau$ is the tokenizer and $t$ the question type. A bidirectional encoder \citep{devlin2019bert,vaswani2017attention} produces $\mathbf{H}\in\mathbb{R}^{n\times d}$, an embedding of the question type is added, a two-layer Transformer head mixes the positions, and a single MLP maps the vector at each anchor to the score $z_j$. The multilingual checkpoint uses mmBERT-base \citep{marone2025mmbert} with 322 million parameters and a budget of 1,024 tokens, beyond which the state is truncated without warning. The English checkpoint and the \emph{typed-decisions} checkpoint, fine-tuned by the developer of Laya on the developer's own benchmark, use ModernBERT-large \citep{warner2025modernbert}. Jev~1.13 is documented only through its interface, a context of 64 thousand tokens per request, English as primary training language, and its training family \citep{typesafe2026docs}. Its architecture, size, data and reward are not published, and nothing about its internals is assumed here.

%% ============================================================================
\section{Measures and their calibration}\label{sec:measures}

Throughout, a decider answers $n$ items. Item $i$ has gold option $y_i$, probabilities $\mathbf{p}_i$, decision $\hat{y}_i=\argmax_j p_{ij}$ and correctness indicator $a_i=\1[\hat{y}_i=y_i]$.

\subsection{Three confidences}

Three functions of $\mathbf{p}$ are reported as confidence by the systems studied:
\begin{equation}
  c_{\max} = \max_j p_j, \qquad c_H = 1-\frac{H(\mathbf{p})}{\log k},\ \ H(\mathbf{p}) = -\sum_j p_j\log p_j, \qquad c_{\text{J}} = \frac{k\,c_{\max}-1}{k-1}.
\end{equation}
Laya returns $c_{\max}$ as \texttt{answer\_confidence} and $c_H$ as \texttt{confidence}, and Jev returns $c_{\text{J}}$ \citep{typesafe2026docs}.

\begin{proposition}\label{prop:affine}
For fixed $k\ge 2$, $c_{\text{J}}$ is a strictly increasing affine function of $c_{\max}$. Hence for every threshold $\tau$ on $c_{\max}$ there is a threshold $\tau'=(k\tau-1)/(k-1)$ on $c_{\text{J}}$ that accepts exactly the same items. For $k=2$, $c_H$ is a strictly increasing function of $c_{\max}$ and thresholds on the two correspond. For $k\ge3$ no such correspondence holds: there are vectors ordered one way by $c_{\max}$ and the other way by $c_H$.
\end{proposition}

The proof is in \Cref{app:proofs}. Because only $c_{\max}$ is the quantity that temperature scaling adjusts and that calibration refers to, all gates below are thresholds on $c_{\max}$, written $C$.

\subsection{Calibration}

\begin{definition}[Top-label calibration]
A decider is calibrated in its top label when $\Prob(\hat{Y}=Y \mid C=c) = c$ for almost every $c\in[0,1]$.
\end{definition}

Calibration is a property of groups of predictions: it states how often the answers given with confidence $c$ are right, and says nothing about any single answer. Its usual estimator partitions $[0,1]$ into $B$ equal bins $\mathcal{B}_b$ and averages the gap between accuracy and confidence \citep{naeini2015obtaining,guo2017calibration}:
\begin{equation}
  \text{ECE} = \sum_{b=1}^{B}\frac{|\mathcal{B}_b|}{n}\,\Big|\,\overline{a}(\mathcal{B}_b) - \overline{c}(\mathcal{B}_b)\,\Big|, \qquad B=15.
  \label{eq:ece}
\end{equation}
ECE is not a proper score and depends on the binning, and with few items per bin it is biased upwards. We therefore also report two strictly proper scores \citep{gneiting2007proper}, the Brier score and the negative log-likelihood,
\begin{equation}
  \text{BS} = \frac{1}{n}\sum_i\sum_j\big(p_{ij}-\1[y_i=j]\big)^2, \qquad \text{NLL} = -\frac{1}{n}\sum_i \log p_{i,y_i}.
  \label{eq:proper}
\end{equation}
A scoring rule $S(\mathbf{p},y)$, with positive orientation, is strictly proper when $\E_{y\sim\mathbf{q}}S(\mathbf{q},y) > \E_{y\sim\mathbf{q}}S(\mathbf{p},y)$ for all $\mathbf{p}\neq\mathbf{q}$: reporting the true distribution is the unique optimal strategy. This is the sense in which a training objective built from such rules ``rewards honest probabilities''. It is a property of the objective, not a guarantee about a trained model on a new distribution, which is why calibration is measured here rather than assumed.

\paragraph{Temperature scaling.} A single inverse temperature $\beta=1/T$ applied to log-probabilities $\mathbf{z}_i$ is fitted on held-out data by minimising
\begin{equation}
  \mathcal{L}(\beta) = \frac{1}{n}\sum_{i=1}^{n}\Big[-\beta z_{i,y_i} + \log\sum_{j=1}^{k} e^{\beta z_{ij}}\Big]
  \label{eq:temp}
\end{equation}
\citep{guo2017calibration}.

\begin{proposition}\label{prop:convex}
$\mathcal{L}$ is convex on $\mathbb{R}$, with $\mathcal{L}'(\beta)=\frac1n\sum_i(\E_{\mathbf{p}_i(\beta)}[z_i]-z_{i,y_i})$ and $\mathcal{L}''(\beta)=\frac1n\sum_i\mathrm{Var}_{\mathbf{p}_i(\beta)}[z_i]\ge 0$, where $\mathbf{p}_i(\beta)=\mathrm{softmax}(\beta\mathbf{z}_i)$. It is strictly convex when some item has non-constant scores. A minimiser need not exist: if every item has $z_{i,y_i}>\max_{j\neq y_i}z_{ij}$, then $\mathcal{L}'(\beta)<0$ for all $\beta$ and the infimum is approached as $\beta\to\infty$.
\end{proposition}

If some calibration item is wrong, $\mathcal{L}'(\beta)$ tends to a positive limit as $\beta\to\infty$, so a minimiser exists whenever $\mathcal{L}'$ is negative somewhere, and the strict convexity makes it unique. Temperatures are fitted below by Newton's method $\beta\leftarrow\beta-\mathcal{L}'(\beta)/\mathcal{L}''(\beta)$, with the iterate clipped to $[0.05,20]$ as a safeguard. The fitted temperatures reported here lie between 0.40 and 1.77, inside that range. A temperature cannot correct a decider that returns probabilities already rounded, as Jev does: rounded to two decimals, about three quarters of its answers put exactly 1.00 on one option, so $z=\log p$ is $-\infty$ on the others and no $\beta$ changes them. Jev is evaluated as returned.

\subsection{Confidence as a detector of correct answers}

Calibration measures whether the value of confidence is honest. It does not measure whether confidence orders cases well. The second property is measured by the receiver operating characteristic of $C$ as a score for the event $A=1$. Accepting the items with $C\ge\tau$ accepts a fraction $\mathrm{TPR}(\tau)=\Prob(C\ge\tau\mid A=1)$ of correct answers and a fraction $\mathrm{FPR}(\tau)=\Prob(C\ge\tau\mid A=0)$ of wrong ones. The area under this curve equals the Mann--Whitney probability
\begin{equation}
  \text{AUROC} = \Prob(C^+ > C^-) + \tfrac12\Prob(C^+=C^-),
  \label{eq:auroc}
\end{equation}
where $C^+$ and $C^-$ are the confidences of an independently drawn correct and wrong answer \citep{hanley1982meaning}. Confidence would be proof of correctness only if AUROC were 1.

\subsection{Explanation fidelity}

A probability states how much the model believes its decision, and an explanation should state why. We test the simplest form of an explanation, a ranking of input words by importance, by removing what it points to \citep{zeiler2014visualizing,samek2017evaluating,petsiuk2018rise}. For a state with words $w_1,\dots,w_m$ and decided option $\hat{o}$, the occlusion importance of word $w_l$ is $I_l = p_{\hat{o}}(s)-p_{\hat{o}}(s\setminus w_l)$. Removing words cumulatively in decreasing order of $I_l$ gives the deletion curve
\begin{equation}
  D(x) = \min\Big\{1,\ \frac{p_{\hat{o}}(s\setminus\{w_{(1)},\dots,w_{(\lfloor xm\rfloor)}\})}{p_{\hat{o}}(s)}\Big\}, \qquad x\in[0,1],
  \label{eq:deletion}
\end{equation}
and its area $\text{AUD}=\int_0^1 D(x)\,dx$, approximated on an 11-point grid. A faithful explanation makes the decision collapse early, giving a small area. The reference is the same curve with words removed in random order. Because the same perturbation defines and evaluates the importance, and because deleting words produces inputs unlike the training data \citep{hooker2019benchmark}, the comparison with the random order is essential and the measure is read as a lower standard than retraining-based benchmarks.

\subsection{The selective gate}\label{sec:gate-def}

A gate decides alone when $C\ge\tau$ and escalates the rest to a person \citep{elyaniv2010foundations,geifman2017selective}. Its coverage and the accuracy among the cases it accepts are
\begin{equation}
  \phi(\tau) = \Prob(C\ge\tau), \qquad \theta(\tau) = \Prob(A=1\mid C\ge\tau).
  \label{eq:gate}
\end{equation}

\paragraph{The slack of a promise.} At a threshold that keeps the promise, the \emph{slack} is $\theta(\tau)-\alpha$, the accuracy the accepted items can lose before the promise breaks. A change of the input that lowers $\theta(\tau)$ by more than the slack breaks the promise whatever rule chose $\tau$, and \Cref{sec:noise,sec:language,sec:battery} measure such changes. How high $\theta$ can be pushed in the first place is bounded by \Cref{prop:ceiling}.

\paragraph{Two error types.} Among accepted answers, a wrong one is a \emph{type I error}: a wrong action executed without anyone seeing it. Among escalated answers, a correct one is a \emph{type II error}: a person reviews what was already right. Reading the gate as a test of $H_0$: ``the answer is wrong'', accepting is rejecting $H_0$, and a promise $\theta(\tau)\ge\alpha$ bounds the fraction of wrong answers among acceptances, $1-\theta(\tau)$, which is a false discovery rate rather than a type I error rate in the Neyman--Pearson sense. Which error is worse is not a statistical question. It depends on the action the system executes and must be fixed with whoever is accountable for the process.

\paragraph{The F1 of the gate.} With $\mathrm{TP}$ the accepted correct answers, $\mathrm{FP}$ the accepted wrong ones and $\mathrm{FN}$ the escalated correct ones, $F_1 = 2\mathrm{TP}/(2\mathrm{TP}+\mathrm{FP}+\mathrm{FN})$. It ignores the escalated wrong answers and weighs the two error types almost symmetrically, so maximising it pushes the gate to accept nearly everything.

\paragraph{Reviews per avoided error.} Relative to a reference rule, a rule that escalates $\Delta_r$ more items per 1,000 and executes $\Delta_e$ fewer wrong actions pays $\Delta_r/\Delta_e$ reviews per avoided wrong action. It is the cost ratio $C_e/C_r$ at which the two rules cost the same, so the stricter rule is worth it exactly when a wrong action costs more than that many reviews.

\paragraph{The Bayes threshold.} Let a wrong automatic action cost $C_e$ and each escalation cost $C_r$, with $0<C_r<C_e$.

\begin{proposition}\label{prop:bayes}
If the decider is calibrated in its top label, the rule that accepts exactly when $C>\tau^*=1-C_r/C_e$ minimises the expected cost per item among all rules that decide from $C$ alone.
\end{proposition}

The proof is a pointwise comparison of two conditional expected costs (\Cref{app:proofs}). The proposition separates two steps that are often merged: calibration, by a proper score, makes the probability honest, and the metric of the problem then chooses the threshold. Fitting a temperature to maximise F1 or minimise cost would distort the probability that the threshold formula needs.

\paragraph{The conditional promise and the point rule.} A gate promises $\theta(\tau)\ge\alpha$. On a calibration sample, $n(\tau)$ items pass the threshold and $c(\tau)$ of them are correct. The \emph{point rule} scans thresholds from the lowest and keeps the first with $\hat{\theta}(\tau)=c(\tau)/n(\tau)\ge\alpha$, requiring $n(\tau)\ge 20$.

\begin{remark}\label{rem:half}
The estimate $\hat\theta(\tau)$ has standard error $\sqrt{\theta(1-\theta)/n}$, which is 0.44 percentage points for $\theta=0.99$ and $n=500$. If $\theta(\tau)$ decreases continuously with coverage and $\hat\theta$ is approximately normal around it, the point rule stops where the estimate first reaches $\alpha$, that is at a threshold whose true conditional accuracy is close to $\alpha$, and the promise then holds on new data with probability close to one half. This is a heuristic argument, not a theorem: the scan introduces dependence between thresholds. \Cref{sec:gate} measures the rate.
\end{remark}

\paragraph{The Beta lower-bound rule.} The rule evaluated here as an alternative replaces the point estimate by a Bayesian lower bound. The same rule, a Beta posterior on precision with a uniform prior, a credibility requirement at a target precision and the qualifying candidate of largest recall, was described by \citet{barkan2023automatic} for setting the threshold of a classifier, and a frequentist analogue of its minimum-threshold sweep is the threshold estimator of \citet{vishwakarma2023promises}. Applied to one threshold, its criterion is the Bayesian binomial reliability-demonstration test with a uniform prior \citep{wilson2021assurance}. It is also the gate of the open companion material for Laya, which motivates its evaluation here. The rule treats $\theta(\tau)$ as uncertain given the accepted cases. Under a uniform prior and exchangeable outcomes among accepted items,
\begin{equation}
  \theta(\tau)\mid c,n \sim \mathrm{Beta}\big(c+1,\ n-c+1\big),
  \label{eq:beta}
\end{equation}
and the rule keeps the lowest threshold for which the 5\% quantile of this posterior is at least $\alpha$, with the same minimum of 20 accepted cases. In words: a threshold is accepted only if, with posterior credibility 0.95, the accuracy \emph{among the accepted cases} exceeds the target. High thresholds leave few accepted cases, which widens the posterior and makes the rule more demanding, and when no threshold qualifies, the gate decides nothing. The statement is Bayesian. It is not a frequentist guarantee, and it inherits a selection effect when qualification is rare: thresholds then qualify mostly on calibration samples that happened to be favourable. The scan matters: in this rule, every candidate threshold is examined and the lowest one that qualifies is kept, wherever it lies in the scan. We call this the \emph{ascending} scan, and the variant that walks down from the highest threshold and stops at the first one that fails the \emph{fixed-sequence} scan.

\paragraph{Learn then Test and SGR.} Two frequentist rules serve as references. Learn then Test (LTT) \citep{angelopoulos2025learn} tests, for each candidate threshold $\tau_j$ with $n_j$ accepted and $c_j$ correct calibration items, the null hypothesis $\theta(\tau_j)<\alpha$ with the binomial $p$-value $p_j=\Prob\big(\mathrm{Bin}(n_j,\alpha)\ge c_j\big)$, in fixed sequence from the highest threshold, skipping thresholds where even $c_j=n_j$ cannot reject ($\alpha^{n_j}>\delta$) and stopping at the first $p_j>\delta$. With probability at least $1-\delta$ over the calibration sample, the chosen threshold satisfies $\theta(\hat\tau)\ge\alpha$. Selection with guaranteed risk (SGR) \citep{geifman2017selective} searches the $m$ candidates by bisection and accepts a threshold when the Clopper--Pearson upper bound of its error rate at level $\delta/\lceil\log_2 m\rceil$ is below $1-\alpha$.

\begin{remark}[Relation to Learn then Test]\label{rem:ltt}
By the identity between the binomial tail and the regularised incomplete beta function, $p_j\le\delta$ if and only if $\alpha$ is at most the $\delta$-quantile of $\mathrm{Beta}(c_j,\,n_j-c_j+1)$, the Clopper--Pearson lower bound, which is the equal-tailed bound of the improper prior $\mathrm{Beta}(0,1)$ \citep{brown2001interval,hofer2022comparing}. The Beta rule of \cref{eq:beta} replaces this prior by the uniform $\mathrm{Beta}(1,1)$, one pseudo-success, and Learn then Test is recovered exactly by the Beta rule with prior $\mathrm{Beta}(0,1)$, $q=\delta$ and the fixed-sequence scan over testable thresholds. The two rules therefore differ in exactly two choices, the prior and the scan, and \Cref{sec:ablation} measures the price of each.
\end{remark}

The ascending scan has been verified on data, not proved, and the two are different statements. The next result states when the scan inherits a guarantee and why it can fail.

\begin{proposition}[When the ascending scan inherits a guarantee]\label{prop:scan}
Order the testable candidate thresholds from the highest, $\tau_1>\tau_2>\dots>\tau_m$, and let $Q_j=1$ when $\tau_j$ qualifies under the Beta rule. The ascending scan returns $\tau_{J_A}$ with $J_A=\max\{j:Q_j=1\}$, and the fixed-sequence scan returns $\tau_{J_F}$ with $J_F=\max\{j:Q_1=\dots=Q_j=1\}$, each returning no threshold when its set is empty. Let $V_A$ and $V_F$ be the events that the threshold returned by each scan breaks the promise, $\theta(\tau)<\alpha$, and $D=\{J_A\ne J_F\}$ the event that the scans disagree.
\begin{enumerate}
  \item[(i)] The scans agree if and only if the qualifying indices form an initial segment $\{1,\dots,k\}$, $k\ge0$, that is, no threshold qualifies below one that fails.
  \item[(ii)] $V_A\subseteq V_F\cup D$, hence $\Prob(V_A)\le\Prob(V_F)+\Prob(D)$ and $\Prob(V_A\cap D^c)\le\Prob(V_F)$.
  \item[(iii)] With the prior $\mathrm{Beta}(0,1)$ and $q=\delta$, the fixed-sequence scan is Learn then Test, and $\Prob(V_A)\le\delta+\Prob(D)$ under the exchangeability that Learn then Test assumes.
\end{enumerate}
\end{proposition}

The proof is in \Cref{app:proofs}. The prefix form of part (i) is the fixed-sequence procedure of multiple testing \citep{bauer1991multiple,hsu1999stepwise}, used for thresholds by risk-controlling prediction sets and Learn then Test \citep{bates2021distribution,angelopoulos2025learn}, and parts (ii) and (iii) are an elementary decomposition by a union bound, close in form to the argument of \citet{angelopoulos2026conformal}, where the risk of an arbitrary calibration algorithm is bounded by that of a valid reference algorithm plus a gap estimated from the calibration data. The event $D$ is observable on the calibration sample, since both scans can be run on it, so a deployment can tell whether the threshold it uses came from a qualification pattern that the guarantee covers. Its probability is not observable and has to be estimated, for instance by resampling the calibration items, without a finite-sample guarantee. The bound in (ii) concerns the joint event, so agreement of the scans does not by itself bound the conditional probability of a violation by $\delta$. Part (i) also says where the scan can fail: a threshold must fail and a lower one, which accepts every item of the higher one and more, must qualify again. That needs the added items to be right often enough to lift the lower bound back over the target, which happens where the conditional accuracy is flat or not monotone near $\alpha$.

\begin{proposition}[Labels the rule needs]\label{prop:nmin}
Under the uniform prior, a threshold with $n$ accepted calibration items can qualify at target $\alpha$ and credibility level $1-q$ only if $n\ge\lceil \ln q/\ln\alpha\rceil-1$.
\end{proposition}

This is the Bayesian success-run bound. With no failure among $n$ items the posterior is $\mathrm{Beta}(n+1,1)$, a computation that goes back to Price's appendix to the essay of \citet{bayes1763essay} and is reviewed by \citet{luko1997attribute}, and its frequentist version with the $\mathrm{Beta}(0,1)$ prior is the success-run formula of reliability demonstration and the rule of three \citep{hanley1983nothing}. For $q=0.05$ the bound is 58 accepted items at $\alpha=0.95$, 298 at $0.99$ and 597 at $0.995$, which makes the minimum of 20 accepted cases of the rule inactive at these targets. The proof is in \Cref{app:proofs}.

\paragraph{What no rule can promise.} Every rule above chooses a threshold, and so chooses which items to accept. None of them changes what can be known about an item from its input. Let $X$ be the input, $Y$ the correct option, $\eta_y(x)=\Prob(Y=y\mid X=x)$, and $m(x)=\max_y\eta_y(x)$ the largest probability that any option has given the input.

\begin{proposition}[The ceiling of a promise]\label{prop:ceiling}
\begin{enumerate}
  \item[(i)] For every decision function $d$ and every acceptance set $S$ with $\Prob(X\in S)>0$, $\Prob\big(d(X)=Y\mid X\in S\big)\le\E\big[m(X)\mid X\in S\big]$.
  \item[(ii)] If $\Prob(m(X)\ge t)=c>0$, then among all decision functions and acceptance sets with $\Prob(X\in S)=c$, the accuracy among accepted items is largest for $d^*(x)=\argmax_y\eta_y(x)$ and $S^*=\{x:m(x)\ge t\}$, where it equals $\Theta^*(c)=\E[m(X)\mid m(X)\ge t]$, and $\Theta^*$ does not increase with $c$.
  \item[(iii)] A promise $\theta\ge\alpha$ can therefore be kept at coverage $c$ only if $\alpha\le\Theta^*(c)$, and at any positive coverage only if $\alpha$ does not exceed the essential supremum of $m(X)$.
  \item[(iv)] Observing a further variable $W$ does not lower the ceiling, $\Theta^*_{X,W}(c)\ge\Theta^*_X(c)$ at every coverage where both are defined, and merging options into coarser ones does not lower $m$.
\end{enumerate}
\end{proposition}

Part (ii) is Chow's reject rule \citep{chow1970optimum} read as a ceiling: the best selective decider ranks items by $m$. The point rule, the Beta rule, Learn then Test and temperature scaling act on the order of a decider's confidence $C$, so the sets they accept have the form $\{C\ge\tau\}$, and part (i) bounds their accuracy whatever the rule. A decider reaches $\Theta^*$ only where its decisions and its order agree with $d^*$ and $m$, which is what more labels, a better model or a combination of deciders can improve. Only part (iv) raises the ceiling itself, by putting more information in the state or by asking a coarser question. Its first half is the conditional Jensen inequality applied to the convex map $v\mapsto\max_y v_y$ of the posterior, the same argument by which more informative observations never lower the expected value of the best decision \citep{blackwell1953equivalent,degroot1962uncertainty}. The ceiling of a promise is that general fact read for the accuracy among accepted items. The proof is in \Cref{app:proofs}, and the finite case of (i) and the two inequalities behind (iv) are checked in Lean~4 with Mathlib (\texttt{proofs/SelectiveCeiling.lean} in the repository).

\subsection{Multi-objective formulation}\label{sec:multi}

The gate trades coverage against risk, and a deployment usually adds constraints by group. Let $R(\tau)=1-\theta(\tau)$ be the selective risk.

\paragraph{Pareto set and costs.} Because coverage $\phi$ falls with $\tau$ while $R$ need not be monotone, only the lower envelope of the risk--coverage curve is Pareto-optimal. Weighting wrong automatic actions against escalations, $J_w(\tau)=w\,\Prob(A=0,\,C\ge\tau)+(1-w)\,\Prob(C<\tau)$, is the cost model of \Cref{prop:bayes} with $C_e/C_r=w/(1-w)$, so each weight $w>1/2$ selects the Bayes threshold $\tau^*=1-(1-w)/w$ under calibration.

\paragraph{Constrained form.} The promise of the gate is the constrained problem
\begin{equation}
  \max_\tau\ \phi(\tau)\quad\text{subject to}\quad R(\tau)\le 1-\alpha,
  \label{eq:econstraint}
\end{equation}
estimated on a calibration sample. The point and Beta rules are two estimators of its solution, and the Beta rule replaces the constraint by $\Prob(R(\tau)>1-\alpha\mid\text{data})\le 0.05$.

\paragraph{Thresholds by class.} When the cost of a wrong action depends on the option, a vector of thresholds $(\tau_1,\dots,\tau_K)$, one per decided label, solves \cref{eq:econstraint} or minimises $J_w$ separately on the items decided as each label, at the price of fewer calibration items per threshold.

\paragraph{Multicalibration.} A single temperature calibrates on average. For a collection $\mathcal{G}$ of groups (here, MASSIVE scenarios) and confidence bins $\mathcal{B}_b$, multicalibration asks that
\begin{equation}
  \big|\,\E[A - C \mid X\in G,\ C\in\mathcal{B}_b]\,\big| \le \kappa \qquad \text{for all } G\in\mathcal{G}\text{ and all } b
  \label{eq:multical}
\end{equation}
with enough mass in each cell \citep{hebertjohnson2018multicalibration}. A low global ECE can hide groups that violate \cref{eq:multical}, which \Cref{sec:conf} measures.

\subsection{Comparing two methods: superiority, non-inferiority, equivalence and efficiency}\label{sec:compare}

Many of the comparisons below ask whether one method is better than another, and a method that is no worse at a lower cost is already a gain. Saying so requires stating what ``no worse'' means and how it is tested, because the usual reading of a confidence interval that contains zero, as evidence that two methods are alike, is not valid \citep{walker2011understanding}. We use the following definitions throughout. Accuracy comparisons use a margin of 2 points, the margin preregistered for the additive head of \Cref{sec:explain} (criterion G4) and kept for every later comparison of accuracy. It is set by the resolution of the measurement, not by a cost: on 2,000 items an accuracy near 0.95 has a 95\% interval of about one point on each side, so a margin of 2 points is the smallest that a paired comparison can show. A caller for whom a wrong action costs much more than a review should use a smaller margin, which the analysis scripts take as a parameter.

Let $Y_A$ and $Y_B$ be an outcome of methods $A$ and $B$ for which larger is worse, such as a violation rate, the share of windows below the promise or an error rate, observed in pairs over independent units (calibration draws, streams or items), and let $\Delta=\E[Y_A-Y_B]$. A margin $\varepsilon>0$ is the largest difference that does not matter in practice. It is fixed before the data are seen, from the smallest difference of practical interest or from a tolerance already stated \citep{lakens2018equivalence}, as the regulatory practice for non-inferiority trials also requires \citep{fda2016noninferiority}.
\begin{itemize}
  \item \emph{Superiority} of $A$: the hypothesis $H_0:\Delta\ge0$ is rejected in favour of $\Delta<0$. It is claimed when the upper limit of the two-sided 95\% interval of $\Delta$ is below zero.
  \item \emph{Non-inferiority} of $A$ to $B$: $A$ may be worse than $B$, but by less than the margin. The hypothesis $H_0:\Delta\ge\varepsilon$ is rejected in favour of $\Delta<\varepsilon$ by a one-sided test at level 0.025, which is the same as requiring the upper limit of the two-sided 95\% interval to lie below $\varepsilon$.
  \item \emph{Equivalence}: $H_0:|\Delta|\ge\varepsilon$ is rejected in favour of $|\Delta|<\varepsilon$ by two one-sided tests at level 0.05 each \citep{schuirmann1987comparison}, which is the same as requiring the 90\% interval to lie inside $(-\varepsilon,\varepsilon)$.
  \item \emph{No difference detected}: an interval that contains zero and meets none of the three conditions. It is reported as such and is never read as equivalence.
\end{itemize}
A method has an \emph{efficiency gain} over another when its outcome is non-inferior and it uses less of at least one resource (reviews, retrains, GPU time, model passes) with the upper limit of the paired 95\% interval of the difference below zero. The gain is \emph{clean} when no other resource and no error rate is larger with an interval above zero, and a \emph{trade-off} otherwise, where one resource is saved by spending another.

The intervals are paired $t$ intervals over independent units, checked by a bootstrap of the same units where reported. For the gate against the plug-in population, the Monte Carlo error of the calibration draws ignores the sampling of the 2,000 items, so the standard deviation of a difference is taken from 200 bootstrap populations, which includes the item part. Proportions measured on the same items are compared by the exact McNemar test. Comparisons fixed in a preregistration are confirmatory. The others are labelled exploratory, their number is stated, and no correction for multiplicity is applied, so an isolated exploratory result is a lead for further work, not a finding.

%% ============================================================================
\section{Experimental setup}\label{sec:setup}

\paragraph{Data.} MASSIVE~1.1 \citep{fitzgerald2023massive}, test split, in pt-PT and en-US. The test split is shuffled with Python's random generator and seed 13 and its first 2,000 utterances are kept, and each becomes a \texttt{choice} question with five options: the gold intent and four distractors drawn from the other test intents by the same generator, shuffled with the gold intent (\texttt{scripts/itens\_massive.py} rebuilds the items from the MASSIVE release). The two languages share the same intents and options item by item, which allows paired comparisons across languages. The instruction is ``What is the user asking for in \texttt{utterance}?'' and each option is described by its intent name with underscores replaced by spaces. Licence CC~BY~4.0.

\paragraph{Systems.} Jev~1.13, answers collected through the API on 2026-09-26 and frozen. Laya at code commit \texttt{4066d5d} and weights revision \texttt{55cf4c4}: the multilingual, English and typed-decisions checkpoints without fine-tuning, and, in Portuguese only, the multilingual and typed-decisions checkpoints fine-tuned on the 11,274 MASSIVE pt-PT training utterances that remain after removing 240 exact duplicates of test utterances (three epochs, batch 32, AdamW with learning rates $2.5\times10^{-5}$ for the encoder and $10^{-4}$ for the head, recipe in the companion lesson repository). Laya inference ran on CPU in bfloat16 autocast, batches of eight sorted by length. \Cref{tab:checkpoints} fixes the names used in the text and the tables.

\begin{table}[!htbp]
\centering
\caption{Deciders and the names used for them. Base: the pre-trained encoder of each Laya checkpoint. Laya-ft is short for the fine-tuned multilingual checkpoint, and the battery and the time-series front refit it on their own data with the drift recipe.}
\label{tab:checkpoints}
\small
\begin{tabular}{@{}llll@{}}
\toprule
Name in the text & In tables & Base & Fine-tuning \\
\midrule
Jev~1.13 (and 1.14) & Jev & not published & by the vendor \\
Laya multilingual & Laya, released & mmBERT-base & none \\
Laya English & Laya English, root & ModernBERT-large & none \\
Laya typed-decisions & Laya typed-dec. & ModernBERT-large & by the developer of Laya \\
Laya multilingual, fine-tuned & Laya-ft & mmBERT-base & MASSIVE pt-PT, 11,274 items \\
Laya typed-decisions, fine-tuned & Laya typed-dec., ft & ModernBERT-large & MASSIVE pt-PT, 11,274 items \\
\bottomrule
\end{tabular}
\end{table}

\paragraph{Protocol.} Items 0 to 999 calibrate: Laya temperatures are fitted by Newton's method (\Cref{prop:convex}) and gate thresholds are chosen there. Items 1,000 to 1,999 report. The gate is also evaluated over 1,000 random calibration/report splits, refitting the Laya temperature on each calibration half. Intervals: Wilson for accuracy \citep{wilson1927probable}, Clopper--Pearson for the vendor's statement \citep{clopper1934use}, percentile bootstrap with 2,000 resamples for AUROC \citep{efron1979bootstrap}, and the exact McNemar test for paired differences \citep{mcnemar1947note}.

\paragraph{How the promise is measured.} A promise about accuracy among accepted items concerns the population the items come from, which is unknown. The idea is to let the 2,000 items stand in for that population and ask, over many calibration samples drawn from them, how often a rule picks a threshold whose accuracy on the whole population falls short. We measure it against a plug-in population under the exchangeability that every rule already assumes. The 2,000 items of a locale are taken as the population. A calibration sample of $n_{\text{cal}}$ items is drawn from it with replacement, the temperature (for Laya) and the threshold are fitted on the sample, and the true selective accuracy of the chosen threshold is computed on the whole population. A \emph{violation} is a true selective accuracy below the target. Each rule is evaluated on 10,000 independent calibration draws, and the violation rate is reported with its one-sided 95\% Clopper--Pearson upper bound. To separate the uncertainty that comes from the items from Monte Carlo error, the same procedure is repeated on 200 bootstrap resamples of the items, each taken as a population. With $\hat p_b$ the rate on resample $b$ and $M=1{,}000$ draws per resample, the Monte Carlo variance is $w=\frac1B\sum_b\hat p_b(1-\hat p_b)/(M-1)$, and the item variance is $\max\{0,\widehat{\mathrm{Var}}_b(\hat p_b)-w\}$. The same split applies to decisions and to paired differences between rules, computed draw by draw. This checks the rules against the empirical distribution, and it is not a statement about the population behind the sample. Three properties of the construction bear on the numbers. A draw in which no threshold qualifies decides nothing and counts as a kept promise, so a rule can keep its rate low by abstaining, which is why decisions are reported next to every rate. A sample of 1,000 items drawn with replacement from 2,000 contains about 787 distinct items (about 442 for a sample of 500). And the population is discrete, so $\theta(\tau)$ is a step function of the threshold, coarsest for Jev, whose probabilities take few distinct values. The guarantee of Learn then Test assumes candidate thresholds fixed before the calibration data are seen. Here, as in common practice, the candidates are the observed confidences and the Laya temperature is fitted on the same sample, so the guarantee is approximate and the plug-in rates show how close it comes. The tolerance is fixed in advance at the nominal level of the rules: a rule passes when the upper bound of its violation rate is at most 0.05. The upper bound reflects Monte Carlo error only. The item-level standard deviation of the violation rates of the bounded rules reaches 0.016, and the mean over bootstrap populations differs from the plug-in value by up to 0.012, so differences of a point or two between rules are within what another sample of 2,000 items could produce. Because a rule whose true violation rate equals the nominal 5\% would fail this tolerance about half of the time, single cases are also read in three categories, with a two-sided 95\% Clopper--Pearson interval on the violation rate: kept when its upper limit is at most 0.05, broken when its lower limit exceeds 0.05, and undetermined otherwise. \Cref{tab:symbols} separates the levels that share the values 0.05 and 95\%.

\begin{table}[!htbp]
\centering
\caption{Symbols and levels. Several of them take the values 0.05 or 95\%, with different roles.}
\label{tab:symbols}
\small
\begin{tabularx}{\linewidth}{@{}lX@{}}
\toprule
Symbol & Meaning \\
\midrule
$\alpha$ & target of the promise, the accuracy among accepted items (0.95 or 0.99 on MASSIVE) \\
$\theta(\tau)$, $\phi(\tau)$ & accuracy among accepted items and coverage at threshold $\tau$ \\
$q$ & credibility level of the Beta lower bound, its $q$-quantile (0.05 unless stated) \\
$\delta$ & level of Learn then Test, the probability over the calibration sample that its threshold breaks the promise (0.05) \\
tolerance & largest acceptable violation rate of a rule against the plug-in population (0.05) \\
$\varepsilon$ & margin of non-inferiority and equivalence (\Cref{sec:compare}), 2 points for accuracy \\
$\lambda$ & forgetting factor of the Beta counts and discount factor of the dynamic adapter (\Cref{sec:drift}) \\
audit & share of accepted items labelled at random in the drift streams (5\%) \\
intervals & 95\% two-sided unless stated, 90\% for equivalence by two one-sided tests \\
\bottomrule
\end{tabularx}
\end{table}

\paragraph{Reproducibility.} Every number in this paper is produced from the released items, the frozen Jev answers and the Laya probabilities per item by five scripts, all with the seed 20261020: \texttt{metrics.py} (accuracy, calibration, AUROC, costs and the curves of the figures), \texttt{deletion.py} (explanation fidelity), \texttt{population\_gate.py} (the gate against the plug-in population), \texttt{ablation\_gate.py} (the split-half ablation) and \texttt{tables.py}, which writes every table of the paper from their outputs, rounding half up on the stored values. Copies of the vendor documentation pages consulted on 2026-10-20 are archived with the data.

%% ============================================================================
\section{Results}\label{sec:results}

\subsection{Accuracy}

\begin{table}[!htbp]
\centering
\caption{Accuracy on the 1,000 reported utterances, macro-F1 over intents, and the paired difference to Jev~1.13 (exact McNemar). FT: fine-tuned on MASSIVE pt-PT. Best per language in bold.}
\label{tab:accuracy}
\small
\begin{tabular}{llcccc}
\toprule
Lang. & System & Accuracy [95\% CI] & Macro-F1 & $\Delta$ vs Jev, pp [95\% CI] & $p$ \\
\midrule
\multirow{6}{*}{pt} & Jev 1.13 & 0.947 [0.931, 0.959] & 0.945 &  &  \\
 & Laya multilingual & 0.731 [0.703, 0.758] & 0.709 & $-21.6$ [$-24.4$, $-18.8$] & $<10^{-40}$ \\
 & Laya English & 0.736 [0.708, 0.762] & 0.690 & $-21.1$ [$-23.9$, $-18.3$] & $<10^{-40}$ \\
 & Laya typed-decisions & 0.739 [0.711, 0.765] & 0.687 & $-20.8$ [$-23.6$, $-18.0$] & $<10^{-40}$ \\
 & Laya multilingual, FT & \textbf{0.973} [0.961, 0.981] & \textbf{0.953} & $+2.6$ [$+1.1$, $+4.1$] & $8.6\times10^{-4}$ \\
 & Laya typed-decisions, FT & 0.955 [0.940, 0.966] & 0.928 & $+0.8$ [$-0.8$, $+2.4$] & 0.40 \\
\midrule
\multirow{4}{*}{en} & Jev 1.13 & \textbf{0.962} [0.948, 0.972] & \textbf{0.958} &  &  \\
 & Laya English & 0.887 [0.866, 0.905] & 0.855 & $-7.5$ [$-9.5$, $-5.6$] & $3.3\times10^{-14}$ \\
 & Laya typed-decisions & 0.901 [0.881, 0.918] & 0.880 & $-6.1$ [$-8.0$, $-4.3$] & $9.6\times10^{-11}$ \\
 & Laya multilingual & 0.815 [0.790, 0.838] & 0.785 & $-14.7$ [$-17.1$, $-12.3$] & $<10^{-30}$ \\
\bottomrule

\end{tabular}
\end{table}

\begin{figure}[!htbp]
\centering
\includegraphics[width=0.6\linewidth]{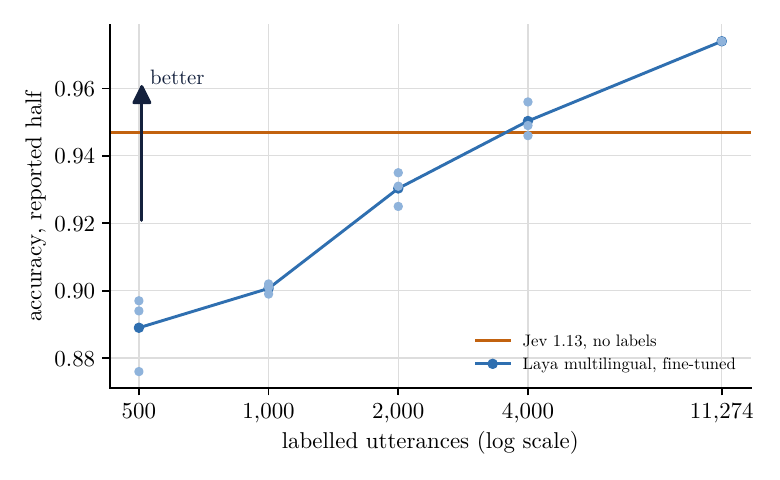}
\caption{Accuracy of the fine-tuned Laya multilingual checkpoint against the number of labelled pt-PT utterances (three seeds per size, mean in the line), with Jev~1.13 without labels as reference.}
\label{fig:labels}
\end{figure}

Without fine-tuning, Jev is more accurate than every Laya checkpoint in both languages (\Cref{tab:accuracy}). In Portuguese the three Laya checkpoints are indistinguishable from each other and about 21 points behind. With labels from the domain the order reverses: the fine-tuned multilingual checkpoint exceeds Jev by 2.6 points ($p<0.001$). Three measurements of the fine-tune appear in the article and differ by recipe. The checkpoint of \Cref{tab:accuracy} is one fine-tune on the full partition with the recipe of the companion lesson (0.973). The label curve of \Cref{fig:labels} comes from the same lesson, with three seeds per size. The paired measurement used for non-inferiority (\Cref{app:cost}) refits with the recipe of the drift experiments and three seeds per size, and gives 0.947 at 4,000 labels and 0.962 at 11,274, 1.5 points above Jev, and the battery of \Cref{sec:battery} uses that recipe on its own grid up to 8,000 labels. The label curve in \Cref{fig:labels} shows that about 4,000 labelled utterances reach Jev's accuracy, in 17 seconds of training on one GPU, a measurement taken from the companion lesson repository with three seeds per size. The choice between the two models is thus less a matter of accuracy than of labels, language, privacy (Jev requires the state to leave the machine) and the need to audit.

\begin{table}[!htbp]
\centering
\caption{What each route costs, MASSIVE pt-PT, reported half. Price per 1,000 decisions from the mean of 384 input tokens over 1,200 calls, at US\$0.042 per million input tokens with output tokens free, as stated by the vendor in October 2026. Inference is CPU time per item for Laya. Jev's latency was not measured, and a local CPU time is not comparable with an API round trip. Training on one RTX 5090 with three seeds. The full-partition times were measured while the GPU was shared with other runs. In bold, the best of each column where an order exists: the highest accuracy, the fewest labels and the state that does not leave the machine.}
\label{tab:cost}
\small
\begin{tabularx}{\linewidth}{@{}>{\raggedright\arraybackslash}Xcccccc@{}}
\toprule
System & Accuracy & Labels & Training & Inference & US\$ per 1,000 & State leaves \\
\midrule
Jev 1.13 & 0.947 & \textbf{0} & none & API, not measured & 0.016 & yes \\
Laya multilingual & 0.731 & \textbf{0} & none & 12.7 ms, CPU & local & \textbf{no} \\
Laya multilingual, 4,000 labels & 0.947 & 4,000 & 17 s, one GPU & as above & local & \textbf{no} \\
Laya multilingual, fine-tuned & \textbf{0.973} & 11,274 & 75 to 105 s, one GPU & 11.5 ms, CPU & local & \textbf{no} \\
\bottomrule

\end{tabularx}
\end{table}

\paragraph{Harder options.} The four distractors of an item are drawn from the other test intents, which rewards a decider that has learned how often each intent occurs. To measure that, the 2,000 items were rebuilt with the distractors drawn first from the gold intent's scenario, 2.4 of the four on average (exploratory, registered before the run). Every decider loses accuracy: Jev 4.2 points in Portuguese and 3.9 in English, the released Laya checkpoints 12 to 20 points and the fine-tuned ones 4.5 and 5.2. With an empty utterance the fine-tuned checkpoints still answer 44\% and 37\% of the hard items correctly, against 20\% by chance, and on the hard items their advantage over Jev lies entirely where the empty utterance already gives the answer: there it is 4.0 and 3.9 points, and where the empty utterance is wrong it is $-0.2$ ($[-2.0,1.8]$) and $-0.6$ ($[-2.3,1.2]$). The edge of the fine-tuned checkpoints over Jev comes from the frequency of the intents they learned, not from a better reading of the utterance.

\paragraph{What each route costs.} Jev needs no labels and costs US\$0.016 per 1,000 decisions at the vendor's price. Laya runs locally in about 12 ms per item on a CPU and is non-inferior to Jev from 4,000 labels on. The cost of each route, the break-even between labels and API spend, and the effect of the processor's instruction set are in \Cref{app:cost}.

\subsection{Confidence as evidence}\label{sec:conf}

\begin{table}[!htbp]
\centering
\caption{Calibration and discrimination on the reported half. ECE with 15 bins as returned and after temperature, Brier score and NLL after temperature, AUROC of confidence as a detector of a correct answer with bootstrap 95\% interval, and ECE per MASSIVE scenario (median and worst over scenarios with at least 30 items). Best per language and column in bold (lowest, except AUROC).}
\label{tab:calibration}
\small
\setlength{\tabcolsep}{4pt}
\begin{tabular}{llccccccc}
\toprule
Lang. & System & $T$ & ECE (as ret.) & ECE ($T$) & Brier & NLL & AUROC [95\% CI] & ECE scen. med./worst \\
\midrule
\multirow{6}{*}{pt} & Jev 1.13 & -- & 0.020 & -- & 0.082 & 0.264 & 0.905 [0.862, 0.943] & 0.039 / 0.077 \\
 & Laya multilingual & 1.77 & 0.110 & 0.045 & 0.370 & 0.764 & 0.825 [0.798, 0.852] & 0.142 / 0.312 \\
 & Laya English & 1.10 & 0.051 & 0.039 & 0.374 & 0.764 & 0.802 [0.773, 0.833] & 0.127 / 0.236 \\
 & Laya typed-decisions & 0.55 & 0.158 & 0.041 & 0.358 & 0.706 & 0.818 [0.790, 0.846] & 0.115 / 0.226 \\
 & Laya multilingual, FT & 1.28 & \textbf{0.018} & \textbf{0.012} & \textbf{0.045} & \textbf{0.095} & 0.928 [0.890, 0.962] & \textbf{0.024} / 0.083 \\
 & Laya typed-dec., FT & 0.83 & 0.021 & 0.013 & 0.064 & 0.130 & \textbf{0.958} [0.936, 0.977] & 0.037 / \textbf{0.057} \\
\midrule
\multirow{4}{*}{en} & Jev 1.13 & -- & \textbf{0.019} & -- & \textbf{0.065} & \textbf{0.256} & \textbf{0.894} [0.833, 0.944] & \textbf{0.029} / \textbf{0.100} \\
 & Laya English & 0.75 & 0.059 & 0.047 & 0.178 & 0.379 & 0.869 [0.839, 0.898] & 0.086 / 0.174 \\
 & Laya typed-decisions & 0.40 & 0.232 & \textbf{0.022} & 0.152 & 0.298 & 0.884 [0.855, 0.910] & 0.063 / 0.180 \\
 & Laya multilingual & 1.64 & 0.073 & 0.048 & 0.271 & 0.550 & 0.849 [0.822, 0.875] & 0.115 / 0.292 \\
\bottomrule

\end{tabular}
\end{table}

\begin{figure}[!htbp]
\centering
\includegraphics[width=\linewidth]{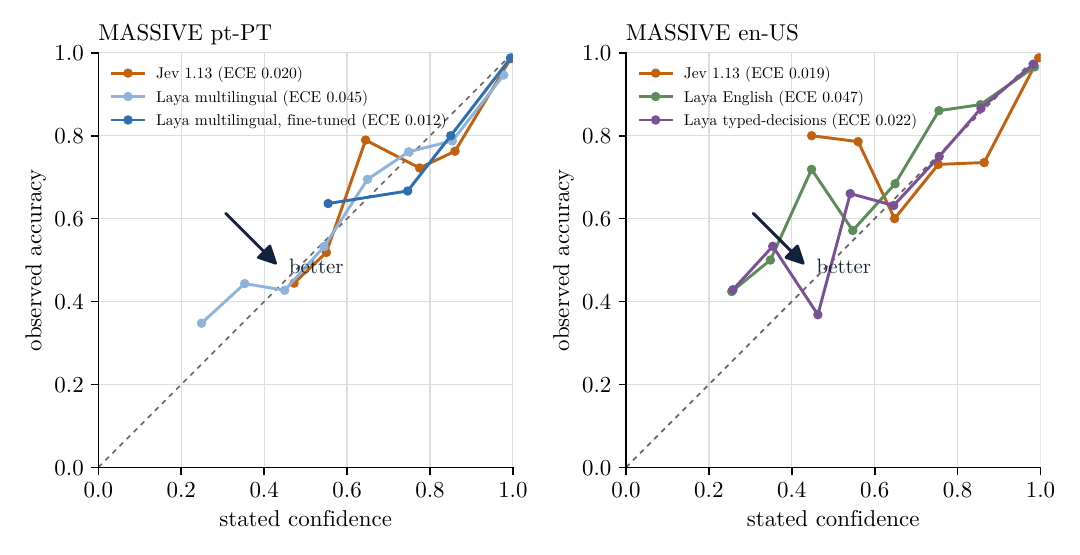}
\includegraphics[width=\linewidth]{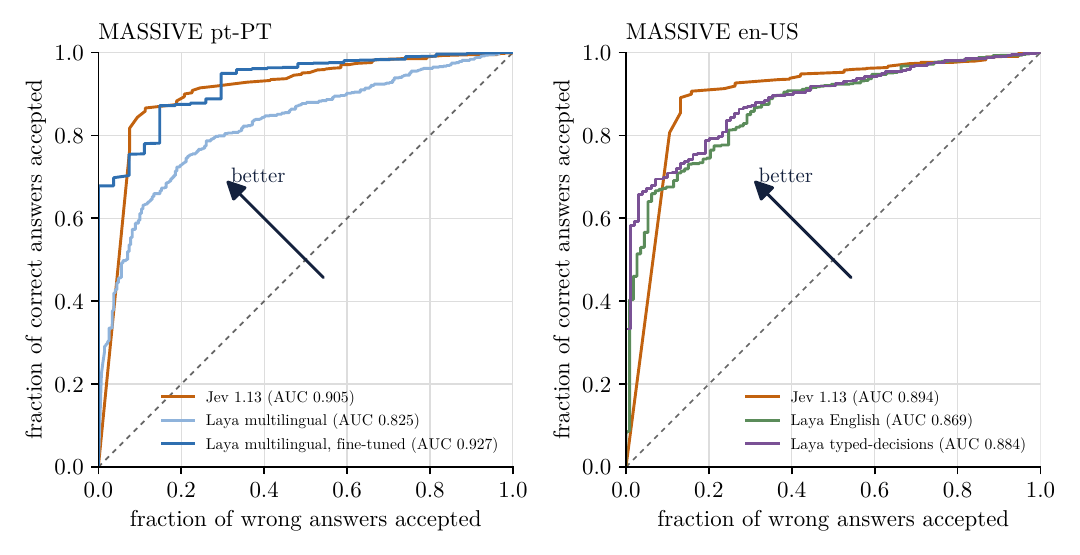}
\caption{Top: reliability diagrams (10 bins with at least 5 items), the diagonal is perfect calibration. Bottom: ROC of confidence as a detector of a correct answer. Arrows point from worse to better.}
\label{fig:conf}
\end{figure}

\paragraph{Ordering, not certifying.} No system reaches an AUROC of one (\Cref{tab:calibration}, \Cref{fig:conf}). Confidence orders cases: a correct answer drawn at random has higher confidence than a wrong one with probability between 0.80 and 0.96. It does not certify any single case. At confidence 0.97 or more, Jev still makes 6 errors among 819 answers in Portuguese and 5 among 862 in English, and the fine-tuned Laya 8 among 896.

\paragraph{A vendor statement, tested.} The vendor writes that outcomes assigned probability 1.0 ``should occur 100\% of the time'' and that these rates ``describe groups of predictions, not a guarantee about any single answer'' \citep{typesafe2026docs}. Jev's probabilities are rounded to two decimals, so 1.00 means $p\ge0.995$. In Portuguese, 724 of the 728 answers at 1.00 were correct, 99.45\% with exact 95\% interval [98.60\%, 99.85\%]. In English, 776 of 780. Read literally, ``100\%'' is rejected, since the interval excludes one. Read with the rounding, it is compatible with calibration at $p\ge0.995$. Each of these errors, alone, looked like certainty.

\paragraph{Calibration repaired, and its limits.} Temperature scaling repairs the global calibration of every Laya checkpoint, including the typed-decisions checkpoint, which is underconfident as returned (temperatures below one) and has the worst ECE as returned, 0.232 in English. Global calibration hides groups: in Portuguese the multilingual checkpoint has ECE 0.045 overall but 0.312 in the \emph{datetime} scenario, and the median scenario ECE of the non-fine-tuned checkpoints is 2.8 to 3.3 times their global ECE in Portuguese and 1.8 to 2.9 times in English. Part of the gap is the upward bias of ECE in small groups, but the order of magnitude is the motivation for \cref{eq:multical}.

\subsection{The gate}\label{sec:gate}

\paragraph{F1 does not know what costs more.} The threshold that maximises the F1 of the gate on the calibration half accepts 97.2\% of the reported items for Jev in Portuguese, with 39 type I and 14 type II errors, and 99.7\% for the fine-tuned Laya, with 25 type I errors and one type II. Wherever a wrong action costs more than a review, this threshold is wrong by construction.

\begin{figure}[!htbp]
\centering
\includegraphics[width=0.62\linewidth]{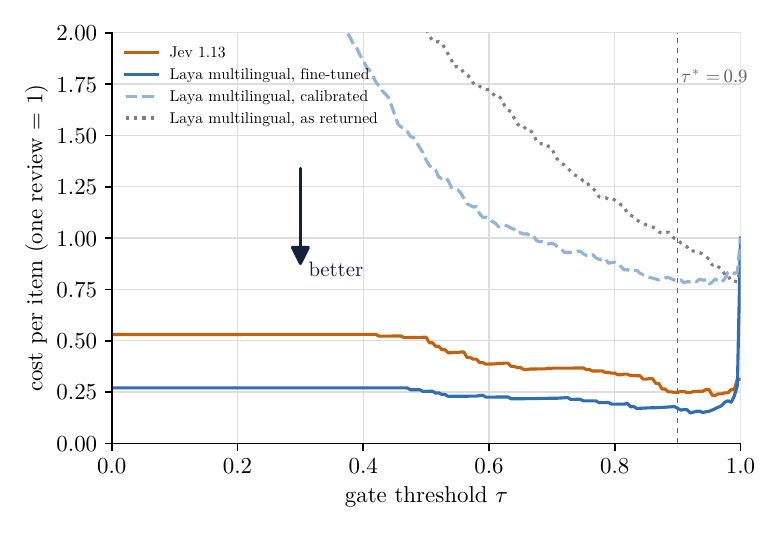}
\caption{Cost per item against the gate threshold for a cost ratio of 10:1 (a wrong automatic action costs ten reviews), MASSIVE pt-PT. The dashed line is the Bayes threshold $\tau^*=0.9$.}
\label{fig:cost}
\end{figure}

\paragraph{The Bayes threshold needs calibration.} \Cref{fig:cost} shows the cost per item against the threshold for a cost ratio of 10:1 ($\tau^*=0.9$). For calibrated deciders the minimum lies near $\tau^*$, and the Bayes threshold, with no fitting, costs 0.248 per item for Jev against 0.234 for the threshold that minimises cost on the calibration half. For the multilingual Laya as returned, the same threshold costs 0.99 per item, against 0.79 after temperature scaling. At 100:1 the gap widens to 3.04 against 1.28. Two qualifications follow from the measurements. With an extreme ratio, the threshold fitted on the calibration half rests on a handful of errors and can be worse on new data than the Bayes threshold: 0.90 against 0.59 for the fine-tuned Laya at 100:1. And a single global temperature need not calibrate the tail where an extreme threshold sits: for Laya English in English at 100:1, the Bayes threshold costs 1.01 after temperature and 0.83 as returned. That difference rests on 5 against 1 accepted errors among 1,000 items, so it is suggestive rather than established, and the split-half evaluation of \Cref{sec:ablation} points the same way: at that ratio, whose threshold is 0.99, the Bayes threshold breaks a promise of 99\% in 29\% of the splits for this checkpoint.

\paragraph{How the promise is measured.} The promise is measured against a plug-in population, as defined in \Cref{sec:setup}.

\begin{table}[!htbp]
\centering
\caption{The gate against the plug-in population, $n_{\text{cal}}=1{,}000$, 10,000 calibration draws. Each cell: violation rate (one-sided 95\% upper bound) / decisions per 1,000 items. Beta rules at $q=0.05$, Learn then Test and SGR at $\delta=0.05$. In bold, per case, the rule with the most decisions among those whose upper bound is at most 0.05.}
\label{tab:gate}
\scriptsize
\setlength{\tabcolsep}{2.5pt}
\begin{tabular}{lllccccc}
\toprule
Target & Lang. & System & Point & Beta, ascending & Beta, fixed seq. & LTT & SGR \\
\midrule
99\% & pt & Jev 1.13 & 0.386 (0.394) / 848 & \textbf{0.014 (0.017) / 400} & 0.014 (0.016) / 338 & 0.014 (0.016) / 338 & 0.002 (0.003) / 156 \\
99\% & pt & Laya multilingual, FT & 0.408 (0.416) / 953 & \textbf{0.020 (0.023) / 879} & 0.019 (0.022) / 851 & 0.019 (0.022) / 851 & 0.002 (0.003) / 776 \\
99\% & pt & Laya typed-dec., FT & 0.504 (0.513) / 911 & \textbf{0.031 (0.034) / 831} & 0.019 (0.022) / 818 & 0.019 (0.021) / 817 & 0.001 (0.001) / 739 \\
99\% & en & Jev 1.13 & 0.487 (0.495) / 880 & \textbf{0.021 (0.024) / 448} & 0.017 (0.019) / 447 & 0.017 (0.019) / 447 & 0.004 (0.005) / 128 \\
99\% & en & Laya typed-decisions & 0.534 (0.542) / 631 & \textbf{0.034 (0.037) / 344} & 0.012 (0.014) / 216 & 0.012 (0.014) / 214 & 0.000 (0.000) / 1 \\
99\% & en & Laya English & 0.533 (0.541) / 503 & \textbf{0.026 (0.028) / 88} & 0.012 (0.013) / 56 & 0.011 (0.013) / 56 & 0.002 (0.003) / 1 \\
\midrule
95\% & pt & Jev 1.13 & 0.416 (0.424) / 993 & \textbf{0.029 (0.032) / 975} & 0.029 (0.032) / 975 & 0.029 (0.032) / 974 & 0.003 (0.005) / 962 \\
95\% & pt & Laya multilingual, FT & \textbf{0.000 (0.000) / 999} & 0.000 (0.000) / 999 & 0.000 (0.000) / 999 & 0.000 (0.000) / 999 & 0.000 (0.000) / 999 \\
95\% & pt & Laya typed-dec., FT & \textbf{0.000 (0.000) / 999} & 0.000 (0.000) / 995 & 0.000 (0.000) / 995 & 0.000 (0.000) / 995 & 0.000 (0.000) / 988 \\
95\% & en & Jev 1.13 & \textbf{0.000 (0.000) / 999} & 0.000 (0.000) / 994 & 0.000 (0.000) / 994 & 0.000 (0.000) / 994 & 0.000 (0.000) / 985 \\
95\% & en & Laya typed-decisions & 0.491 (0.499) / 868 & \textbf{0.034 (0.037) / 827} & 0.032 (0.035) / 812 & 0.032 (0.035) / 811 & 0.003 (0.005) / 793 \\
95\% & en & Laya English & 0.497 (0.505) / 831 & \textbf{0.037 (0.041) / 750} & 0.024 (0.026) / 474 & 0.023 (0.026) / 471 & 0.003 (0.004) / 694 \\
\bottomrule

\end{tabular}
\end{table}

\begin{figure}[!htbp]
\centering
\includegraphics[width=\linewidth]{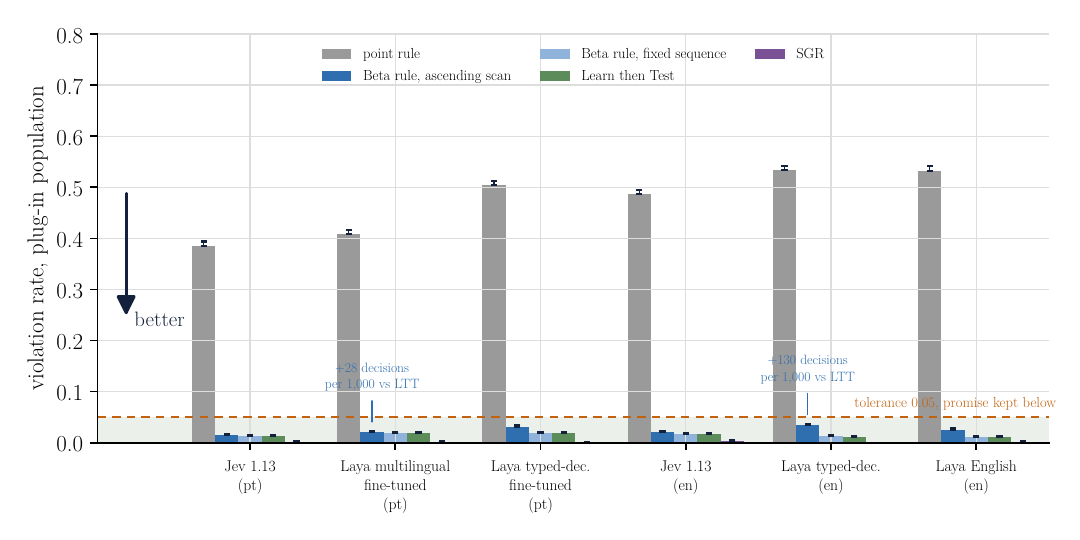}
\caption{Violation rate of the promise of 99\% against the plug-in population, for the five rules of \Cref{tab:gate}, with the one-sided 95\% upper bound as error bar. The dashed line is the tolerance of 0.05, and the shaded band below it is where the promise is kept. The annotations mark the two cases in which the ascending Beta rule decides more items than Learn then Test with a 95\% interval above zero that includes the item-level uncertainty (\Cref{sec:compare}).}
\label{fig:gate}
\end{figure}

\paragraph{Five rules for one promise.} \Cref{tab:gate} and \Cref{fig:gate} compare the point rule, the ascending Beta rule, the same rule with the fixed-sequence scan, Learn then Test and SGR. Wherever the target binds, the point rule violates the promise in 39\% to 53\% of the calibration draws, the coin flip of \Cref{rem:half}. Every rule built on a bound stays within the tolerance in every case, with 1,000 and with 500 calibration items, and the largest upper bound among them is 0.044. Their rates still differ. Learn then Test and the fixed-sequence Beta rule violate in at most 3.2\% of the draws, SGR in at most 0.4\% but decides about one item per 1,000 for Laya typed-decisions and Laya English in English at 99\%, and the ascending Beta rule has the highest rate among the bounded rules, up to 3.7\%. These results replicate, for typed deciders, the contrast between naive and certified thresholds reported for binary System One decisions \citep{messoudi2026calibration} and for signal detectors \citep{zhou2026false}, ablated across nine bound families on MASSIVE among other datasets \citep{basu2026cross}, and seen on the Laya typed-decisions checkpoint itself, where a frozen point-estimate gate missed its error target and, in an exploratory analysis, a rule requiring a Wilson upper bound to clear the target brought the held-out error to 0.101 against a target of 0.10 \citep{nandakishore2026laya}.

\begin{table}[!htbp]
\centering
\caption{Paired differences against Learn then Test on the same calibration draws, plug-in population, $n_{\text{cal}}=1{,}000$: decisions per 1,000 items and violation rate, with the item-level standard deviation from 200 bootstrap populations in parentheses. In bold, coverage differences larger than twice their item-level standard deviation.}
\label{tab:vsltt}
\scriptsize
\setlength{\tabcolsep}{3pt}
\begin{tabular}{lllcccc}
\toprule
 & & & \multicolumn{2}{c}{Beta, ascending $-$ LTT} & \multicolumn{2}{c}{Beta, fixed seq. $-$ LTT} \\
\cmidrule(lr){4-5}\cmidrule(lr){6-7}
Target & Lang. & System & decisions & violation & decisions & violation \\
\midrule
99\% & pt & Jev 1.13 & $+61.8$ (46.5) & $+0.001$ (0.002) & $+0.2$ (0.6) & $+0.000$ (0.000) \\
99\% & pt & Laya multilingual, FT & \textbf{$+28.4$ (11.1)} & $+0.001$ (0.003) & \textbf{$+0.4$ (0.2)} & $+0.000$ (0.000) \\
99\% & pt & Laya typed-dec., FT & $+13.9$ (8.0) & $+0.012$ (0.004) & \textbf{$+0.3$ (0.2)} & $+0.001$ (0.000) \\
99\% & en & Jev 1.13 & $+1.3$ (4.1) & $+0.004$ (0.003) & $+0.4$ (0.8) & $+0.000$ (0.000) \\
99\% & en & Laya typed-decisions & \textbf{$+129.9$ (52.2)} & $+0.022$ (0.006) & $+1.3$ (1.5) & $+0.001$ (0.000) \\
99\% & en & Laya English & $+32.2$ (32.5) & $+0.014$ (0.007) & $+0.0$ (0.2) & $+0.000$ (0.000) \\
\midrule
95\% & pt & Jev 1.13 & $+0.3$ (0.2) & $+0.000$ (0.001) & \textbf{$+0.1$ (0.0)} & $+0.000$ (0.000) \\
95\% & pt & Laya multilingual, FT & $+0.0$ (0.0) & $+0.000$ (0.000) & $+0.0$ (0.0) & $+0.000$ (0.000) \\
95\% & pt & Laya typed-dec., FT & $+0.1$ (0.1) & $+0.000$ (0.000) & $+0.0$ (0.0) & $+0.000$ (0.000) \\
95\% & en & Jev 1.13 & $+0.3$ (0.2) & $+0.000$ (0.000) & $+0.1$ (0.0) & $+0.000$ (0.000) \\
95\% & en & Laya typed-decisions & $+16.9$ (57.7) & $+0.002$ (0.002) & $+1.0$ (1.6) & $+0.000$ (0.000) \\
95\% & en & Laya English & $+279.3$ (171.6) & $+0.014$ (0.007) & $+3.0$ (2.3) & $+0.000$ (0.000) \\
\bottomrule

\end{tabular}
\end{table}

\paragraph{What the ascending scan buys, and at what price.} \Cref{tab:vsltt} isolates the difference to Learn then Test. The fixed-sequence Beta rule is Learn then Test with one pseudo-success (\Cref{rem:ltt}): it decides at most 3.0 more items per 1,000 and violates at most 0.06 points more often. The ascending Beta rule differs by its ascending scan, and that is where both its gain and its cost come from. Read with the definitions of \Cref{sec:compare}, both rules keep the promise in all 24 combinations of system, target and calibration size, so the ascending Beta rule is non-inferior to the tolerance wherever Learn then Test is. In no combination does it decide fewer items with an interval below zero, and it decides more, with a 95\% interval that includes the item-level uncertainty above zero, in 4 of the 24. Two of these are clean gains, 1.1 more items per 1,000 for Jev in Portuguese at 95\% with 500 calibration items and 28 more for the fine-tuned multilingual checkpoint at 99\% (interval $[6,51]$), with no detectable change in violation. Two are trade-offs, 130 more for Laya typed-decisions in English at 99\% ($[27,233]$) at a violation rate higher by 2.2 points ($[0.8,3.7]$), and 279 more for Laya English in English at 95\% with 500 calibration items ($[16,542]$) at 1.4 points more ($[0.1,2.7]$), both still within the tolerance. The 24 combinations include both calibration sizes, and \Cref{tab:vsltt} shows the size of 1,000, where the same difference for Laya English has an item-level standard deviation of 172 and is not detected. In the other 20 combinations no difference in decisions is detected, and the item-level noise reaches 172 decisions per 1,000. With Holm's correction over the 24 combinations, none of the four differences in decisions remains detected (\texttt{scripts/holm.py}). In the static setting, therefore, Learn then Test remains the rule that carries a guarantee, and the ascending Beta rule decides more items on average where the conditional accuracy is not monotone, a gain that the item-level uncertainty of this front does not confirm once the 24 comparisons are corrected for multiplicity.

\begin{figure}[!htbp]
\centering
\includegraphics[width=\linewidth]{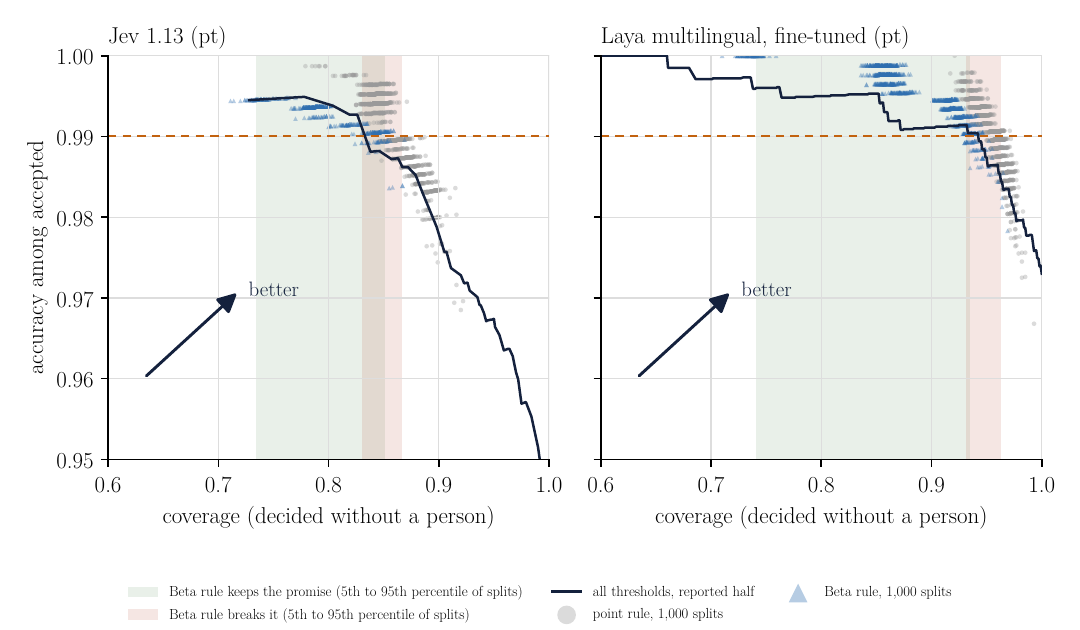}
\caption{What is decided alone against what is right, MASSIVE pt-PT, target 99\%, split-half evaluation. The line is every threshold on the reported half, and each point is one of 1,000 splits. Splits in which a rule qualifies no threshold decide nothing and are not shown: 608 of 1,000 for the Beta rule with Jev, none for the fine-tuned Laya. Shaded bands: the coverage (5th to 95th percentile) of the Beta-rule splits that kept the promise on the reported half (green) and of those that broke it (red). For the fine-tuned Laya the rule keeps the promise between coverages of about 0.74 and 0.93 and breaks it above, where the curve flattens near the target, the region that \Cref{prop:scan} and the measured disagreement of the scans point to.}
\label{fig:accov}
\end{figure}

\paragraph{The exchange rate.} Against the point rule, a bounded rule pays for each avoided wrong action with extra escalations. With the split-half evaluation of \Cref{app:tables}, the ascending Beta rule spends about 14 extra reviews per avoided wrong action for the fine-tuned Laya, so it pays off as soon as a wrong action costs more than 14 reviews, as in an assistant that books events. For Jev, rounded probabilities create a step at 1.00 that leaves the rule without a qualifying threshold in 52\% to 61\% of the splits, and the rate rises to 78 to 92 reviews per avoided error, which pays off only for critical actions. \Cref{fig:accov} places the splits on the accuracy--coverage plane: the point rule straddles the target line, the Beta rule moves left and up.

\paragraph{What lower coverage buys.} The curves of \Cref{fig:accov} also show where escalating more is worth it. On the reported half in Portuguese, lowering the coverage from 100\% to 90\% removes 32 of Jev's 53 errors at about 3 reviews per avoided error, and 19 of the 27 errors of the fine-tuned Laya at about 5. From 90\% to 80\% the price rises to about 6 and 25 reviews per avoided error, and below 80\% almost nothing is left to avoid. Jev keeps 4 errors among the 728 answers it returns with probability 1.00, and the fine-tuned Laya reaches no error at 65\% coverage, 650 items whose Beta lower bound is still only 0.995. The marginal price is the cost ratio at which each coverage is optimal, which is how \Cref{prop:bayes} reads on data. A target of 99.5\% asks for a coverage where each further error costs tens of reviews.

\subsection{Why the bound works, and where it fails}\label{sec:ablation}

To attribute the behaviour of the rules, we score 60 configurations on 1,000 random calibration/report splits of the 2,000 items (\Cref{app:tables}). They cover:
\begin{itemize}
  \item the point rule,
  \item the Beta rule with $q\in\{0.01,0.025,0.05,0.10,0.20\}$, the uniform, Jeffreys and $\mathrm{Beta}(0,1)$ priors and three scans, plus one variant with a minimum of 100 accepted items,
  \item Learn then Test and SGR with $\delta\in\{0.01,0.05,0.10,0.20\}$,
  \item the Bayes threshold for cost ratios of 10, 100 and 1,000,
  \item per-class thresholds with the point and the Beta rule.
\end{itemize}
The full grid runs with 1,000 calibration items at targets of 95\% and 99\%, with temperature scaling on and off for Laya. Jev is always used as returned. The core rules also run with 500 and 250 calibration items, at 99.5\%, and on the non-fine-tuned checkpoints. The split-half rates include report-side noise. They serve to attribute effects, while the plug-in population above serves to judge the promise.

\paragraph{Why it works.} Almost the whole effect comes from replacing the point estimate by a bound. How much the prior weighs depends on which prior. Replacing the uniform prior of the rule by $\mathrm{Beta}(0,1)$ changes the split-half rate by at most 0.007 at $q=0.05$, and against the plug-in population, with the definitions of \Cref{sec:compare}, it keeps the promise in all 24 combinations of system, target and calibration size and changes decisions by at most 2.6 per 1,000. The Jeffreys prior, more liberal by half a pseudo-observation of each kind, decides more items in 16 combinations, by up to 245 per 1,000, but violates more often in 8 and breaks the promise in 5 of the 24, two with Jev, one with Laya typed-decisions and two with Laya English, all in English except one Jev case, and four of them at 99\%, where the uniform prior keeps it. The prior is therefore not neutral. Priors at least as conservative as the uniform one keep the ascending scan within the tolerance at these sample sizes, and the Jeffreys prior does not. The credibility level $q$ acts as $\delta$ does, with violation rates rising with $q$. The bound also makes the rule abstain when the evidence is insufficient. With 250 calibration items at 99\%, no bounded rule qualifies a threshold for any system, in agreement with \Cref{prop:nmin}, while the point rule still promises and breaks the promise in 50\% to 65\% of the splits. Every rule that counts accepted successes uses only the order of the confidences, so the point rule, the Beta rule, Learn then Test and SGR are nearly invariant to temperature scaling. For the fine-tuned multilingual checkpoint at 99\%, the Beta rule gives 0.073 broken and 883 decisions after temperature scaling and 0.069 and 882 without. The invariance is not exact, because a temperature can reorder items whose probability vectors differ in shape: for Laya typed-decisions in English at 99\%, the Beta rule decides 277 items per 1,000 after temperature scaling and 180 without.

\paragraph{Where it fails.} The ascending scan of the rule keeps the lowest threshold that qualifies anywhere, and so exploits stretches where the conditional accuracy is not monotone. This is the source of both its additional coverage and its failures in \Cref{tab:gate,tab:vsltt}. When qualification is rare, the chosen threshold is also optimistic. For Laya English in English at 99\%, the calibration accuracy among accepted items exceeds the report accuracy by 1.2 points on average, against $-0.1$ to $0.4$ points in the other eleven cases. Per-class thresholds either break the promise often or decide almost nothing. With 59 predicted intents, each class has about 17 calibration items, far below the minimum of \Cref{prop:nmin}: at 99\%, the per-class point rule breaks the promise in 34\% to 85\% of the splits, and the per-class Beta rule decides no item.

\paragraph{When the ascending scan breaks the promise.} \Cref{prop:scan} locates every excess violation of the ascending scan in the draws where the two scans disagree, and the plug-in population measures it. Over 10,000 draws per combination, the fixed-sequence Beta rule violates within 0.0011 of Learn then Test in all 24 combinations, the scans disagree in 0\% to 60\% of the draws, and the inclusion of part (ii) holds in every combination. Disagreement alone is not the danger. Where the ascending threshold covers about as much as the fixed-sequence one, the disagreeing draws violate no more often than the others, as for Laya English in English at 95\%, which disagrees in 59\% of the draws. The danger is a jump in coverage. In the three combinations where the disagreeing draws cover more than 10 points above the rest, 0.91 to 0.95 against 0.74 to 0.80, the promise breaks in 17\% to 98\% of them (the fine-tuned checkpoints at 99\% with 500 calibration items, and Jev in English at 99\%). The ascending scan therefore fails where the accuracy--coverage curve of \Cref{fig:accov} flattens near the target at high coverage, and a deployment can detect the risk by comparing the two scans on its own calibration sample.

\paragraph{Across every evaluation.} The tolerance used so far counts a promise as kept only when the upper bound of its violation rate is at most 0.05, so a rule whose true rate equals the nominal 5\% would fail about half of the time. A fairer reading sorts each evaluation into three categories with a two-sided 95\% Clopper--Pearson interval: kept when the upper bound is at most 0.05, broken when the lower bound exceeds 0.05, and undetermined otherwise. \Cref{tab:beta-audit} applies it to the 190 evaluations of this article in which calibration and evaluation items are exchangeable: the static gate of this section, the same design for the three released checkpoints in Portuguese, the battery of \Cref{sec:battery} with MASSIVE pt-PT as its reference, and Laya calibrated on each MASSIVE locale. Every case left undetermined with 2,000 draws was run again with 10,000. Against the point rule, the usual choice when no guarantee is sought, the ascending Beta rule always pays: the point rule breaks the promise in 147 of the 190 cases with mean violation rates of 29\% to 48\%, and the Beta rule brings them to 1.5\% to 2.7\% at a cost of 51 to 165 decisions per 1,000. Against Learn then Test, which keeps the promise in all 190 cases, the Beta rule decides more items on average in 128 cases and fewer in none, differences between means over calibration draws that do not include the uncertainty of the items, 16 more per 1,000 for Jev, 34 for the fine-tuned Laya and 73 for Laya as released, almost twice as many. Its price is a violation rate near the nominal level without a guarantee: the mean rises from 1.0\% to 1.7\% for Learn then Test to 1.5\% to 2.7\%, six cases are undetermined, and one is broken, the fine-tuned checkpoint calibrated in Armenian at 5.5\% (95\% interval $[5.1,6.0]$). In every case where the Beta rule exceeds the nominal level, the two scans disagree in 64\% to 79\% of the draws, as \Cref{prop:scan} anticipates. The Beta rule is therefore the better trade when more automatic decisions matter more than a formal guarantee, and it gains most with the weakest decider. Where the promise needs a guarantee, Learn then Test remains the rule.

\begin{table}[!htbp]
\centering
\caption{The three rules over the 190 exchangeable evaluations of the article, by kind of decider, each case with its tightest measurement (10,000 draws where 2,000 left it undetermined). Kept: upper bound of the violation rate at most 0.05. Undetermined and broken: the Beta cases with the interval straddling 0.05 or above it. Mean violation rate and mean decisions per 1,000 over the cases. In bold, per row, the rule that keeps the promise most often, the lowest mean violation and, between the two bounded rules, the one that decides more.}
\label{tab:beta-audit}
\small
\setlength{\tabcolsep}{3pt}
\begin{tabular}{@{}lcccccccccccc@{}}
\toprule
 & & \multicolumn{3}{c}{Kept} & \multicolumn{2}{c}{Beta} & \multicolumn{3}{c}{Mean violation} & \multicolumn{3}{c}{Decisions per 1,000} \\
\cmidrule(lr){3-5}\cmidrule(lr){6-7}\cmidrule(lr){8-10}\cmidrule(lr){11-13}
Decider & Cases & point & Beta & LTT & undet. & broken & point & Beta & LTT & point & Beta & LTT \\
\midrule
Jev 1.13 & 18 & 5 & \textbf{18} & \textbf{18} & 0 & 0 & 0.290 & 0.015 & \textbf{0.012} & 659 & \textbf{494} & 478 \\
Laya as released & 82 & 5 & 78 & \textbf{82} & 4 & 0 & 0.477 & 0.027 & \textbf{0.010} & 277 & \textbf{162} & 89 \\
Laya fine-tuned & 90 & 32 & 87 & \textbf{90} & 2 & 1 & 0.321 & 0.024 & \textbf{0.017} & 737 & \textbf{686} & 652 \\
\bottomrule

\end{tabular}
\end{table}

\paragraph{Checked on new items.} The plug-in population treats the 2,000 items as the population. To check that reading, the gate was calibrated on the same kind of draws and read both on the sample and on 2,989 Portuguese and 3,003 English MASSIVE items outside it, the dev partition and the test rows not sampled, with Jev called again so that its answers carry the variation of the API (exploratory, registered before the run). In four of the six cases the new items confirm the plug-in reading: the point rule breaks the promise in 35\% to 51\% of the draws where it broke on the sample, and the bounded rules stay at or below 3.8\%. In the other two the bounded rules break on the new items, Learn then Test included. The fine-tuned multilingual checkpoint, whose accuracy falls from 0.975 on the sample to 0.965 on the new items, breaks a promise of 99\% in 30\% of the draws under Learn then Test and 41\% under the Beta rule. Laya English, with a similar accuracy on both (0.886 and 0.893) but a confidence that ranks the new items differently, breaks a promise of 95\% in 6.1\% and 10.9\% of the draws. A guarantee for exchangeable items does not survive a shift as small as the one from the sampled test rows to the dev partition, and the lower accuracy of the fine-tuned checkpoint on new items is consistent with the near duplicates noted in the limitations.

%% ============================================================================
\section{When the input is noisy}\label{sec:noise}

Every measurement so far uses MASSIVE as written, clean and in lower case. Real messages arrive with typing errors, words out of order and, in Portuguese, without accents. Clean training data of this kind is, in effect, combed, and it does not prepare a decider for the wild. The question for a caller is whether such noise changes the decisions and whether the promise of a gate calibrated on clean text survives it. Five operators were applied to the report items, deterministic given a seed: keyboard typos (a QWERTY neighbour, a deleted, a doubled or two transposed letters) hitting 5\%, 10\% or 20\% of the words, adjacent words swapped, lower case without punctuation, accents removed, and a mixed profile that combines the last three with typos. Laya, as released and fine-tuned, and Jev~1.13, through the API with fresh answers for clean and noisy text alike (19,000 calls, US\$0.31), were measured on every condition. All of this is exploratory, registered before the runs.

\begin{table}[!htbp]
\centering
\caption{Natural noise, report items. Accuracy on clean text and under typos at 10\% of the words and the mixed profile at 10\% and 20\%, and the violation rate of the promise of 95\% by the ascending Beta rule, with thresholds chosen on 1,000 calibration samples of clean items and applied to the clean and to the mixed-10\% report items. In bold, per language, the highest accuracy and the lowest violation rate of each column.}
\label{tab:noise}
\small
\begin{tabularx}{\linewidth}{@{}>{\raggedright\arraybackslash}Xcccccccc@{}}
\toprule
 & & \multicolumn{4}{c}{Accuracy} & \multicolumn{2}{c}{Violation at 95\%} \\
\cmidrule(lr){3-6}\cmidrule(lr){7-8}
System & Lang. & clean & typo 10\% & mixed 10\% & mixed 20\% & clean & mixed 10\% \\
\midrule
Jev 1.13 & pt & 0.950 & 0.939 & \textbf{0.939} & \textbf{0.934} & \textbf{0.00} & \textbf{0.15} \\
Laya multilingual & pt & 0.732 & 0.686 & 0.646 & 0.636 & 0.18 & 0.98 \\
Laya multilingual, fine-tuned & pt & \textbf{0.973} & \textbf{0.944} & \textbf{0.939} & 0.930 & \textbf{0.00} & 1.00 \\
\midrule
Jev 1.13 & en & 0.961 & 0.947 & \textbf{0.948} & \textbf{0.943} & \textbf{0.00} & 0.75 \\
Laya multilingual & en & 0.816 & 0.752 & 0.744 & 0.726 & 0.01 & \textbf{0.18} \\
Laya multilingual, fine-tuned & en & \textbf{0.973} & \textbf{0.949} & 0.939 & 0.916 & \textbf{0.00} & 0.97 \\
\bottomrule

\end{tabularx}
\end{table}

\paragraph{Accuracy degrades, unevenly.} Paired against clean text with an item bootstrap, typos and the mixed profile cost Jev 0.7 to 2.2 points of accuracy on average, and removing accents costs it nothing detectable (\Cref{tab:noise}). The fine-tuned Laya loses 2.4 to 5.7 points and the released checkpoint 4.6 to 9.6. Swapping adjacent words and dropping punctuation change almost nothing for any of them. Fine-tuning on clean labels made Laya 2.3 points more accurate than Jev on clean Portuguese text and 1.2 points on clean English text, and noise shrinks the difference to between $-2.7$ and $+0.5$ points, Jev being ahead by 2.7 under the heaviest mixed noise in English.

\paragraph{The promise does not travel.} A threshold chosen on clean items keeps the promise of 95\% on clean report items for every system, and breaks it on noisy ones. For the fine-tuned Laya the promise breaks in 11 of the 12 typo and mixed conditions of the two languages, with violation rates of 95\% to 100\%, and for Jev in 9 of the 12, with rates of 13\% to 81\%, by the Beta rule and by Learn then Test alike. A loss of a few points of accuracy is enough, because it falls on the accepted items and pushes their accuracy below the target. This is the failure of exchangeability that the drift streams of \Cref{sec:drift} also meet, produced here by nothing more than ordinary typing.

\begin{figure}[!htbp]
\centering
\includegraphics[width=\linewidth]{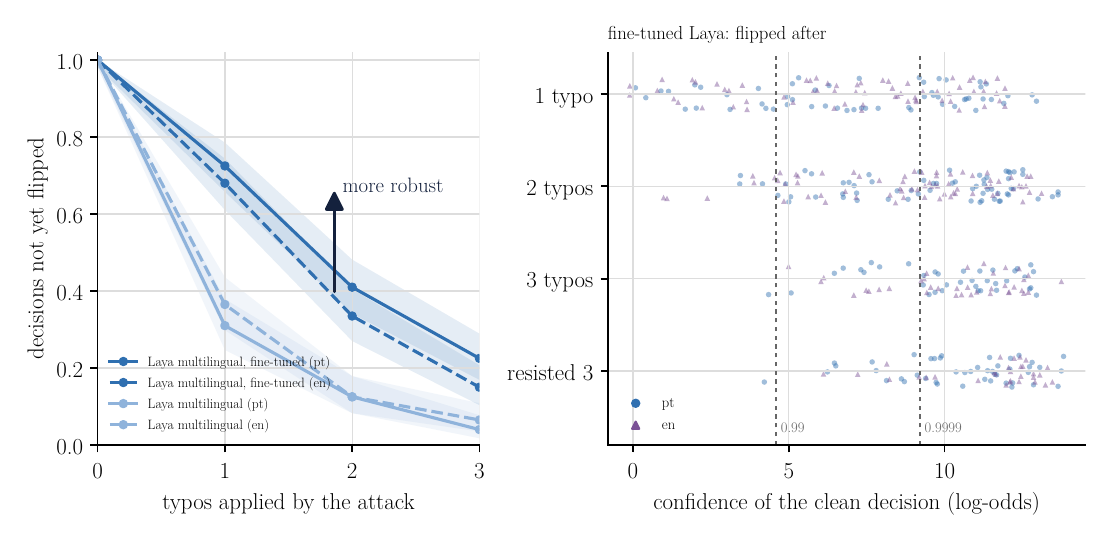}
\caption{Greedy typo attack on Laya, 200 report items per language. Left: share of decisions not yet changed after each number of typos, with 95\% intervals, for the released and the fine-tuned multilingual checkpoint. Fine-tuning delays the fall but does not prevent it. Right: for the fine-tuned checkpoint, the confidence of each clean decision, in log-odds, grouped by the number of typos that changed it. The groups overlap across the whole range, so a confidence of 0.99 or 0.9999 (dashed lines) does not tell which decisions will survive. Jev was not attacked, only measured under random noise.}
\label{fig:adversarial}
\end{figure}

\paragraph{A few typos flip confident decisions.} A greedy search that applies, one at a time, the typo that most lowers the probability of the clean decision changes the decision of the fine-tuned Laya on 27.5\% of 200 Portuguese items with a single typo (95\% interval $[21,34]$), 59\% with two and 77.5\% with three, and of the released checkpoint on 69\%, 88\% and 96\%. English is similar. The confidence gives no warning. Among the decisions flipped, the median probability of the clean decision was 0.9999, and in Portuguese 88\% of them had a probability of at least 0.99. \Cref{fig:adversarial} shows both sides, and \Cref{tab:adversarial} shows what such an attack looks like.

\begin{table}[!htbp]
\centering
\caption{Typos that change a decision of the fine-tuned Laya. The examples are not chosen by hand: in each language, the item of median clean confidence among those changed by one typo, and in Portuguese also among those changed by two, as fixed in the implementation notes before they were seen. The gold intent is the clean decision in all three.}
\label{tab:adversarial}
\small
\begin{tabularx}{\linewidth}{@{}cXXll@{}}
\toprule
Lang. & Clean utterance & Attacked utterance & Clean decision & Decision after the typos \\
\midrule
pt & existe um tosquiador só de gatos na cidade & exiate um tosquiador só de gatos na cidade & \texttt{general\_quirky} (0.9993) & \texttt{play\_game} (0.92) \\
en & open temperature & open temprature & \texttt{weather\_query} (0.9993) & \texttt{play\_music} (0.33) \\
pt & eliminar uma lista & eiminar uma lisha & \texttt{lists\_remove} (0.99997) & \texttt{iot\_hue\_lightup} (0.90) \\
\bottomrule

\end{tabularx}
\end{table}

\paragraph{Training with noise is a gain of method.} The fine-tuning recipe of the original head, with half of the training items passed through a random operator, keeps the clean accuracy and raises the noisy one. Against the same recipe without noise, paired over items and averaged over three seeds, the clean accuracy is non-inferior with the 2-point margin of \Cref{sec:compare} (Portuguese $+0.3$ with interval $[-0.1,0.8]$, English $+0.3$ with interval $[-0.2,0.8]$), and the accuracy under the mixed profile at 10\%, the condition fixed in advance, is higher ($+1.6$, $[0.7,2.4]$, and $+0.8$, $[0.1,1.5]$). By the protocol of \Cref{sec:compare} this is superiority on noisy text with non-inferiority on clean text, at no cost in labels, not an efficiency gain, which would require a resource saved. It recovers about half of what noise takes in Portuguese, from 0.930 to 0.946 under the mixed profile against 0.964 on clean text.

For a caller, the reading is direct. A gate is calibrated for the inputs it was calibrated on, and typing noise is enough to break it, so the calibration items must look like the traffic, noise included. Fine-tuning for accuracy should add noise to its training data, which costs nothing on clean text. And a probability of 0.9999 does not certify that a decision would survive a single typo.

%% ============================================================================
\section{When the language changes}\label{sec:language}

Noise is one change of the input that a deployment meets. Language is another, and a decider offered as multilingual invites the question of whether a promise calibrated in one language holds in the others. MASSIVE 1.1 gives the same utterances in 52 locales \citep{fitzgerald2023massive}. Each of the 2,000 items of the article was mapped to its MASSIVE identifier, and the utterance with the same identifier replaced the Portuguese one in every locale, with the options and the gold intent kept, so the items are parallel across locales. Jev~1.13 answered the 1,000 report items of the 50 locales not measured before (50,000 calls, US\$0.83), and Laya multilingual answered all items as released and as fine-tuned in Portuguese. The analysis was registered before the runs and is exploratory. A deployment that starts in English calibrates in English. The threshold is therefore chosen on the 1,000 calibration items of en-US, with the temperature fitted on the same draw for Laya, and applied to the report items of each locale, which share no item with the calibration items. The check is the plug-in procedure of \Cref{sec:gate} with 2,000 draws, run by the tool released with the article (\texttt{promise\_check}), whose test reproduces the plug-in results of \Cref{sec:gate} within four standard errors.

\begin{figure}[!htbp]
\centering
\includegraphics[width=\linewidth]{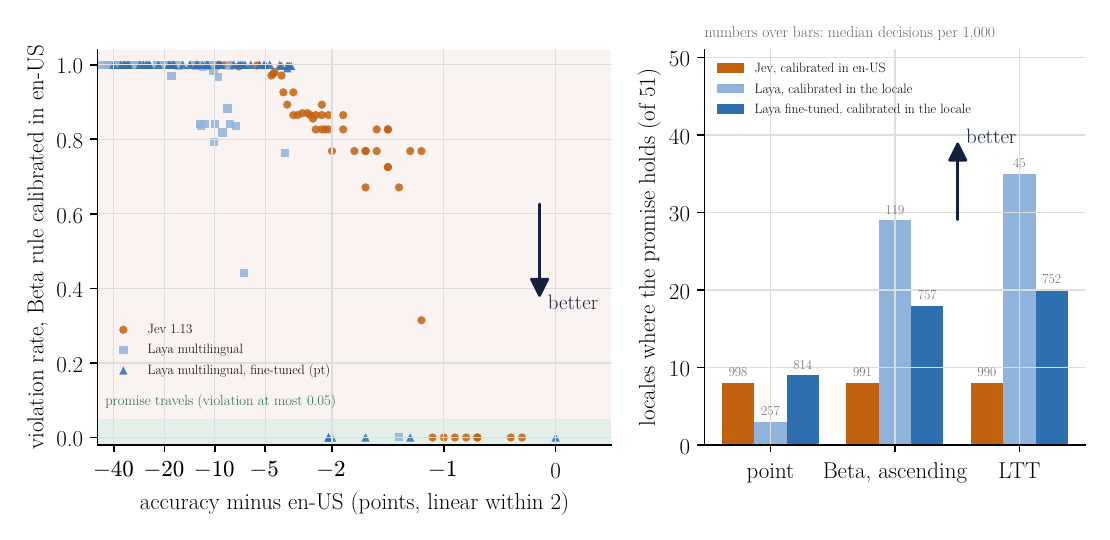}
\caption{The promise of 95\% across the 51 locales of MASSIVE other than en-US, report items. Left: violation rate of the Beta rule calibrated on the en-US calibration items, against the accuracy lost relative to en-US on the same items. The green band is where the promise travels (violation at most 0.05). Jev keeps it only where the loss stays below about one point. Right: locales where the promise holds (upper bound of the violation rate at most 0.05), for Jev calibrated in en-US and for Laya calibrated on the calibration items of each locale, with the median decisions per 1,000 over the bars. Calibrating in the locale recovers the promise for Laya with the bounded rules, at the price of fewer decisions.}
\label{fig:languages}
\end{figure}

\paragraph{Accuracy.} Jev loses 0.3 to 9.4 points against en-US, most in Javanese, Swahili, Khmer and Welsh, and is non-inferior to en-US with the 2-point margin of \Cref{sec:compare} in 6 of the 51 locales. The released Laya loses 1.4 to 47.2 points and is non-inferior in none. Its median loss is 20.7 points in the 23 locales not written in Latin script and 12.4 in the 28 that are, a descriptive contrast, since no grouping was tested. Fine-tuning in Portuguese raises Laya's accuracy in every locale, by 14.9 to 30.5 points with every lower bound above 11.7, so the labels of one language help all the others measured. The fine-tuned checkpoint still loses up to 40.2 points (Amharic) and is non-inferior to en-US only in pt-PT, the language it was tuned on. \Cref{tab:languages} in \Cref{app:languages} gives every locale, and a machine-readable copy is released with the data.

\paragraph{Calibration error stays low, and the promise does not travel.} Jev's ECE stays between 0.009 and 0.030 in every locale, against 0.016 on the en-US items of this front, in line with the stable calibration error that \citet{fadli2026does} report for Jev on Indonesian regional languages. The promise does not follow it. Calibrated in en-US, the promise of 95\% holds in 8 of the 51 locales for Jev, by the point rule, the Beta rule and Learn then Test alike, and breaks in the other 43 with violation rates of 0.31 to 1.00 (\Cref{fig:languages}). The reason is arithmetic. At 95\% Jev accepts more than 99\% of the items in en-US, whose accuracy among accepted items is 96.2\%, so the promise has a slack of 1.2 points. Every locale that loses at most 1.1 points keeps it, and every locale that loses 1.2 points or more breaks it, with a Spearman correlation of $-0.95$ between the loss and the violation rate. The released Laya keeps the promise calibrated in English only in Japanese, and the fine-tuned checkpoint in six locales, pt-PT, es-ES, fr-FR, it-IT, ja-JP and zh-CN. \citet{ahuja2022calibration} found multilingual encoders miscalibrated in low-resource languages and in languages typologically distant from English, and found temperature scaling and a few examples in the language helpful. The measurements here add that a well calibrated probability is not enough. What has to travel is the threshold, and it travels only as far as the slack of the promise allows.

\paragraph{Calibrating in each language.} Laya can be calibrated on the 1,000 calibration items of each locale and checked on its report items. The released checkpoint then keeps the promise in 3 of the 51 locales with the point rule, in 29 with the Beta rule and in 35 with Learn then Test, at median decisions per 1,000 of 257, 119 and 45. The fine-tuned checkpoint keeps it in 9, 18 and 20, at 814, 757 and 752. This check evaluates one fixed report half rather than a plug-in population, so a locale whose accepted items are few is judged on a small sample, and the counts carry that variability. The pattern is nonetheless that of \Cref{sec:gate}, repeated over 51 languages: a bound keeps the promise where the point estimate breaks it, and pays for it in decisions, most where the decider is weakest.

For a caller, a decider that serves several languages needs a calibration sample in each of them. A low ECE in every language does not replace it, and neither does a small loss of accuracy. The slack of the promise in the calibration language, the accuracy among accepted items minus the target, is the loss of accuracy that a promise can absorb, and here it was about one point.

%% ============================================================================
\section{When the domain changes}\label{sec:battery}

\begin{figure}[!htbp]
\centering
\includegraphics[width=\linewidth]{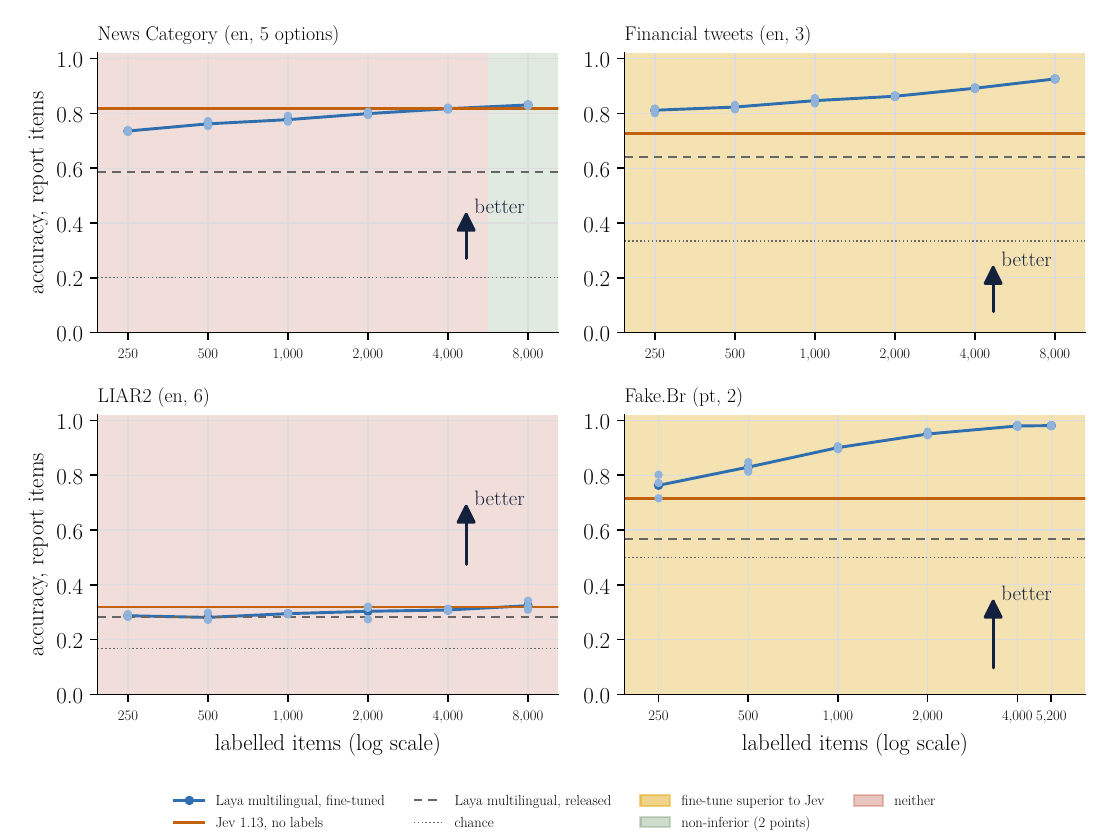}
\caption{Label curves of the battery. Accuracy of the fine-tuned Laya on the report items against the labelled items used (three seeds and their mean), with Jev without labels and the released Laya as reference lines. Gold where the fine-tune is more accurate than Jev with a 95\% interval above zero, green where it is non-inferior with the margin of 2 points, red where it is neither.}
\label{fig:battery}
\end{figure}

Every measurement so far uses intents from MASSIVE. To see which findings belong to the method and which to that dataset, the same code was run on four further domains, registered before any answer was collected and exploratory throughout. News Category headlines \citep{misra2022news} ask for one of five categories drawn from 42. Financial tweets ask whether a tweet is bearish, bullish or neutral (Twitter Financial News sentiment, MIT licence, \url{https://huggingface.co/datasets/zeroshot/twitter-financial-news-sentiment}). LIAR2 \citep{xu2024enhanced} asks for the fact-checker's rating of a political claim on six levels, from the statement alone. Fake.Br \citep{monteiro2018contributions} asks whether a Portuguese news text is fake or real. Its size-normalised texts pair each fake item with a true one on the same topic, and pairs were drawn whole so that no topic is split between the items and the training data. FakeRecogna \citep{garcia2022fakerecogna}, planned for this slot, was replaced before any answer was collected, because 71\% of its fake items carry a verdict marker such as ``\#boato'' in the title, against 1.4\% of its real items. Each domain gives 2,000 items, half for calibration and half for report. Jev~1.13 answered every item, the report items under typos and the mixed profile at 10\%, and the empty state (17,003 calls, US\$0.30). Laya multilingual answered as released and after fine-tuning on 250 to 8,000 labelled items of the domain with three seeds, 32 to 43 seconds of training for the largest size on one GPU. \Cref{tab:battery} gives the reading across domains, \Cref{fig:battery} the label curves and \Cref{app:battery} one report card per domain.

\begin{table}[!htbp]
\centering
\caption{The battery, report items. Columns, from the left: accuracy of Jev, of the released Laya and of the largest fine-tune (mean of three seeds), the labels from which the fine-tune is non-inferior (margin of 2 points) and superior to Jev, Jev's ECE as returned, the cases in which each rule keeps the promise against the plug-in population (of 10, targets 0.90 and 0.95 for five deciders counting each seed) with the tolerance of \Cref{sec:gate}, the median decisions per 1,000 of the fine-tunes at 0.95, and the seeds in which the Beta promise of the fine-tune at 0.95, calibrated on clean text, survives typos at 10\%. The first row is MASSIVE pt-PT measured with the same code and grid, with Jev's earlier answers reused. In LIAR2 no rule accepts any item, so every promise is kept by abstention. In bold, per row, the highest accuracy, the rule that keeps the promise most often and the rule that decides most, where they differ.}
\label{tab:battery}
\scriptsize
\setlength{\tabcolsep}{2.6pt}
\begin{tabular}{lccccccccccccc}
\toprule
 & & \multicolumn{3}{c}{Accuracy} & \multicolumn{2}{c}{Labels for} & & \multicolumn{3}{c}{Promise kept} & \multicolumn{2}{c}{Decisions} & Typos \\
\cmidrule(lr){3-5}\cmidrule(lr){6-7}\cmidrule(lr){9-11}\cmidrule(lr){12-13}
Domain & Opt. & Jev & Laya & Laya-ft & NI & sup. & ECE Jev & point & Beta & LTT & Beta & LTT & \\
\midrule
MASSIVE pt (ref.) & 5 & 0.947 & 0.732 & \textbf{0.960} & 4,000 & none & 0.020 & 7 & 8 & \textbf{10} & 996 & 996 & 0 of 3 \\
\midrule
News Category & 5 & 0.817 & 0.585 & \textbf{0.831} & 8,000 & none & 0.076 & 0 & 6 & \textbf{9} & \textbf{517} & 331 & 0 of 3 \\
Financial tweets & 3 & 0.728 & 0.642 & \textbf{0.926} & 250 & 250 & 0.171 & 3 & 9 & \textbf{10} & \textbf{842} & 532 & 0 of 3 \\
LIAR2 & 6 & 0.319 & 0.282 & \textbf{0.324} & none & none & 0.162 & 10 & 10 & 10 & 0 & 0 & no decisions \\
Fake.Br (pt) & 2 & 0.716 & 0.567 & \textbf{0.981} & 250 & 250 & 0.087 & 7 & \textbf{10} & \textbf{10} & 999 & 999 & 3 of 3 \\
\bottomrule

\end{tabular}
\end{table}

\paragraph{Labels decide the comparison, and the domain decides the labels.} The fine-tune is more accurate than Jev from 250 labels on financial tweets and on Fake.Br, reaching 0.926 against 0.728 and 0.981 against 0.716. On news headlines it needs 8,000 labels to be non-inferior and is not detectably more accurate. On LIAR2 it is never non-inferior, and neither model moves far from always answering ``false'', which is right for 28.5\% of the report items. The 4,000 labels of MASSIVE are therefore a property of that task, not a constant of the method, and the label curve has to be measured in the domain at hand.

\paragraph{Calibration does not travel either.} Jev's ECE, 0.020 on MASSIVE in Portuguese, the reference row of the battery, is 0.076 on news headlines, 0.087 on Fake.Br, 0.162 on LIAR2 and 0.171 on financial tweets, where 0.82 of confidence goes with 0.60 of accuracy among the most confident LIAR2 decisions. Laya after the temperature fitted on the calibration half stays between 0.006 and 0.048. A calibration measured on one task licenses nothing about another, and the temperature, fitted on a few hundred labelled items of the domain, is the cheap repair.

\paragraph{The gate.} Across the four domains, the point rule keeps the promise in 20 of 40 cases and the bounded rules in most: Learn then Test in 39 and the Beta rule in 35. The five Beta cases outside the tolerance have estimated violation rates of 0.042 to 0.050, above 0.05 only by the upper bound of 2,000 draws, and four of them are fine-tunes on news headlines or financial tweets. At 0.95 the fine-tunes decide a median of 517 items per 1,000 with the Beta rule against 331 with Learn then Test on news headlines, and 842 against 532 on financial tweets. The trade measured on MASSIVE recurs: more decisions for a tolerated excess of risk that sits at the tolerance, not beyond it.

\paragraph{Typos break the promise again.} Calibrated on clean text, the 0.95 promise of the fine-tune breaks under typos in every seed on news headlines and financial tweets, with violation rates of 0.21 to 0.92. Jev's promise breaks as well where Jev decides, 0.20 at 0.95 and 0.50 at 0.90 on news headlines, although its accuracy under typos changes by $-0.5$ points with an interval containing zero. As on MASSIVE, a loss too small to detect in accuracy is enough to break a promise that sits on the accepted items. On Fake.Br typos cost nothing, but the mixed profile drops the fine-tune by 34 to 40 points and breaks its promise in every draw, which the next paragraph explains.

\paragraph{What the fine-tune read in Fake.Br.} Three tests registered after this result separate form from content. Removing punctuation alone costs the fine-tune 34.3 points (95\% interval $[31.4,37.2]$), while lower case costs 0.7, removing accents 0.4 and typos 0.5. A logistic regression on surface features only, with no word in it, reaches 0.807 ($[0.781,0.830]$): fake items carry more quotation marks, exclamation marks and full stops, true items more commas and hyphens. And Laya fine-tuned on text without punctuation or capitals still reaches 0.931 on such text. The wording carries most of the signal, and the fine-tune leaned on punctuation because the training data offered it, which a validation split from the same source cannot reveal. Removing the form from the training data costs 7 points on the original text (0.909 against 0.983) and keeps 0.931 on text without it, where the original fine-tune falls to 0.606, the same exchange as the training with noise of \Cref{sec:noise}.

\paragraph{The empty state follows the training labels.} On news headlines, whose options change from item to item, the fine-tune answers 0.32 to 0.37 of the report items correctly with an empty state, against 0.17 before fine-tuning and 0.23 for Jev, and picks the most frequent training category among the options in 45\% to 48\% of the items, the pattern of MASSIVE. Where the options are fixed, the empty state yields one answer for all items: the majority class for every model on financial tweets (neutral, 64.5\%), and on LIAR2 ``false'' for Jev and two of the three fine-tunes.

\paragraph{When no promise can be made.} On LIAR2 no rule accepts a single item at 0.90 or 0.95, and \Cref{prop:ceiling} gives a reason compatible with this: no rule can promise more than the accuracy the input allows. Among the 20 most confident report items Jev is right in 45\% and the fine-tunes in 60\% to 65\%, so no threshold isolates a group that reaches 90\%. The authors of LIAR2 report 0.32 for the statement alone with their own model trained on the full training split, the accuracy reached here, which is compatible with a low ceiling for that input, though the ceiling itself is not observed. The measures that follow were taken after this result and are exploratory. Better estimates move a decider towards the ceiling: the mean of Jev's and the fine-tunes' probabilities is right in 0.349 of the report items and in 67\% of its 100 most confident, against 50\% and 60\% for each alone. More information in the state can raise the ceiling, but only if it is informative and clean. The speaker record published with LIAR2 counts the rating of the statement itself, which is at least one in every row, so it was replaced by the record of the speaker in a disjoint half of the training split. A record that left out only the row itself let the fine-tunes identify training labels through the speaker and was not used. With the clean record in a structured state, accuracy changes by $+2.1$ points for Jev ($[-0.6,4.8]$) and by $-2.5$ to $+1.7$ for the fine-tunes, all with intervals containing zero. A coarser question raises the ceiling at once: on two levels, the false side against the true side, whose majority is 0.577, the 100 most confident decisions are right in 95\% to 97\% of the items, so a promise of 90\% becomes possible for a tenth of them, on a question that no longer asks for the rating. And a set-valued answer always keeps its promise: split conformal sets cover the rating in 89\% to 91\% of the report items at the 90\% level, with 3.9 to 4.6 of the six ratings in a set. The size of the set is the measure of how little the statement says.

For a caller, the battery turns the findings of MASSIVE into checks that a deployment can run in its own domain with \texttt{promise\_check}: the label curve against the closed model, the calibration of each model on the domain, the gate with a bounded rule, the promise under the noise the traffic will carry, a test of what the fine-tune reads, and the ceiling of the promise when no rule decides.

%% ============================================================================
\section{When a Bayesian retraining loop is useful under drift}\label{sec:drift}

Every rule of \Cref{sec:gate} assumes that the items used to choose the threshold and the items the gate later decides are exchangeable. A deployed decider loses that assumption when its inputs or the meaning of its labels change \citep{gama2014survey,lu2018learning}, and the promise of the gate then has to be kept by watching it, retraining when it breaks and qualifying the threshold again. Deciding to act only when a change harms the deployed model, rather than whenever the inputs move, is the question of sequential risk tracking \citep{podkopaev2022tracking}. A Bayesian loop for the three steps is studied here. Its online update, exponential forgetting of Beta evidence with the Laplace predictive, is the Beta reputation system of \citet{josang2002beta}, the update of discounted Thompson sampling \citep{raj2017taming} and the binomial case of the discount-factor dynamic models of \citet{westharrison1997bayesian}, and what the loop adds is its coupling to the acceptance gate, with a trigger that chooses between retraining and re-qualifying the threshold. This section asks when that loop is useful compared with the state-of-the-art alternatives for the same task, measured on streams with a known change, under a preregistration frozen before any pilot run. Every implementation choice the preregistration left open was recorded before the run it affects.

\subsection{The loop and its comparators}

The decider is the multilingual Laya checkpoint on five-option intent questions built as in \Cref{sec:setup}, with the promise of an accuracy of at least $\alpha=0.95$ among the items it decides alone and $q=\delta=0.05$. Labels come from two sources, each at the cost of one review: every escalated item, and a uniform audit of 5\% of the accepted items. Escalated items are selected by low confidence and misrepresent the error rate among accepted items, so the drift monitors read the audit labels only, while promotion and re-qualification use all labels.

The loop has three Beta steps.
\begin{itemize}
  \item \emph{Trigger.} The audited accepted items give a Beta posterior of the accuracy among accepted items, with counts discounted by a factor $\lambda$ per window of 500 items, and retraining starts when $P(\theta<\alpha)\ge1-q$.
  \item \emph{Promotion.} The retrained and the current model are compared on a hold-out of recent labelled items, and the new model replaces the current one when $P(\text{new better})\ge1-q$ under $\mathrm{Beta}(b+1,c+1)$ on the $b$ and $c$ discordant pairs.
  \item \emph{Re-qualification.} The threshold is chosen again with the Beta lower-bound rule of \Cref{sec:gate-def}.
\end{itemize}

Eight comparators share the decider, the retraining procedure, the audit rate and the label stream. The frozen model never retrains. The periodic arm retrains every $K$ items. Recalibration only re-estimates the accuracy per confidence bin by Laplace's rule of succession. DDM \citep{gama2004learning}, ADWIN \citep{bifet2007learning} and Page--Hinkley \citep{page1954continuous} watch the audited error stream, with promotion always, by the exact McNemar test \citep{dietterich1998approximate} or by the high-confidence safety test of \citet{thomas2019preventing}, as tuning options. Two arms never retrain and only move the threshold, Learn then Test qualified again in every window and adaptive conformal inference (ACI) \citep{gibbs2021adaptive}. Every retrain fine-tunes the current model for two epochs on the most recent 4,000 labelled items with 20\% replay of the initial training set, and the arms that do not use the Beta rule re-qualify with Learn then Test.

\subsection{Streams and measures}

Three synthetic drifts are built from MASSIVE. In D1, five of the 18 scenarios, drawn before freezing, are absent from the initial training and enter the stream after the change, a knock-out shift reversed \citep{rabanser2018failing}. In D2, the language changes from Portuguese to English on parallel items. In D3, two pairs of intents, also drawn before freezing, swap their labels, a change of concept rather than of inputs. Each stream has 9,000 items, with the change at a point drawn uniformly between items 3,000 and 5,000, abrupt or gradual over 1,000 items \citep{gama2014survey}. A stream without drift measures false retrains. Thirty replicates per drift and pattern give 1,890 runs over the nine arms. The real streams use the News Category Dataset \citep{misra2022news}, HuffPost headlines from 2012 to 2022 whose earlier years form the HuffPost part of Wild-Time \citep{yao2022wild}. Each of ten transitions between consecutive years is a stream family with five random orders and the change on 1 January. Temporal change does not always degrade a model \citep{agarwal2022temporal}, so a transition counts only if the frozen model loses at least 2 points of accuracy across it.

The primary measure is the cost per item at a cost ratio of 10, where a wrong automatic action costs 10 and each escalation and each audit label costs 1, with retraining free in the primary analysis and costing 50 in a sensitivity analysis. The promise is checked on the true selective accuracy of each post-change window, measured on the test split reweighted to the stream's current mixture. Twenty configurations per arm were tuned on separate streams. The loop selected $\lambda=0.95$ and $q=0.01$, both at the edge of the preregistered grid, and the best comparator was the periodic arm with $K=3{,}000$ and promotion always, at a tuning cost of 0.618 against 0.611 for the loop.

The loop counts as useful, which is the claim the preregistration ties to the title, if its cost is lower than the best comparator's with a paired 95\% interval excluding zero in at least two of the three drifts, if the share of post-change windows below $\alpha$ exceeds the comparator's by at most 0.02, and if it shows no more false retrains on the control stream. The real streams are supporting evidence. In the main run, the true selective accuracy of the frozen model moves by $+0.1$ points across the language change of D2, against $-6.0$ for D1 and $-9.5$ for D3, so D2 is read as a drift that does not degrade the model. The pilot value for D2 is not used, because the pilot cached encodings by item identifier and MASSIVE gives the same identifier to an utterance in every locale. The main run keys the cache by content.

\subsection{Results}

\begin{table}[!htbp]
\centering
\caption{Drift experiment, main run: mean over the 60 streams of each drift (30 replicates, abrupt and gradual) and the 30 control streams. Cost per item at $C_e/C_r=10$, share of post-change windows of 500 items whose true selective accuracy is below $\alpha=0.95$, and retrains per stream. In bold, the lowest cost and the fewest windows below $\alpha$ per column. Retrains are not ranked, because under a drift that degrades the model fewer retrains is not better.}
\label{tab:drift}
\small
\begin{tabular}{lcccccccccc}
\toprule
 & \multicolumn{4}{c}{Cost per item} & \multicolumn{3}{c}{Windows below $\alpha$} & \multicolumn{3}{c}{Retrains} \\
\cmidrule(lr){2-5}\cmidrule(lr){6-8}\cmidrule(lr){9-11}
Arm & D1 & D2 & D3 & control & D1 & D2 & D3 & D1 & D3 & control \\
\midrule
Author's loop & 0.764 & \textbf{0.402} & 0.825 & \textbf{0.433} & 0.86 & \textbf{0.00} & 0.93 & 0.9 & 1.2 & 0.0 \\
Periodic ($K=3{,}000$) & 0.720 & 0.466 & 0.806 & 0.482 & 0.83 & 0.26 & 0.62 & 3.0 & 3.0 & 3.0 \\
DDM & 0.843 & 0.448 & 0.838 & 0.500 & 0.91 & 0.08 & 0.65 & 2.2 & 2.1 & 2.0 \\
ADWIN & 0.795 & 0.427 & 0.883 & 0.475 & 0.88 & 0.03 & 0.96 & 0.9 & 0.0 & 0.0 \\
Page--Hinkley & 0.815 & 0.428 & 0.843 & 0.476 & 0.93 & 0.05 & 0.77 & 1.2 & 1.0 & 0.2 \\
Recalibration only & 0.742 & 0.406 & 0.781 & 0.436 & 0.57 & \textbf{0.00} & 0.38 & 0.0 & 0.0 & 0.0 \\
LTT per window & \textbf{0.668} & 0.413 & \textbf{0.773} & 0.442 & \textbf{0.37} & \textbf{0.00} & \textbf{0.21} & 0.0 & 0.0 & 0.0 \\
ACI & 0.917 & 0.430 & 0.879 & 0.478 & 0.94 & 0.03 & 0.95 & 0.0 & 0.0 & 0.0 \\
Frozen & 0.920 & 0.427 & 0.883 & 0.475 & 0.94 & 0.03 & 0.96 & 0.0 & 0.0 & 0.0 \\
\bottomrule

\end{tabular}
\end{table}

By the preregistered rule, the loop does not improve on the best comparator selected in tuning, the periodic arm (\Cref{tab:drift}). Paired over the 60 streams of each drift, its cost is higher in D1 by 0.044 (95\% interval $[0.002,0.087]$) and in D3 by 0.020 ($[0.004,0.035]$), and lower only in D2, by 0.064 ($[-0.079,-0.049]$). Non-inferiority on the windows below the promise holds in D2 and fails in D1 (difference 0.034, upper bound 0.113) and D3 (0.305, upper bound 0.387). On the control stream the loop never retrains, against three retrains of the periodic arm. The preregistered sensitivity analysis charges 50 reviews per retrain. Under it the difference against the periodic arm is no longer detected in D1 (0.033, $[-0.009,0.075]$) or D3 (0.010, $[-0.006,0.025]$), and the loop is cheaper by 0.081 in D2 and 0.066 on the control stream (both intervals below $-0.04$), so where retraining has a real cost the loop is no longer detectably more expensive than the periodic arm on any stream type, which is not the same as equivalent. The claim the preregistration ties to the title is therefore not supported, and the title does not make it.

The question of when the loop is useful has a sharper answer than the rule. The loop pays off where the stream does not degrade the model. On the control stream and on D2, the change of language that costs the frozen model almost nothing, it has the lowest mean cost of the nine arms, 0.433 and 0.402, because it does not retrain where there is nothing to repair, while the periodic arm retrains three times and DDM between 1.5 and 2 times. Against recalibration only and Learn then Test per window, which do not retrain either, no difference in cost is detected. Where the drift degrades the model, the loop reacts late and little. Its median first retrain comes 1,924 items after the change in D1 and 2,792 in D3, with 0.9 and 1.2 retrains per stream, and 86\% and 93\% of the post-change windows stay below the promise. The trigger reads only the audited accepted items, about 25 per window, and it asks for a posterior probability of at least 0.99 that the promise is broken. No arm that retrains keeps the promise under D1 or D3. The periodic arm, the best of them, still leaves 83\% and 62\% of the windows below $\alpha$, with three retrains and three promotions per stream.

The arms that do best under a degrading drift move only the threshold. Learn then Test qualified again in every window has the lowest cost in D1 and D3 and the fewest windows below the promise, 37\% and 21\%, and it pays for this in reviews, with 302 and 572 labels per 1,000 items against 110 and 133 for the loop. At a cost ratio of 100 it is the cheapest arm on all four stream types. Paired with the loop, as an exploratory comparison, its cost is lower by 0.096 in D1 ($[0.067,0.126]$) and by 0.052 in D3 ($[0.036,0.068]$), with intervals containing zero in D2 and on the control stream. This arm ranked last in tuning, on shorter streams of 4,000 items that change earlier, so it was not the preregistered comparator, and whether the length of the tuning streams explains the reversal was not tested. Recalibration alone, sits between the two in D1 and D3 and is close to the loop where nothing degrades.

\begin{table}[!htbp]
\centering
\caption{Efficiency at equal promise in the drift experiment (exploratory, 64 comparisons, \Cref{sec:compare}). Outcome: share of windows below $\alpha$ (post-change windows, all windows on the control stream), margin 0.02. Resources: labels per 1,000 items, retrains and GPU seconds, paired over 60 streams per drift and 30 control streams. C: clean efficiency gain. T: trade-off, one resource saved and another resource or the error rate larger. E: equivalent, no resource saved. NI: non-inferior, no resource saved. --: non-inferiority not shown, because the method is worse or the interval is too wide. Clean gains, the best outcome of a comparison, in bold.}
\label{tab:drift-eff}
\small
\begin{tabular}{llcccc}
\toprule
Method & Against & D1 & D2 & D3 & control \\
\midrule
Author's loop & Periodic & -- & \textbf{C} & -- & \textbf{C} \\
Author's loop & DDM & \textbf{C} & \textbf{C} & -- & \textbf{C} \\
Author's loop & ADWIN & -- & NI & NI & -- \\
Author's loop & Page--Hinkley & T & T & -- & T \\
Author's loop & Recalibration only & -- & E & -- & E \\
Author's loop & LTT per window & -- & T & -- & T \\
Author's loop & ACI & NI & NI & NI & NI \\
Author's loop & Frozen & NI & NI & NI & -- \\
\midrule
Recalibration only & Author's loop & T & E & T & E \\
Recalibration only & Periodic & T & \textbf{C} & T & \textbf{C} \\
Recalibration only & DDM & T & \textbf{C} & T & \textbf{C} \\
Recalibration only & ADWIN & T & NI & NI & -- \\
Recalibration only & Page--Hinkley & T & T & T & T \\
Recalibration only & LTT per window & -- & T & -- & T \\
Recalibration only & ACI & NI & NI & NI & NI \\
Recalibration only & Frozen & NI & NI & NI & -- \\
\bottomrule

\end{tabular}
\end{table}

The comparison protocol of \Cref{sec:compare} makes the trade explicit (\Cref{tab:drift-eff}). Where the stream does not degrade the model, the loop is a clean efficiency gain over the periodic arm and DDM, with the same or a better promise and no retrains, and it is equivalent to recalibration only. Against Learn then Test per window it saves 94 labels per 1,000 items in D2 and 97 on the control stream at equal promise, but accepts 8 and 9 more wrong actions per 1,000, a trade-off. Under D1 and D3 it is not non-inferior to the periodic arm, to recalibration only or to Learn then Test per window. Recalibration only, Laplace's rule per bin with forgetting, is non-inferior to every arm except Learn then Test per window under D1 and D3, never retrains, and is a clean gain over the periodic arm and DDM on the stable streams. With Holm's correction over the 64 comparisons, 19 of the 27 gains remain. The eight lost are against DDM and Page--Hinkley, and every gain of the loop and of recalibration over the periodic arm on the stable streams remains. Under degradation its gains are trade-offs, fewer retrains and less GPU time paid with more labels, 190 more per 1,000 items than the loop in D1.

For a caller, the reading is the following. When the stream is stable or its change is harmless, the loop avoids needless retraining at the lowest cost measured. When the change degrades the model and escalation is affordable, re-qualifying the threshold every window keeps the promise more often than any retraining arm, at the price of more reviews. Even the best arm leaves 37\% of the post-change windows of D1 below the promise, so a degrading drift of this size calls for a human decision on the model, not only for an automatic loop.

\paragraph{Which component drives the loop.} The attribution ablation swaps one component of the loop at a time on the same 60 streams per drift, and credits a component when swapping it raises the cost with an interval above zero. Only the trigger is credited. Replacing the posterior under the uniform prior by the one-sided Clopper--Pearson bound on the same counts makes the loop retrain less, 0.6 and 0.8 fewer times per stream, and raises its cost by 0.046 in D1 ($[0.025,0.066]$) and 0.012 in D3 ($[0.006,0.017]$). The other three components work against the loop. Counting per window instead of discounting lowers the cost in D1 by 0.048 ($[-0.068,-0.029]$), with 0.7 more retrains per stream. Promoting every retrained model lowers it in D1 by 0.019 and in D3 by 0.008, so the Beta test on discordant pairs rejects useful models. Re-qualifying the threshold with Learn then Test leaves the cost unchanged but lowers the share of post-change windows below the promise by 6 points in D1, with an upper bound just below zero, and by 22 points in D3 ($[-30,-14]$). The weak link of the loop is therefore its last step. After a promotion, the threshold is chosen again on the most recent labels, which over-represent escalated, hard items, and the ascending scan of the Beta rule is exactly the procedure that \Cref{prop:scan} shows to lose its guarantee when the qualification pattern is not monotone. In D2 no swap changes the cost by more than 0.001.

\paragraph{Repairing the re-qualification.} A second ablation, designed after the first and reported in \Cref{app:repair}, tests repairs of the weak step. Re-qualifying on items held out of the retraining set repairs the loop on D1, and the fixed-sequence scan and weights for the probability of being labelled repair it on D3.

\subsection{When the drift has a pattern}\label{sec:drift-pattern}

The loop and the recalibration arm track the accuracy of each confidence bin with a forgetting Beta, a local-level model that can follow a change only after it has happened. Many deployments meet changes that recur or announce themselves: traffic that alternates between languages, noise that rises at some hours, an input feature that shifts before the accuracy does. A dynamic generalised linear model with binomial observations \citep{west1985dynamic,westharrison1997bayesian} keeps the per-bin tracking and lets trend, seasonality and inputs enter, so that the gate can use the accuracy predicted for the next window instead of the accuracy estimated for the last one.

\paragraph{The model.} For confidence bin $b$ and window $t$, let $\theta_{b,t}$ be the accuracy of the items in the bin and $\eta_{b,t}=\operatorname{logit}\theta_{b,t}$. The state $\beta_{b,t}$ holds a level, a harmonic pair of known period $P$ and the coefficients of window inputs $x_t$ computed from the texts alone,
\begin{equation}
  \eta_{b,t}=F_t^\top\beta_{b,t},\qquad \beta_{b,t}=G\,\beta_{b,t-1}+\omega_{b,t},\qquad y_{b,t}\mid\theta_{b,t}\sim\mathrm{Bin}(n_{b,t},\theta_{b,t}),
  \label{eq:dglm}
\end{equation}
with $F_t=(1,1,0,x_t^\top)^\top$, $G$ block diagonal with $1$, the rotation by $2\pi/P$ and the identity, and the evolution variance set by component discount factors, so that the prior variance of the state is $R_t=\Delta^{-1/2}GC_{t-1}G^\top\Delta^{-1/2}$ with $\Delta$ the diagonal of discounts. Given the prior moments $f=F_t^\top a_t$ and $q=F_t^\top R_tF_t$ of $\eta_{b,t}$, conjugate updating \citep{west1985dynamic} represents the prior of $\theta_{b,t}$ by the Beta whose logit has these moments,
\begin{equation}
  \psi(r)-\psi(s)=f,\qquad \psi'(r)+\psi'(s)=q,
  \label{eq:betamoments}
\end{equation}
with $\psi$ the digamma function, updates it to $\mathrm{Beta}(r+y,\,s+n-y)$ with the labels of the window, and returns to the state through the posterior moments $(f^*,q^*)$ of the logit: $m_t=a_t+R_tF_t(f^*-f)/q$ and $C_t=R_t-R_tF_tF_t^\top R_t(1-q^*/q)/q$. West, Harrison and Migon solve \cref{eq:betamoments} by $r=(1+e^f)/q$ and $s=(1+e^{-f})/q$, which is accurate only for small $q$. With a few labels per bin and window, as an audit of 5\% split into 20 bins gives, $q$ stays near one, the approximation returns $s<1$, and every window without an error pushes the logit up without bound. The adapter therefore solves \cref{eq:betamoments} exactly by Newton's method.

\paragraph{The gate.} Before window $t$ is gated, its inputs $x_t$ are computed and each bin receives its one-step-ahead predictive $\mathrm{Beta}(r_{b,t},s_{b,t})$. The set $A_t$ of accepted bins holds the bins whose predictive $q_b$-quantile is at least $\alpha$, for a fixed level $q_b$ per bin, or, in the variant that matches the recalibration arm, whose predictive mean $r_{b,t}/(r_{b,t}+s_{b,t})$ is. When $A_t$ is empty the gate acts on nothing in window $t$ and the promise holds vacuously. The lower-bound variant run in the experiments uses $q_b=0.05$ in every bin.

\begin{proposition}[When the dynamic gate keeps the promise]\label{prop:dyngate}
Let $w_{b,t}\ge0$ be the share of the items of window $t$ that fall in bin $b$, and $\Theta_t=\sum_{b\in A_t}w_{b,t}\theta_{b,t}/\sum_{b\in A_t}w_{b,t}$ the accuracy among accepted items.
\begin{enumerate}
  \item[(i)] If $\theta_{b,t}\ge\alpha$ for every $b\in A_t$, then $\Theta_t\ge\alpha$. The condition is sufficient, not necessary.
  \item[(ii)] If \cref{eq:dglm} holds as a Bayesian model and every bin is gated at level $q_b=q/B$, with $B$ the number of bins, then, whenever $A_t$ is non-empty, $\Prob(\Theta_t\ge\alpha\mid\text{labels up to }t-1\text{ and inputs up to }t)\ge1-q$. The run with $q_b=0.05$ controls each bin, not their mixture.
  \item[(iii)] With the level alone, a single discount $\lambda$ and exact moment matching, and to first order in $1/r$ and $1/s$, one window of evolution and updating maps $(r,s)$ to $(\lambda r+y,\ \lambda s+n-y)$, the forgetting Beta with the same $\lambda$.
\end{enumerate}
\end{proposition}

Part (iii) is an approximation to first order, not an identity. No component can anticipate a change of $\theta_{b,t}$ that the past labels and the inputs $x_t$ carry no information about, since by part (iv) of \Cref{prop:ceiling} only information that predicts $\theta_{b,t}$ raises what a gate can promise. Part (i) explains why per-bin gates decide less than threshold gates at the same target: they ask every accepted bin to reach the target, while the promise asks it only of their mixture. Part (ii) is a Bayesian statement under a correctly specified model, not a finite-sample frequentist guarantee, and it does not survive a misspecified evolution. Part (iii) places the forgetting Beta of the loop and of the recalibration arm inside the family as its simplest member. On simulated Bernoulli streams with a level, a ramp and a seasonal path, the two give predictive means within 0.010 of each other on average with 5 labels per window and within 0.0006 with 100, for $\lambda$ between 0.8 and 0.95. The proof is in \Cref{app:proofs}, and parts (i) and (ii) are checked in Lean~4 with Mathlib (\texttt{proofs/DynamicGate.lean}). The closest frequentist counterpart is conformal PID control \citep{angelopoulos2023conformal}, whose scorecaster forecasts the next conformal score with seasonal and exogenous models, building on adaptive conformal inference \citep{gibbs2021adaptive}. It controls the coverage of intervals rather than the accuracy among accepted decisions.

\paragraph{Streams with a pattern.} Two streams of 9,000 items add a known cycle to the setting of \Cref{sec:drift}. In S1, Portuguese items are mixed with the same items after typing noise at 20\%, and the share of noisy items at item $t$ is $\tfrac12\big(1-\cos(2\pi t/2000)\big)$, so the cycle degrades the model and recurs every four windows. In S2, Portuguese and English items alternate with the same share, a cycle that, like D2, does not degrade the model. The seasonal variants are given the period, as a deployment would know a daily or weekly cycle, and the inputs $x_t$ are two window means computed from the texts alone, the share of words absent from the Portuguese training vocabulary and the share of 30 frequent English function words. The level discount, tuned on the tuning streams of \Cref{sec:drift}, is $\delta=0.80$. Each comparison follows the protocol of \Cref{sec:compare}, paired over 30 streams per cycle and 60 per drift, with the margin of 0.02 on the share of windows below $\alpha$. It is exploratory, written before the runs, and not corrected for multiplicity.

\begin{table}[!htbp]
\centering
\caption{The dynamic adapter on streams with a pattern and under the degrading drifts: mean cost per item and share of windows below $\alpha=0.95$ (all windows on S1 and S2, post-change windows on D1 and D3), over 30 streams per cycle and 60 per drift. The seasonal variants were run on S1 and S2 only. In bold, the lowest value per column.}
\label{tab:adapter}
\small
\begin{tabular}{lcccccccc}
\toprule
 & \multicolumn{2}{c}{S1, noise cycle} & \multicolumn{2}{c}{S2, language cycle} & \multicolumn{2}{c}{D1} & \multicolumn{2}{c}{D3} \\
\cmidrule(lr){2-3}\cmidrule(lr){4-5}\cmidrule(lr){6-7}\cmidrule(lr){8-9}
Arm & Cost & Below $\alpha$ & Cost & Below $\alpha$ & Cost & Below $\alpha$ & Cost & Below $\alpha$ \\
\midrule
Recalibration only & 0.614 & 0.51 & \textbf{0.407} & \textbf{0.00} & 0.742 & 0.57 & 0.781 & 0.38 \\
Recalibration, $\lambda=0.80$ & 0.646 & 0.52 & 0.420 & \textbf{0.00} & 0.728 & \textbf{0.26} & \textbf{0.762} & 0.22 \\
LTT per window & \textbf{0.493} & \textbf{0.01} & 0.420 & \textbf{0.00} & \textbf{0.668} & 0.37 & 0.773 & \textbf{0.21} \\
Author's loop & 0.578 & 0.45 & 0.408 & \textbf{0.00} & 0.764 & 0.86 & 0.825 & 0.93 \\
Periodic & 0.578 & 0.44 & 0.456 & 0.22 & 0.720 & 0.83 & 0.806 & 0.62 \\
Dynamic, level & 0.642 & 0.51 & 0.417 & \textbf{0.00} & 0.729 & 0.28 & 0.763 & 0.23 \\
Dynamic, seasonal & 0.716 & 0.32 & 0.481 & \textbf{0.00} & -- & -- & -- & -- \\
Dynamic, seasonal and inputs & 0.717 & 0.31 & 0.477 & \textbf{0.00} & 0.764 & 0.33 & 0.824 & 0.28 \\
Dynamic, same, lower bound & 0.905 & 0.08 & 0.707 & \textbf{0.00} & -- & -- & -- & -- \\
\bottomrule

\end{tabular}
\end{table}

\paragraph{What the adapter changes.} Where the cycle or the change does not degrade the model, the dynamic variants only add cost (\Cref{tab:adapter}). On S2, D2 and the control stream, recalibration keeps the promise in every window, and against it the level model costs 0.007 to 0.010 more per item and the seasonal models about 0.07 more, all paired intervals above zero, because they escalate more without any window to repair.

Where the cycle degrades the model and its period is known, the seasonal components help the promise and pay for it in reviews. On S1 they cut the share of windows below $\alpha$ from 51\% under recalibration to 32\% (paired difference $-0.19$, $[-0.25,-0.12]$), and the inputs add nothing measurable (31\%). The gain comes with 267 more labels per 1,000 items, which buy 16 fewer wrong actions per 1,000 and raise the cost by 0.10 ($[0.09,0.11]$), a trade-off. The lower-bound variant brings the windows below $\alpha$ to 8\%, at a cost of 0.91. Learn then Test qualified again in every window does better on this stream than any dynamic variant, with 0.7\% of windows below $\alpha$ at a cost of 0.49 and 334 labels per 1,000 items. When the cycle is fast, re-qualifying the threshold every window serves the promise better than predicting the accuracy of each bin.

Under the degrading drifts, the level model keeps the promise more often than recalibration and at a lower cost. The share of post-change windows below $\alpha$ falls from 57\% to 28\% in D1 ($[-0.34,-0.25]$) and from 38\% to 23\% in D3 ($[-0.18,-0.12]$), with the cost lower by 0.014 and 0.017 (both intervals below zero) and 147 and 80 more labels per 1,000 items. In D3 it costs 0.010 less than Learn then Test per window ($[-0.020,-0.0005]$, exploratory) with 82 fewer labels per 1,000 items, but its share of windows below $\alpha$, 23\% against 21\%, is not shown to be non-inferior.

The gain comes from forgetting faster, not from the dynamic model. By part (iii) of \Cref{prop:dyngate}, the level model with $\delta=0.80$ is, to first order, the forgetting Beta with $\lambda=0.80$, while the recalibration arm was tuned on a grid that stopped at $\lambda=0.90$, the value it chose. Recalibration with $\lambda=0.80$, run on the same streams to separate the two, matches the level model: their costs differ by at most 0.004 per item on every stream, the level model is equivalent to it on S1, S2 and the control stream, leaves 2.2 more points of post-change windows below $\alpha$ in D1 ($[0.6,3.8]$) and shows no detectable difference in D3. Against the tuned recalibration, the faster forgetting lowers the share of windows below $\alpha$ by 31 points in D1 and 16 in D3 at a lower cost, and costs 0.011 to 0.013 more per item on the streams that do not degrade the model. The tuning grid of the recalibration arm stopped short of the better rate.

For a caller, the reading is the following. A known cycle that degrades the model justifies seasonal components only when more reviews are affordable, and even then re-qualifying a threshold every window is the stronger option. A cycle or a change that does not degrade the model needs no dynamic component. Under an abrupt degradation, what helps the per-bin gate is to forget faster, whatever model implements it.

\subsection{Numeric decisions over real time}\label{sec:m3}

The streams above are built from text. To see a gate on decisions about numbers, over real time and in a domain with financial series, we use the 1,428 monthly series of the M3 competition \citep{makridakis2000m3}. The decision is whether the next value of a series exceeds its last one, at each of its last 36 time points, so that each time point, an origin, is a window of 1,428 decisions. The decider is exponential smoothing with additive error, damped trend and seasonality of period 12 \citep{hyndman2002state}, refitted at every origin on the last 48 observations, whose probability of a rise is the predictive probability that the next value exceeds the last. Threshold gates are calibrated once on origins 1 to 12, or on the 12 origins that precede each origin and slid by one. The per-bin gates of \Cref{sec:drift-pattern} are warmed up on origins 1 to 12, and the dynamic model gets a seasonal pair of period 12 origins. The promise is the accuracy among acted decisions in each origin, for targets of 0.60, 0.70 and 0.80. Jev and Laya decide on a sample of 150 series, 25 per category, at the last 12 origins, with the last 24 values as the state. Laya is fine-tuned in sliding blocks, on 8,000 decisions of the other series at the 24 origins before each block of six test origins, with three seeds. The analysis is exploratory and was written before the runs.

\begin{table}[!htbp]
\centering
\caption{Deciders on the direction of the next value, 150 M3 series at their last 12 origins (1,800 decisions, 52.2\% of rises). Accuracy and macro-F1 with 95\% intervals from a bootstrap over series, share of decisions that predict a rise, accuracy minus that of ETS on the same decisions, and accuracy of the 100 most confident decisions. Laya fine-tuned averages the probabilities of three seeds. In bold, the highest accuracy.}
\label{tab:m3-deciders}
\small
\begin{tabular}{lccccc}
\toprule
Decider & Accuracy & Macro-F1 & Rise & Minus ETS & Top 100 \\
\midrule
ETS & \textbf{0.673} $[0.643, 0.701]$ & 0.673 & 0.50 & -- & 0.88 \\
Jev & 0.540 $[0.514, 0.568]$ & 0.510 & 0.23 & $-0.133$ $[-0.168, -0.097]$ & 0.76 \\
Laya as released & 0.476 $[0.449, 0.503]$ & 0.324 & 0.00 & $-0.197$ $[-0.239, -0.154]$ & 0.56 \\
Laya fine-tuned & 0.522 $[0.495, 0.551]$ & 0.343 & 1.00 & $-0.151$ $[-0.187, -0.116]$ & 0.60 \\
\bottomrule

\end{tabular}
\end{table}

A typed decider does not read the direction of a series written as numbers (\Cref{tab:m3-deciders}). Laya collapses onto one answer, a fall or no change in 99.7\% of the decisions as released and a rise in every decision after fine-tuning, and its accuracy is that of the constant answer in both cases, 0.476 and 0.522. Jev answers a rise in 23\% of the decisions, and its accuracy, 0.540 with interval $[0.514,0.568]$, contains the 0.522 of always answering a rise. Exponential smoothing, ETS in the table, reaches 0.673 on the same decisions, and its 100 most confident decisions are right in 88\% of the cases. For a decision about numbers, the decider is a model of the series, and the gate goes on top of it.

\begin{table}[!htbp]
\centering
\caption{Gates on exponential smoothing over the 1,428 M3 series and the 24 test origins: share of origins whose accuracy among acted decisions is below the target, and decisions acted per 1,000. Sliding gates are calibrated on the 12 preceding origins. Per-bin gates accept a confidence bin when its predicted accuracy reaches the target. In bold, the lowest share below the target per column.}
\label{tab:m3-gates}
\small
\begin{tabular}{lcccc}
\toprule
 & \multicolumn{2}{c}{Target 0.70} & \multicolumn{2}{c}{Target 0.80} \\
\cmidrule(lr){2-3}\cmidrule(lr){4-5}
Gate & Below target & Per 1,000 & Below target & Per 1,000 \\
\midrule
Point, calibrated once & 0.42 & 795 & 0.29 & 350 \\
Learn then Test, calibrated once & 0.33 & 763 & 0.17 & 311 \\
Point, sliding & 0.50 & 828 & 0.42 & 398 \\
Beta rule, sliding & 0.42 & 792 & 0.38 & 372 \\
Learn then Test, sliding & 0.42 & 790 & 0.17 & 184 \\
Forgetting Beta per bin & \textbf{0.00} & 382 & \textbf{0.00} & 143 \\
Dynamic per bin & \textbf{0.00} & 365 & \textbf{0.00} & 166 \\
Dynamic per bin, lower bound & \textbf{0.00} & 282 & \textbf{0.00} & 147 \\
\bottomrule

\end{tabular}
\end{table}

At a target of 0.60 every gate keeps the promise in every origin, because the decider is right in 67\% of all decisions. At 0.70 and 0.80, threshold gates, whether calibrated once or slid over the preceding origins, fall below the target in 17\% to 50\% of the origins, while the per-bin gates never do (\Cref{tab:m3-gates}). A threshold calibrated on pooled origins promises the accuracy of their average, and the accuracy of the next origin varies around it. The per-bin gates pay in decisions, acting on 282 to 382 per 1,000 against 763 to 828 for the threshold gates at a target of 0.70, and part of their record comes from asking every accepted bin to reach the target, which by part (i) of \Cref{prop:dyngate} makes them conservative, so the two families are two operating points rather than one method better than the other. The seasonal pair of the dynamic model adds nothing measurable over the forgetting Beta here. The picture holds in the 145 financial series, where exponential smoothing is right in 65\% of the decisions and, at a target of 0.70, the sliding Learn then Test gate falls below the target in 29\% of the origins with 622 decisions per 1,000, against 4\% for the per-bin gates with 266 to 294. The selective ceiling leaves room for a demanding promise: the 1,000 most confident of the 34,272 test decisions are right in 97\% of the cases.

For a caller, a promise made for each period is kept only by a gate that checks each period. A threshold calibrated on a pooled past keeps the average and breaks in a third to a half of the periods at a demanding target, while tracking the accuracy of each confidence bin keeps it at the price of acting on fewer decisions.

%% ============================================================================
\section{When an explanation of a typed decision is useful}\label{sec:explain}

The deletion curves of \Cref{sec:expl} ask whether one ranking of words points to what a decision rests on. A caller who acts on an explanation needs more than a ranking. It needs to know which words carry the decision, whether they carry it alone, together or interchangeably, and how far that statement can be trusted when the decider returns probabilities rounded to two decimals, as Jev does. The problem is practical. Before acting on a decision, escalating it or auditing it, the caller wants to know whether a small change to the input could overturn it, and which words that change would touch. This section treats the explanation of one decision as a cooperative game over the words of the state and asks, for each kind of explanation, under which conditions it answers that question. Two conditions of use matter most, access to unrounded probabilities, as with open weights, and a closed API that returns two decimals. Each claim was measured against a criterion frozen before any report item was read, and the criterion decides what is reported, while its use for the caller is stated at the end of each part. Every implementation choice was recorded before the run it affects, and the changes made after a failure are stated where they occur.

\subsection{Probability is not explanation}\label{sec:expl}

\begin{figure}[!htbp]
\centering
\includegraphics[width=\linewidth]{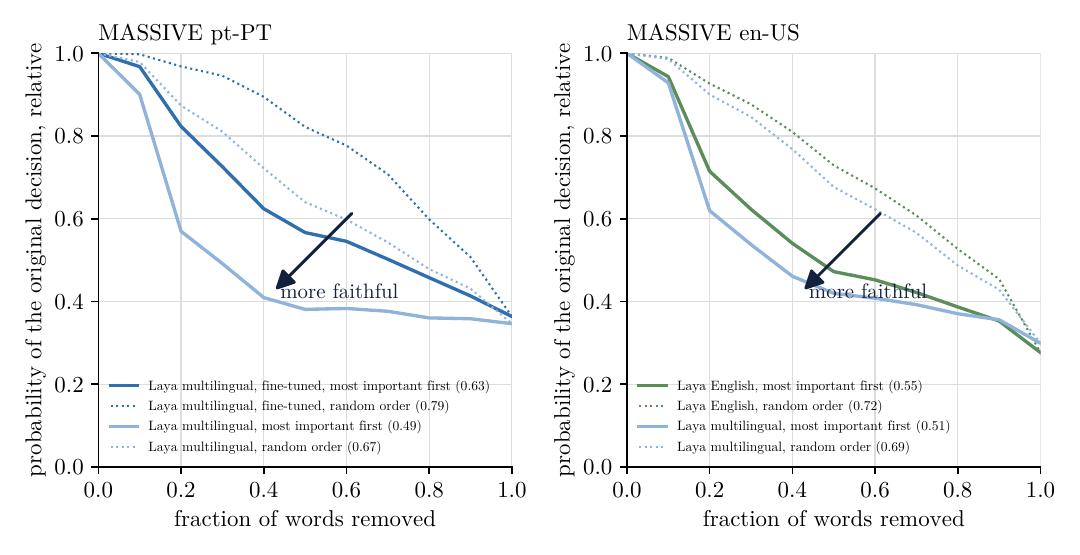}
\caption{Deletion curves by occlusion, mean over the 1,000 reported utterances: words removed most important first (solid) against random order (dotted), with areas in parentheses. A faster fall means a more faithful explanation.}
\label{fig:deletion}
\end{figure}

\begin{table}[!htbp]
\centering
\caption{Area under the deletion curve on the 1,000 reported utterances, words removed most important first (MoRF) and in random order, and the Spearman correlation between the probability of the decision $p_0$ and its MoRF area. The gap is random minus MoRF area: a larger gap means the ranking points to the words the decision rests on, beyond what any removal does. The largest gap per language is in bold.}
\label{tab:deletion}
\small
\begin{tabular}{llcccc}
\toprule
Lang. & Checkpoint & AUD, MoRF & AUD, random & gap & Spearman$(p_0,\text{AUD})$ \\
\midrule
pt & Laya multilingual, FT & 0.631 & 0.790 & 0.159 & $+0.25$ \\
pt & Laya multilingual & 0.490 & 0.675 & \textbf{0.185} & $-0.38$ \\
\midrule
en & Laya multilingual & 0.514 & 0.693 & \textbf{0.179} & $-0.13$ \\
en & Laya English & 0.555 & 0.723 & 0.169 & $+0.09$ \\
\bottomrule

\end{tabular}
\end{table}

Removing the words that occlusion ranks first makes the decision fall faster than removing words at random in all four configurations measured (\Cref{fig:deletion,tab:deletion}). The explanation therefore carries information on average. The decision-level question is different. The Spearman correlation between the probability of a decision and the area of its deletion curve is $+0.25$ for the fine-tuned multilingual checkpoint and $-0.38$ for the same checkpoint without fine-tuning, in Portuguese, and $-0.13$ and $+0.09$ for the multilingual and English checkpoints in English: weak to moderate, and of varying sign. A smaller area means a more faithful explanation, so a negative correlation means that more confident decisions have more faithful occlusion explanations. Two Portuguese utterances illustrate it with the fine-tuned checkpoint. ``qual é a diferença horária entre a califórnia e nova york'' is decided as \emph{datetime\_convert} with probability 0.997, and the word ``horária'' alone accounts for 0.995 of it (area 0.05): the decision rests on one word, and changing that word changes it. ``me notificar sobre os eventos no meu calendário'' is decided as \emph{calendar\_set} with probability 0.9999, and no single word changes the decision (area 0.998): the evidence is redundant and occlusion does not reveal it. The probability does not distinguish the two situations. Jev cannot be measured this way from frozen answers, since through the API each utterance would require one request per perturbation.

\subsection{The decision as a coalition game}

The units of an utterance are its words when it has at most eight, and otherwise eight contiguous spans of nearly equal length, so that $d\le8$ units give $2^d$ coalitions. For a coalition $S$, $x_S$ keeps the units in $S$ and deletes the others. With $j$ the option decided on the full utterance and $k^*$ the runner-up, the value of $S$ is the log-odds $y(S)=\operatorname{logit}p_j(x_S)$ or, when two options are compared, $y_{jk^*}(S)=\log p_j(x_S)-\log p_{k^*}(x_S)$. The game is enumerated exactly, and every attribution below is a function of it \citep{robnik2008,zintgraf2017,covert2021removing}:
\begin{equation}
  m_T=\sum_{U\subseteq T}(-1)^{|T|-|U|}\,y(U),\qquad F_i=m_{\{i\}},\qquad W_i=y(N)-y(N\setminus\{i\}),
  \label{eq:mobius}
\end{equation}
where $m_T$ are the Möbius terms \citep{fumagalli2024}, $F_i$ is the effect of a unit added to the empty state, $W_i$ is the occlusion effect of \Cref{sec:expl} in log-odds, and $E_i$ is the exact Shapley value \citep{shapley1953,lundberg2017,sundararajan2020}. Deleting a unit is one choice of removal among several \citep{covert2021removing}, so a second operator, which replaces each removed word by a training word unrelated to the options, is used as a check.

A method that ranks the units defines an order $\pi$. Its deletion area is $\sum_{k=1}^{d}[y(N)-y(N\setminus\pi_{1..k})]$, and the statistic $u_i$ is the percentile of that area among all $d!$ orders of the same item, with ties counted as one half. A value $u_i=1$ means that no order makes the decision fall faster. Unlike the area of \Cref{eq:deletion}, $u_i$ is not normalised by the initial probability and compares each item only with itself.

The four Laya cases are the multilingual and typed-decisions checkpoints in Portuguese and English, on the 1,000 report items of each language. Jev was enumerated in versions 1.13 and 1.14 on the same items, with all coalitions of an item packed in one request at a total cost of US\$2.51, at the price of US\$0.042 per million input tokens with output tokens free of charge, as stated by the vendor in October 2026. A stability check comes first. Asked the same question three times, Jev returns the same decision in 99.3\% of the items, but only 94.7\% (version 1.13 against the frozen answers) and 94.8\% (1.14, between repetitions) of its probabilities stay within 0.01, below the preregistered 95\%. The Jev analyses therefore use fresh answers only. Most Jev games are censored: the full utterance is returned with probability at least 0.995 in 71\% to 79\% of the items, depending on version and language.

\paragraph{What the empty state already decides.} A typed decider returns a distribution even for an empty state, and the coalition game records it as $y(\varnothing)$. The decision on the empty state measures how much of a decision the model would take without reading the input. Against a chance level of 0.20 from the option positions, the decision on the empty state agrees with the decision on the full utterance in 21\% to 23\% of the items for the multilingual checkpoint, 27\% to 28\% for the typed-decisions checkpoint and 27\% to 29\% for Jev, and in 41.9\% (95\% interval $[38.8,45.0]$) for the multilingual checkpoint fine-tuned in Portuguese. A different quantity takes almost the same value: that checkpoint answers 42\% of the items correctly with no input at all, against 18\% for the same checkpoint before fine-tuning. Fine-tuning thus nearly doubled the share of the decision that the options alone carry. The share follows the frequency of the intents in the training data. On the empty state, the fine-tuned checkpoint picks the most frequent training intent among the five options in 65.7\% of the items (95\% interval $[62.8,68.6]$, against 20\% without a frequency preference), and the Spearman correlation between its empty-state probabilities and the log frequency of the intents is 0.59. Before fine-tuning the same checkpoint picks it in 14.7\% of the items, with a correlation of $-0.20$. The pattern is consistent with fine-tuning having installed the label frequencies of the training set as a prior that the decision carries before reading any word, a reading based on one fine-tune and a correlation of 0.59. The empty state is the same in both languages, so the figures do not depend on the locale. For a caller, the empty-state probe is a one-request audit of how much of a decision is structural rather than read from the input.

\paragraph{Pairwise terms suffice for the game.} If the game were dominated by interactions of order three or more, no pairwise account of a decision could be faithful. The least-squares projection of $y$ on the constant, the units and their pairs explains a median share of 0.86 to 0.93 of the variance over the $2^d$ coalitions in the four Laya cases (\Cref{tab:explain-gates}), above the criterion of 0.6. Truncating the Möbius expansion at order two is a different and worse approximation, with negative median shares, so the order-two model is fitted, not read off the expansion. For inputs of up to eight units, an account of a decision by single words and pairs is therefore adequate. The exact game costs at most 256 passes of an open-weight decider and, with the coalitions of an item packed in one request, about US\$0.0006 per decision of Jev. The exact game is also the cheaper route in model passes. It needs a median of 64 passes per decision on these items, against 514 for KernelSHAP with 256 pairs of coalitions, and the exact Shapley value then takes about 0.1 ms on one CPU thread. The Gibbs sampler of the Bayesian model of the next subsection adds 60 to 83 ms per item.

\begin{table}[!htbp]
\centering
\caption{Explanation front on the report items: what was asked, what was measured against the preregistered criterion, and when the answer is useful to a caller. Coverage is the share of exact values inside the 90\% interval. Pt-ft is the fine-tuned multilingual checkpoint rounded to two decimals as a stand-in for Jev. Where a row compares methods, the best one is in bold. HEDGE is the hierarchical explanation method of \citet{chen2020hedge}. G1 to G5 are the criteria of the preregistration.}
\label{tab:explain-gates}
\small
\begin{tabularx}{\linewidth}{@{}p{0.2\linewidth}XX@{}}
\toprule
Question & Measured (criterion) & Useful when \\
\midrule
Are words and pairs enough? & median order-two share 0.86 to 0.93 (passes, $\ge0.6$) & always, for inputs of up to eight units \\
Intervals for the effect of a word under rounding & coverage of $W_i$ 0.55 to 0.59 and of $E_i$ \textbf{0.64 to 0.72}, KernelSHAP 0.59 to 0.63 (fails, $L\ge0.85$) & not available through a two-decimal API, where rounding bounds alone certify a relevant word in 60\% to 70\% of Jev decisions and the most relevant one in 17\% to 21\%, and not needed with unrounded probabilities and short inputs \\
Can redundant evidence be detected? & F1 0.300 for the Bayesian model, \textbf{0.901 for the Shapley interaction index}, 0.889 for HEDGE (fails for the Bayesian model) & with an interaction index or HEDGE, also through the API \\
Does confidence signal a faithful explanation? & positive slope in 4 of 4 Laya cases, small, Jev 1 of 4, planted intents at ranks 2, 3, 6 of 56 (passes) & no, faithfulness has to be measured per decision \\
Does the explanation depend on the options? & approximate invariance in 3 of 4 Laya and 8 of 8 Jev comparisons, 68\% to 79\% of items change decision or runner-up & only for the option set it was computed on \\
Which ranking deletes fastest? & \textbf{exact Shapley and HEDGE} first, occlusion last, in Laya and in Jev (Kendall 0.6 to 1.0) & exact Shapley or HEDGE, not occlusion \\
Does an additive head explain itself? & loss of 28.7 (pt) and 31.8 (en) points, word scores at 0.024 against the null 0.035 (fails) & no: the penalised residual ($\lambda=1$, 3 seeds) loses 3.7 to 3.8 points, calibrates worse and does not explain itself by word either, so the exact game of the original head is preferable for short inputs \\
\bottomrule
\end{tabularx}
\end{table}

\subsection{Credible intervals under rounding}

The order-two model of a game is $y(S)=\beta_0+\sum_{i\in S}\beta_i+\sum_{i<k\in S}\gamma_{ik}+\varepsilon$, with normal priors whose scales $\tau_1,\tau_2$ are chosen on calibration items by marginal likelihood. A probability rounded to two decimals only places $y(S)$ in an interval, unbounded when it reads 1.00, so the likelihood is interval-censored and the posterior is sampled by Gibbs with truncated-normal augmentation. The posterior gives 90\% intervals for $W_i=\beta_i+\sum_k\gamma_{ik}$ and $E_i=\beta_i+\tfrac12\sum_k\gamma_{ik}$. Because the exact values of Jev are unknown, the criterion is applied to Laya with its probabilities rounded as Jev's are, and the coverage of one unit per item is scored with the ascending Beta rule. The interval should contain the exact value with a lower bound $L\ge0.85$.

It does not (\Cref{tab:explain-gates}). The intervals of $W_i$ cover 58.8\% of the exact values in Portuguese and 55.4\% in English, and those of $E_i$ 63.8\% and 71.8\%. The sampler is correct: with the truth drawn from the prior, it covers 90.3\%, and 88.3\% with half of the coalitions censored at 1.00. With an exact order-two truth fitted to calibration games, coverage falls to 81\%, and to 72\% to 74\% when the full utterance reads 1.00. The shortfall lies in what rounded data can identify. When a decision and most of its coalitions read 1.00, the data say that the log-odds exceed $\operatorname{logit}0.995$ and nothing about how far. In Portuguese, 84\% of the items have at least one unit whose removal leaves the rounded probability at 1.00.

The two published uncertainty-aware estimators of the Shapley value do no better on the same rounded games and the same event. The unbiased KernelSHAP of \citet{covertlee2021}, which covers 90.8\% and 89.6\% on unrounded calibration games with five and eight units, covers 59.2\% and 63.2\% of the exact $E_i$. BayesSHAP \citep{slack2021} covers 30.7\% and 33.3\% on its own probability scale and 24.2\% and 22.8\% in log-odds. Paired on the same items and units, the Bayesian intervals cover more often than KernelSHAP in English (71.6\% against 63.2\%, exact McNemar $p<0.001$), no difference is detected in Portuguese (63.5\% against 59.2\%, $p=0.051$), and they cover more often than BayesSHAP on either scale in both languages ($p<10^{-40}$). Under two-decimal rounding, none of the methods tested yields a calibrated interval for the effect of a word. By the preregistered reading, the Bayesian intervals are reported as descriptive, and Jev is reported with the bounds that rounding alone identifies, $W_i\in[a(N)-b(N\setminus\{i\}),\,b(N)-a(N\setminus\{i\})]$, where $[a(S),b(S)]$ is the log-odds interval of the rounded probability of coalition $S$. These bounds are often one-sided or absent. Across the two versions and languages, 58\% to 67\% of the units have no finite bound, because the decision and the utterance without the unit both read 1.00, and only 19\% to 25\% have both bounds. They still answer part of the caller's question. In 60\% to 70\% of the decisions at least one unit is certain to lower the log-odds of the decision when removed, while the unit that matters most is identified in only 17\% to 21\%. The version matters more for the explanation than for the decision. Versions 1.13 and 1.14 take the same decision on 98.6\% of the Portuguese and 99.1\% of the English items, but give disjoint identified sets for 13\% and 10\% of the units, so an explanation obtained through the API holds for the pinned version only. The posterior means of the order-two model, reported without a criterion as preregistered, give the same picture. The paired difference of $E_i$ between versions has a median of $-0.03$ (Portuguese) and $-0.07$ (English) with interquartile ranges of $[-0.29,0.19]$ and $[-0.38,0.17]$ in log-odds, its sign changes for 7\% of the units, and the unit with the largest effect changes in 8.6\% and 9.9\% of the items. For a caller of a closed API that rounds, the consequence is direct. None of these methods provides a usable uncertainty statement about the effect of a word, and the failure is not confined to censored decisions, since on the uncensored ones the intervals of $W_i$ still cover only 60\% to 68\%. A caller with unrounded probabilities and inputs of up to eight units does not need such intervals, because the exact game gives the effects themselves. A second regime, which fits the same model to 256 random coalitions of unrounded games, covers held-out coalitions (92.1\% and 89.4\%) but not $W_i$ (48.5\% and 58.5\%), and leaves the article as preregistered.

\subsection{Unique, redundant and synergic evidence}

The second utterance of \Cref{sec:expl}, whose decision survives the removal of any single word, is a case of redundant evidence. An explanation that distinguishes such cases from evidence carried by one word, or only by two words together, would tell the caller what kind of change can overturn a decision. The protocol of \citet{bastings2022} gives a planted truth. Five invented tokens are inserted into the Portuguese training data with forced labels: one token that decides an intent alone (unique), two interchangeable tokens for a second intent (redundant), and two tokens that decide a third intent only together (synergic). The multilingual checkpoint learns all three in three seeds, with a lower bound on accuracy between 0.968 and 0.994.

Learning a rule is not the same as encoding it in the shape the rule suggests. The Möbius signature of the planted type, for instance $m_{A_1A_2}\le-\tfrac12\min(m_{A_1},m_{A_2})$ for the redundant pair, holds in every redundant test item, but only in 35\% to 38\% of the unique items and 3\% to 6\% of the synergic ones. A unique token alone drives the probability of its intent to 1 in single precision, which leaves no room for the main effect to separate from its interactions, and in the synergic case each token alone already moves the log-odds by a median of 2.4, while the joint term is about 17. The planted truth is kept only for the redundant type.

On that type, the Bayesian model fails. Its macro F1 over the labels unique, redundant, synergic and none is 0.300, against 0.901 for the pairwise Shapley interaction index on the same rounded game, 0.889 for HEDGE \citep{chen2020hedge} and 0.773 for the sign of $W_i-F_i$. The exact unrounded Möbius terms reach 0.632. A plausible reason, not tested here, is that the interval-censored posterior leaves the pairwise terms of saturated coalitions too uncertain for any credible interval to exclude zero. Under rounding, the interaction indices that average over many coalitions recover the planted redundancy, and the Bayesian decomposition does not. For the caller's question, whether removing one word can overturn a decision, the useful tool is therefore an interaction index or HEDGE on the observed game, which works through the API as well. The planted test supports this use for redundant evidence, the case of the second utterance of \Cref{sec:expl}, and not for unique or synergic evidence.

\subsection{Confidence and the faithfulness of the Shapley order}

\Cref{sec:expl} found the correlation between a decision's probability and its occlusion area weak and of varying sign. The coalition game allows a sharper version of the question. A hierarchical model relates $u_i$ of the exact Shapley order to the log-odds of the decision,
\begin{equation}
  u_i\sim\mathcal N(\mu_i,\,s_i^2+\sigma^2),\qquad \mu_i=\alpha_{c(i)}+b_{c(i)}\operatorname{logit}p_{0i}+\eta\,d_i,
  \label{eq:g5}
\end{equation}
with intercepts and slopes partially pooled over the intent $c(i)$ of the decision, fitted by NUTS with four chains. The non-centred parameterisation was adopted after the centred fits failed to converge ($\hat R$ up to 1.51), and the sign of $b$ had already been seen when the change was made. For Jev, $u_i$ is averaged over 100 posterior draws of the order and $s_i$ is its standard deviation, and only the uncensored items enter.

In all four Laya cases the posterior probability of a positive slope is 1.00 to two decimals (\Cref{tab:explain-gates}). The effect is small: $b$ lies between 0.004 and 0.010 per unit of log-odds, so ten units of log-odds raise $u_i$ by 0.04 to 0.10 on a scale where its mean lies between 0.93 and 0.95, and the Spearman correlation between $p_0$ and $u_i$ is 0.20 to 0.25. In Jev's uncensored stratum, of 207 to 284 items, the sign is decided in one of four cases. The model is checked on the planted checkpoint: it places the three planted intents at ranks 2, 3 and 6 of 56 by $\alpha_c$, within the top 10\% required. For Laya, then, a more confident decision has a slightly more faithful Shapley order. The relation is too weak for the probability to select which explanations to trust. The measure here, the fidelity of the Shapley order, differs from the deletion area of \Cref{sec:expl}, and the two agree only in that the relation is weak, while its direction depends on the checkpoint and on the measure. A caller who needs a faithful explanation of a decision has to measure it for that decision, at the cost given above.

\subsection{Explanations across options and orders}

A typed decider receives its options from the caller, and an explanation of $j$ against $k^*$ could change when the other three options change. On 200 items per language, the three options other than $j$ and $k^*$ were replaced twice by other intents, and the set of units with the largest $W^{jk^*}_i$ was compared with the original. The rate of change was then compared with that of a placebo in which the original options are only permuted. The difference $D$ shows approximate invariance in three of the four comparisons and a dependence in one ($D=0.167$, 95\% interval $[0.024,0.310]$, Portuguese, second replacement). The evidence is thin. In about 75\% of the items the replacement changes the decision or the runner-up, so the item leaves the comparison, and the remaining 42 to 52 items per comparison give wide intervals. Jev, with the largest $W^{jk^*}_i$ taken from the posterior mean of the order-two model on its rounded difference game, shows approximate invariance in all eight comparisons of the two versions and languages, with 68\% to 76\% of the items excluded for the same reason. That exclusion rate is itself the finding that matters for use. A contrastive explanation holds for the option set it was computed on, and when the caller changes the options it has to be recomputed rather than reused. The same distinction between changing the options and only reordering them is a design principle of Jeva \citep{li2026jeva}. Its option representations are computed independently of the request and reused across requests, while the request is refined against the full candidate set, so that the decision is meant to depend on which options are offered and not on their order.

On the same items, six orders were scored by $u_i$ under both removal operators and for both deletion and insertion. The rankings of the methods agree across operators, with Kendall's $\tau$ between 0.6 and 1.0. Paired over items under deletion, the exact Shapley order deletes faster than the occlusion order $W_i$ of \Cref{sec:expl} in all six cases of Laya and Jev, by 0.05 to 0.23 in percentile, and faster than KernelSHAP in all six, by 0.007 to 0.013. Against BayesSHAP the difference is detected in three cases, against the first-entry order $F_i$ in five, and against HEDGE in none, so the exact Shapley value and HEDGE share the lead. Occlusion, the cheapest explanation, is among the two least faithful by its own criterion of deletion. The mean percentile of the Shapley order under deletion is 0.92 in Portuguese and 0.93 in English, so the best order is close to, but not at, the fastest possible. On Jev's rounded games the same ranking appears in all four cases, with Kendall's $\tau$ between 0.6 and 1.0, and occlusion falls further behind, with a mean percentile of 0.71 to 0.75 against 0.94 to 0.96 for the Shapley order. Rounding leaves the occlusion effect at exactly zero for most units, so its order is decided by the tie-break. When a single ranking of words is wanted for a short input, the exact Shapley order or HEDGE is the one to use, and occlusion, though the cheapest, is the one to avoid.

\subsection{An additive head}

A head built so that the contribution of each word is exact loses 28.7 points of accuracy in Portuguese and 31.8 in English, and the head that results is additive in positions rather than in words. Exactness by construction is therefore not a useful route to explanations here, and the measurement is reported in full in \Cref{app:additive}.

%% ============================================================================
\section{A validation protocol}\label{sec:validation}

Two kinds of validation are needed: of each statement made about a decider, and of each decision it takes. For statements, we tag every claim with its source: \emph{documentation} (what the vendor states, on the access date), \emph{code} (what the program does, at a pinned commit), \emph{measured} (what happened on our data, reproducible with the released script and seed), and \emph{not published} (nothing can be asserted). The tags matter in practice. Between 2026-09-29 and 2026-10-20 the vendor documentation changed from a text-only input to experimental image input and a new model version, 1.14, and a statement made without a date became false without anyone changing it. A vendor statement can also be tested, as the one about probability 1.0 was.

For decisions, we group the operational controls into seven, a grouping that the same talk also uses \citep{ibm2026jev}, and turn each into a measurement with the instruments of \Cref{sec:measures} (\Cref{tab:controls}). None of them requires access to the model's internals, so all apply equally to the open and the closed decider. Only the measurement of explanation fidelity does.

\begin{table}[!htbp]
\centering
\caption{Seven operational controls turned into measurements, and where the article measures each.}
\label{tab:controls}
\small
\begin{tabularx}{\linewidth}{@{}lXl@{}}
\toprule
Control & What to measure & Measured in \\
\midrule
Validation & a fixed schema for the state, a token count against the context budget, the language, and the format in which the content is sent, which is part of the input & \Cref{sec:noise,sec:language} \\
Signals & a second, independent decider and the McNemar table of disagreements & \Cref{sec:results} \\
Logging & state, question, options, model version and probabilities of every call & the released package \\
Drift & a fixed probe set re-run over time, with accuracy, ECE and McNemar between windows, and the accuracy of each confidence bin tracked online & \Cref{sec:drift} \\
Anomaly & shifts in option frequencies, cases with high $c_{\max}$ but low $c_H$, and the decision on the empty state & \Cref{sec:conf,sec:explain} \\
Escalation & a gate with a lower-bound rule, Learn then Test where a guarantee is required & \Cref{sec:gate} \\
Verification & an audit sample of the decisions taken alone & \Cref{sec:drift} \\
\bottomrule
\end{tabularx}
\end{table}

For comparisons between methods, the definitions of \Cref{sec:compare} turn a reading of means into a statement that another team can check. Four practices carry most of the weight. The margin is fixed before the data, from the smallest difference that matters or from a stated tolerance, and written down with its reason. Non-inferiority and equivalence are claimed only by their tests, and an interval that contains zero is reported as no difference detected. Resources are measured alongside the outcome, so that a method that keeps the promise at a lower cost is credited and a method that saves one resource by spending another is reported as a trade-off. And the uncertainty of every difference includes the sampling of the items, not only the Monte Carlo error of the procedure, since in the gate of \Cref{sec:gate} the Monte Carlo error alone would show a detectable coverage gain in 21 of the 24 combinations, against 4 once the item part is included.

%% ============================================================================
\section{What a caller can do}\label{sec:whattodo}

The sections above end with what their measurements mean for whoever deploys a typed decider. \Cref{tab:whattodo} gathers those readings in one place, each with the measurement behind it.

\begin{table}[!htbp]
\centering
\caption{What the measurements of the article say to a caller of a typed decider.}
\label{tab:whattodo}
\small
\begin{tabularx}{\linewidth}{@{}p{0.2\linewidth}XXl@{}}
\toprule
Situation & What was measured & What to do & Where \\
\midrule
Promising an accuracy among accepted decisions & the point estimate breaks the promise in 147 of 190 evaluations, Learn then Test keeps it in all, the ascending Beta rule in 183 with more decisions & Learn then Test where a guarantee is required, the ascending Beta rule where more automatic decisions matter more, never the point estimate & \Cref{sec:gate} \\
Few calibration labels & a threshold needs at least $\lceil\ln q/\ln\alpha\rceil-1$ accepted items to qualify, 58 at 95\% & collect at least that many accepted items, or let the gate abstain & \Cref{prop:nmin} \\
Noisier inputs than the calibration items & a few points of accuracy lost on typed text break the promise & calibrate on items that look like the traffic, noise included, and train with noise & \Cref{sec:noise} \\
Several languages & calibrated in English, Jev keeps the promise in 8 of 51 locales, with a slack of 1.2 points & calibrate in each language, and read the slack as the loss the promise can absorb & \Cref{sec:language} \\
A new domain & labels for non-inferiority to Jev range from 250 to 8,000, Jev's ECE from 0.076 to 0.171 & measure the label curve and the calibration in the domain, and test what a fine-tune reads & \Cref{sec:battery} \\
No rule decides anything & the selective ceiling lies below the target & add informative inputs to the state, ask a coarser question or answer with sets & \Cref{prop:ceiling} \\
Change over time & the promise of a fixed threshold does not survive a change that degrades the model, and no retraining arm keeps it & audit accepted decisions, re-qualify the threshold every window where reviews are affordable, and forget faster in per-bin gates. Add seasonal components only for a known cycle that degrades the model & \Cref{sec:drift,sec:drift-pattern} \\
Explaining a decision & the probability does not select faithful explanations, and the empty state reveals what the options alone decide & measure fidelity for the decision at hand, with the exact Shapley order or HEDGE for short inputs & \Cref{sec:explain} \\
Numeric decisions over time & on M3, Jev and Laya answer as a constant would, exponential smoothing is right in 67\% of the decisions, and threshold gates break the promise in 17\% to 50\% of the origins at targets of 0.70 and 0.80 & decide with a model of the series and gate each period, accepting fewer decisions & \Cref{sec:m3} \\
API or local model & a local fine-tune is non-inferior to Jev from 4,000 labels & weigh labels, privacy and auditability rather than the price per call & \Cref{app:cost} \\
\bottomrule
\end{tabularx}
\end{table}

%% ============================================================================
\section{Limitations}

Most measurements use one dataset, MASSIVE, with five options per question. The language front extends them to 52 locales without fine-tuning beyond Portuguese, and the battery to four further domains with two to six options, but each of those domains is one dataset. Only exact duplicates of test utterances were removed from the fine-tuning set. Near duplicates may remain and inflate the accuracy of the fine-tuned checkpoints. The deletion area is normalised by the initial probability and capped at one, so when that probability is close to one, as for the fine-tuned checkpoint, its range is restricted and the correlation with the probability is partly mechanical and attenuated. The fine-tuned checkpoints were trained in Portuguese only. Jev was measured mostly in version 1.13, with version 1.14 only in the explanation front, and its rounding to two decimals limits every analysis that needs its probabilities near one. Laya probabilities are stored with five decimals and clipped at $10^{-6}$ before temperature scaling. Occlusion measures one kind of explanation and produces inputs unlike the training data. The coalition games of \Cref{sec:explain} delete words and inherit that limitation, which the replacement operator only partly addresses. Their gates were evaluated on Laya rounded to two decimals as a stand-in for Jev, which assumes that rounding, not the model, drives the failure of coverage. The planted truth holds for the redundant type only. The noise of \Cref{sec:noise} is synthetic, drawn from a fixed set of operators, and the traffic of a real deployment may hold other kinds, such as abbreviations, mixed languages and voice transcription errors. The cost ratios 10:1 and 100:1 are illustrative scenarios. The gate is evaluated on random splits of one fixed sample, which makes calibration and report items exchangeable. Under drift or with correlated items, such as consecutive video frames, neither the Beta rule nor Learn then Test keeps its promise without further assumptions.

%% ============================================================================
\section{Conclusion}

A typed decider's ``0.97'' licenses a statement about groups: among many answers given with that confidence, about 97\% are right, provided the decider is calibrated on the data at hand. It does not certify a single answer and it does not explain the decision. The measurements here make each of these limits concrete, and the gate turns them into an engineering choice: separate calibration from decision, choose the threshold from the cost of errors, and state the promise as a conditional accuracy that the threshold rule must sustain on new data. A point estimate breaks such a promise in 147 of the 190 exchangeable evaluations measured here. A lower bound keeps it. Learn then Test does so with a guarantee, in all 190 exchangeable evaluations here. The ascending Beta rule, which differs from it only in the prior and the order of the scan, decides more items on average than Learn then Test in 128 of them and fewer in none, a gain the item-level uncertainty of the static front does not confirm after correction for multiplicity, at a violation rate that stays near the nominal level without a guarantee, demonstrably above 5\% in one evaluation, at 5.5\%. It is the better trade when more automatic decisions matter more than a formal guarantee, and it gains most with the weakest decider. Every counting rule depends on the order of the confidences rather than on their calibration, which a Bayes threshold does not escape. The same reading applies to explanations. For short inputs the exact coalition game of a decision is affordable, also through Jev's API, and it says which words carry the decision and whether they carry it interchangeably. Rounding to two decimals takes away the uncertainty statement. Through the API a caller can certify that some word matters in most decisions but rarely which matters most, and an explanation holds only for the pinned version and for the options it was computed on. Under drift the exchangeability behind every rule here fails, and the measurements again say when a method helps rather than whether it wins. The Bayesian loop avoids needless retraining where the stream is stable or its change harmless, at a cost equivalent to recalibration alone and lower than that of periodic retraining and DDM. Under a drift that degrades the model it reacts late and does not improve on periodic retraining, and the arms that keep the promise most often re-qualify the threshold in every window and pay for it in reviews. The labels that every loop consumes can also come from the people who receive the decisions, by marking each answer right, partly right or wrong, which turns the audit sample into online calibration data and is the setting we leave for future work. The ablation credits the loop's trigger and shows that its re-qualification step, the ascending Beta scan on hard recent labels, is where the promise breaks. On real news streams, where the change from year to year is mostly mild, the loop has the lowest mean cost in nine of ten transitions, five of which do not pass the preregistered check of degradation, and the preregistered summary over the five that do detects no difference.

A promise also does not travel by itself. Calibrated in English, Jev keeps a promise of 95\% in 8 of 51 locales, a fine-tuned Laya needs from 250 to 8,000 labels to match Jev depending on the domain, and on the direction of a numeric series neither decider does better than a constant answer, while exponential smoothing is right in 67\% of the decisions. The rule for a caller follows from all of this: calibrate on data that looks like the traffic, choose the threshold with a lower bound, check the promise again when the traffic changes, and audit the decisions the gate takes alone.

\section*{Data and code availability}
The reference package is deposited on Zenodo \citep{brom2026package}, \href{https://doi.org/10.5281/zenodo.23676316}{doi:10.5281/zenodo.23676316}: the parallel items in the 52 locales, Jev's decisions with the probability of the decided option, Laya's probabilities for every option, the coalition games, the drift runs and a datasheet. Jev's probabilities for the options not decided are not distributed, since the vendor's customer agreement forbids using its output to train a model that imitates it. The code, including \texttt{promise\_check} and its test, and the derived measurements and figures are available at \url{https://github.com/pcbrom/typed-deciders-audit} and in the same deposit. The fine-tuning recipe and the lesson material are at \url{https://github.com/pcbrom/laya-jev}.

\section*{Acknowledgments}
The authors thank the Laborat\'{o}rio de Processamento e Reconhecimento Inteligente de Sistemas e M\'{e}todos Anal\'{\i}ticos (L-PRISMA) of the Federal Institute of Bras\'{\i}lia, registered in the CNPq Directory of Research Groups, and the Laborat\'{o}rio de Modelo Computacional e Intelig\^{e}ncia para Transporte (TransLab) of the Department of Computer Science, University of Bras\'{\i}lia.

\section*{Use of AI-assisted tools}
AI-assisted tools were used for code development and English language editing. Experimental design, analysis and interpretation are the authors'.

\bibliographystyle{unsrtnat}
\bibliography{references}

\appendix
\section{Proofs}\label{app:proofs}

\paragraph{Proof of \Cref{prop:affine}.} For $k\ge2$, $c_{\text{J}}=\frac{k}{k-1}c_{\max}-\frac{1}{k-1}$ with slope $k/(k-1)>0$. Hence $c_{\max}\ge\tau \iff c_{\text{J}}\ge(k\tau-1)/(k-1)$, and both thresholds accept the same items. For $k=2$, $\mathbf{p}=(c_{\max},1-c_{\max})$ with $c_{\max}\in[1/2,1]$, and the binary entropy is strictly decreasing on that interval, so $c_H$ is strictly increasing in $c_{\max}$. For $k=3$, take $\mathbf{p}=(0.6,0.2,0.2)$ and $\mathbf{p}'=(0.59,0.41,0)$: $c_{\max}(\mathbf{p})=0.60>0.59=c_{\max}(\mathbf{p}')$, while $c_H(\mathbf{p})=0.135<0.384=c_H(\mathbf{p}')$. A threshold on $c_H$ between the two values accepts $\mathbf{p}'$ and rejects $\mathbf{p}$, and every threshold on $c_{\max}$ that accepts $\mathbf{p}'$ also accepts $\mathbf{p}$. Appending zero probabilities extends the pair to every $k>3$. $\square$

\paragraph{Proof of \Cref{prop:convex}.} Write $\ell_i(\beta)=-\beta z_{i,y_i}+\log\sum_j e^{\beta z_{ij}}$. Differentiating, $\ell_i'(\beta)=\sum_j p_{ij}(\beta)z_{ij}-z_{i,y_i}$ and $\ell_i''(\beta)=\sum_j p_{ij}(\beta)z_{ij}^2-\big(\sum_j p_{ij}(\beta)z_{ij}\big)^2=\mathrm{Var}_{\mathbf{p}_i(\beta)}[z_i]\ge0$, using $\frac{d}{d\beta}p_{ij}(\beta)=p_{ij}(\beta)\big(z_{ij}-\sum_l p_{il}(\beta)z_{il}\big)$. The variance is positive whenever $z_i$ is not constant, since $\mathbf{p}_i(\beta)$ has full support. Averaging over $i$ gives the statements on $\mathcal{L}$. If $z_{i,y_i}>\max_{j\neq y_i}z_{ij}$ for every $i$, then $\sum_j p_{ij}(\beta)z_{ij}<z_{i,y_i}$ for every finite $\beta$, because the expectation of $z_i$ under a full-support distribution is strictly below its maximum, hence $\mathcal{L}'(\beta)<0$ everywhere and no minimiser exists. $\square$

\paragraph{Proof of \Cref{prop:bayes}.} A rule that decides from $C$ accepts on a set $S\subseteq[0,1]$. Its expected cost is $\E\big[\1[C\in S]\,C_e\,(1-A)+\1[C\notin S]\,C_r\big]=\E\big[\1[C\in S]\,C_e\,(1-\Prob(A=1\mid C))+\1[C\notin S]\,C_r\big]$. Under top-label calibration $\Prob(A=1\mid C=c)=c$, so the integrand at $C=c$ is $C_e(1-c)$ if $c\in S$ and $C_r$ otherwise. The pointwise minimum is attained by including $c$ in $S$ exactly when $C_e(1-c)<C_r$, that is $c>1-C_r/C_e$. Ties may be resolved either way. Pointwise minimisation of the integrand minimises the expectation. $\square$

\paragraph{The Beta posterior of \cref{eq:beta}.} With $\theta\sim\mathrm{Uniform}(0,1)=\mathrm{Beta}(1,1)$ and $c$ successes in $n$ exchangeable Bernoulli($\theta$) trials, the posterior density is proportional to $\theta^{c}(1-\theta)^{n-c}$, which is $\mathrm{Beta}(c+1,n-c+1)$. Its predictive mean $(c+1)/(n+2)$ is Laplace's rule of succession, used per confidence bin in the gating rule under drift.

\paragraph{Proof of \Cref{prop:scan}.} (i) If the qualifying indices are $\{1,\dots,k\}$, both maxima equal $k$, and both sets are empty when $k=0$. Conversely, if $J_A=J_F=k\ge1$, then $Q_1=\dots=Q_k=1$ by the definition of $J_F$ and $Q_j=0$ for $j>k$ by the definition of $J_A$, so the qualifying set is $\{1,\dots,k\}$. If both scans return nothing, no index qualifies. If $J_F$ is empty and $J_A$ is not, the scans disagree, and some $Q_j=1$ follows $Q_1=0$. (ii) On $D^c$ the two scans return the same threshold, or none, so $V_A\cap D^c=V_F\cap D^c\subseteq V_F$, and $V_A=(V_A\cap D)\cup(V_A\cap D^c)\subseteq D\cup V_F$. The inequalities follow from monotonicity and subadditivity of $\Prob$. (iii) By \Cref{rem:ltt}, with the prior $\mathrm{Beta}(0,1)$ and $q=\delta$, $Q_j=1$ if and only if $p_j\le\delta$, and a threshold is testable for the Beta rule, $\delta^{1/n_j}\ge\alpha$, if and only if $\alpha^{n_j}\le\delta$, the testability condition of Learn then Test. The fixed-sequence scan over testable thresholds is then the procedure of Learn then Test, whose guarantee gives $\Prob(V_F)\le\delta$, and (ii) completes the bound. $\square$

\paragraph{Proof of \Cref{prop:nmin}.} For fixed $n$, the ratio of the densities of $\mathrm{Beta}(c+2,n-c)$ and $\mathrm{Beta}(c+1,n-c+1)$ is proportional to $\theta/(1-\theta)$, which is increasing, so the family is ordered by likelihood ratio in $c$ and hence stochastically increasing in $c$. Every quantile is therefore largest at $c=n$, where the posterior is $\mathrm{Beta}(n+1,1)$ with distribution function $x^{n+1}$ and $q$-quantile $q^{1/(n+1)}$. The rule qualifies only if $q^{1/(n+1)}\ge\alpha$, that is $\ln q\ge(n+1)\ln\alpha$, and since $\ln\alpha<0$ this is $n+1\ge\ln q/\ln\alpha$. As $n$ is an integer, $n\ge\lceil\ln q/\ln\alpha\rceil-1$. $\square$

\paragraph{Proof of \Cref{prop:ceiling}.} (i) By the tower property, $\Prob(d(X)=Y,\,X\in S)=\E\big[\1[X\in S]\,\eta_{d(X)}(X)\big]\le\E\big[\1[X\in S]\,m(X)\big]$, since $\eta_y(x)\le m(x)$ for every $y$. Dividing by $\Prob(X\in S)$ gives the claim. (ii) The decision $d^*$ attains equality in (i) on every set. Let $\Prob(X\in S)=c=\Prob(X\in S^*)$. Then $\Prob(X\in S\setminus S^*)=\Prob(X\in S^*\setminus S)$, $m<t$ on $S\setminus S^*$ and $m\ge t$ on $S^*\setminus S$, so
\[
  \begin{aligned}
  \E\big[m(X)\1[X\in S]\big]-\E\big[m(X)\1[X\in S^*]\big]&=\E\big[m(X)\1[X\in S\setminus S^*]\big]-\E\big[m(X)\1[X\in S^*\setminus S]\big]\\
  &\le t\,\Prob(X\in S\setminus S^*)-t\,\Prob(X\in S^*\setminus S)=0,
  \end{aligned}
\]
and by (i) no pair with coverage $c$ exceeds $\E[m(X)\mid X\in S^*]=\Theta^*(c)$. For $c_1<c_2$ with levels $t_1\ge t_2$, the set $S^*_2$ is the union of $S^*_1$ and $\{t_2\le m<t_1\}$, so $\Theta^*(c_2)$ is a weighted mean of $\Theta^*(c_1)$ and of a conditional mean of $m$ that is below $t_1\le\Theta^*(c_1)$, hence $\Theta^*(c_2)\le\Theta^*(c_1)$. (iii) If $\alpha$ exceeds the essential supremum $M$ of $m(X)$, then $\E[m(X)\mid X\in S]\le M<\alpha$ for every $S$ of positive probability, and (i) applies. (iv) Let $\zeta(x,w)$ be the vector of probabilities $\zeta_y(x,w)=\Prob(Y=y\mid X=x,W=w)$ and $m_{X,W}=\max_y\zeta_y$. The posterior given the input alone is the conditional mean of the finer one, $\eta(X)=\E[\zeta(X,W)\mid X]$, and $v\mapsto\max_y v_y$ is convex as a maximum of linear functions, so the conditional Jensen inequality gives $m(X)=\max_y\E[\zeta_y(X,W)\mid X]\le\E[m_{X,W}(X,W)\mid X]$. The set $S^*$ of coverage $c$ for $X$ depends on $X$ only, so the decision $\argmax_y\zeta_y$ restricted to it has accuracy $\E[m_{X,W}\mid X\in S^*]\ge\E[m(X)\mid X\in S^*]=\Theta^*_X(c)$, and (ii) applied to $(X,W)$ gives $\Theta^*_{X,W}(c)\ge\Theta^*_X(c)$. For merged options, the probability of a merged option is the sum of the probabilities of its members, which is at least the probability of each member. $\square$

\paragraph{Proof of \Cref{prop:dyngate}.} (i) $\Theta_t$ is a weighted mean of the $\theta_{b,t}$ over $A_t$ with nonnegative weights, so it is at least their minimum, which is at least $\alpha$. (ii) Given the labels up to $t-1$ and the inputs up to $t$, the set $A_t$ is fixed. Every $b\in A_t$ has $\Prob(\theta_{b,t}<\alpha\mid\cdot)\le q/B$ by the gating rule, so by the union bound $\Prob(\exists\,b\in A_t:\theta_{b,t}<\alpha\mid\cdot)\le|A_t|\,q/B\le q$, and on the complement every accepted bin reaches $\alpha$, so $\Theta_t\ge\alpha$ by (i). (iii) With the level alone, $F_t=1$ and $G=1$, so the prior moments of the logit are $f=m_{t-1}$ and $q=C_{t-1}/\lambda$, and the update gives $m_t=f^*$ and $C_t=R_t-R_t(1-q^*/q)=q^*$, because $R_t=q$. The state therefore holds the moments of the logit of $\mathrm{Beta}(r+y,s+n-y)$, and evolution divides its variance by $\lambda$ while keeping its mean. Since $\psi(x)=\ln x+O(1/x)$ and $\psi'(x)=1/x+O(1/x^2)$, \cref{eq:betamoments} reads $\ln(r/s)=f$ and $1/r+1/s=q$ to first order, and dividing $q$ by $\lambda$ with $f$ fixed is solved by $(\lambda r,\lambda s)$. One window maps $(r,s)$ to $(\lambda r,\lambda s)$ by evolution and then to $(\lambda r+y,\lambda s+n-y)$ by updating. For the remark after the proposition, the predictive distribution is a function of the past labels and of $x_t$, so part (iv) of \Cref{prop:ceiling} applies with the past and $x_t$ as the observed variables. $\square$

\section{What each route costs}\label{app:cost}

This appendix holds the cost comparison summarised in \Cref{sec:results}: the measured costs of each route, the break-even between a local fine-tune and a priced API, and the effect of the processor's instruction set on local inference.

\begin{figure}[!htbp]
\centering
\includegraphics[width=0.66\linewidth]{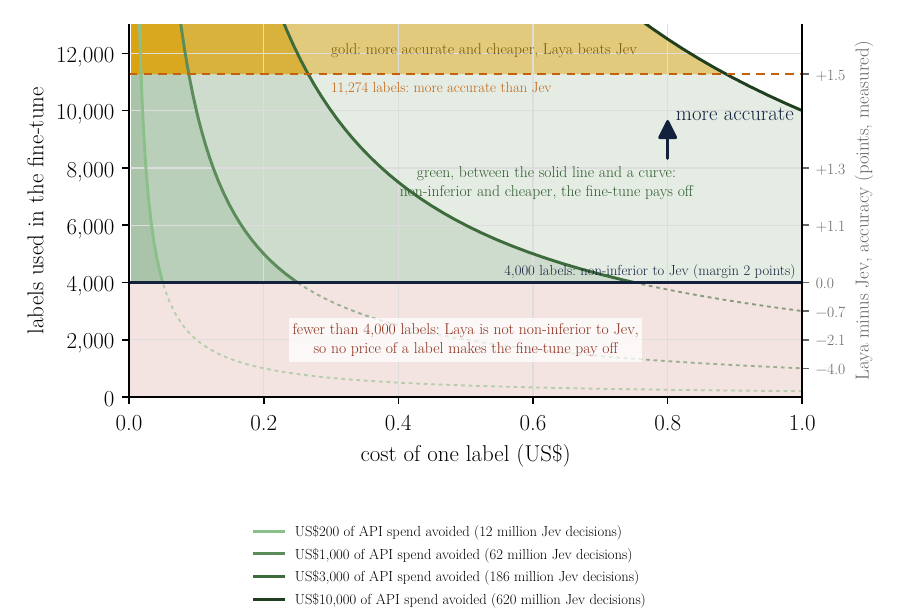}
\caption{When fine-tuning Laya pays off, for any priced API. Two conditions must hold. The fine-tune must be non-inferior to Jev in accuracy, which the paired test with a margin of 2 points establishes from 4,000 labels on (solid line). Below it, in the red band, no price of a label makes the fine-tune worthwhile. And its labels must cost less than the API spend they replace, $L\,c_\ell\le B$ with $B=N\,p$ the cost of the $N$ decisions served at a price $p$ per decision. For a fixed $B$ this boundary is a hyperbola, so the number of labels and the price of a label trade off non-linearly. The fine-tune pays off in green, between the solid line and a hyperbola, and the green deepens where it pays off already at a smaller API spend. The right axis gives, at the label counts of the test grid only, the measured difference in accuracy between the fine-tuned Laya and Jev, averaged over three seeds. The curves are labelled with the API spend avoided and, in parentheses, the number of Jev decisions it buys at the measured price. The dashed line marks the 11,274 labels at which Laya becomes more accurate than Jev, and in gold, above it and below a hyperbola, Laya beats Jev on both counts, more accurate and cheaper. Superiority was measured at 11,274 labels, the whole training pool, and the gold band above that count assumes that more labels do not lower the accuracy, which was not tested. Each hyperbola is a lower bound on the volume needed, because local inference is not free, and it does not price that accuracy gain, which would enlarge the region wherever an error has a cost.}
\label{fig:breakeven}
\end{figure}

\paragraph{What each route costs.} \Cref{tab:cost} puts the measured costs side by side. Jev needs no labels and no training, and a decision costs 384 input tokens at the price stated by the vendor, US\$0.016 per 1,000 decisions. Its latency was not measured. Laya runs locally in about 12 ms per item on a CPU. Fine-tuned on subsets of the training partition with three seeds and compared item by item with Jev's frozen answers, it is non-inferior to Jev, with the 2-point margin of \Cref{sec:compare}, from 4,000 labels on (paired difference 0.0 points, 95\% interval $[-1.5,1.5]$), and more accurate with the full 11,274 (1.5 points, $[0.1,2.9]$). At 3,500 labels the lower bound is $-2.1$, just outside the margin. Training on 4,000 labels takes 17 seconds on one GPU. The fixed cost of the local route is the labelling, which was not measured and depends on the domain. With $c_\ell$ the cost of a label, $L$ the number of labels and $p$ Jev's price per decision, the local model pays for its labels after
\begin{equation}
  N^*=\frac{L\,c_\ell}{p}
  \label{eq:breakeven}
\end{equation}
decisions. $N^*$ is a lower bound, since local inference is not free either. At an illustrative $c_\ell$ of US\$0.05 and $L=4{,}000$, $N^*$ is 12 million decisions, and at US\$0.50 it is 124 million. Unless the volume reaches tens of millions of decisions, the price per call therefore barely enters the choice between the two models. What decides it is whether the state may leave the machine, whether each decision must be auditable, and whether the accuracy required is beyond what the closed model reaches without labels, as it is for the fine-tuned checkpoint here. Rewritten as $L\,c_\ell\le N\,p$, the condition is a product, and \Cref{fig:breakeven} draws its boundary for any price of a label, any number of labels and any API price.

\paragraph{The processor sets part of the local cost.} The 12 ms of \Cref{tab:cost} use bfloat16 autocast on a CPU with AVX-512 BF16 instructions. On that processor (AMD Ryzen 9 9900X3D, six threads, batches of eight sorted by length), the released multilingual checkpoint takes 30.0 ms per item in Portuguese and 29.7 in English in float32, 14.1 and 13.8 ms in bfloat16 autocast and 11.0 and 10.8 ms with the encoder quantised to int8 on the fly. Against float32 on the same items, bfloat16 is equivalent by two one-sided tests with the 2-point margin (Portuguese $-0.3$ points, 95\% interval $[-0.7,0.0]$, English $+0.1$, $[-0.2,0.5]$) and keeps the decision in 99.7\% and 99.6\% of the items, a clean gain of 2.1 times in speed. Dynamic int8 is 2.7 times faster but loses 4.2 points in Portuguese ($[-6.9,-1.5]$) and 2.9 in English ($[-5.2,-0.7]$), changes 25.6\% and 17.2\% of the decisions and is not non-inferior, so its speed is bought with accuracy. On a processor with the bfloat16 instructions the faster precision is therefore free, and the cheaper integer one is a trade-off that has to be measured on the task before it is used.

\section{An additive head}\label{app:additive}

The additive head summarised in \Cref{sec:explain} is reported here in full.

An explanation that is exact by construction requires a model whose decision is a sum over words \citep{agarwal2021}. Attention layers do not learn additive models on their own \citep{leemann2024}, so the additive form has to be imposed. In the additive head, each state token $t$ contributes $e_{t,j}=\operatorname{LN}(h_t)^\top UV^\top\operatorname{LN}(\tilde m_j)$ to the logit of option $j$, with $U,V$ of rank 64 and $\tilde m_j$ the representation of option $j$ with an empty state. The logit is a bias plus the sum of these contributions, so the state enters only through the sum. The head was fine-tuned with the recipe and seeds of the original head.

Over three seeds, the additive head reaches an accuracy between 0.614 and 0.749 in Portuguese and between 0.589 and 0.713 in English, against 0.960 to 0.972 for the original head trained the same way. The paired loss is 28.7 points in Portuguese (95\% interval $[-30.8,-26.6]$) and 31.8 in English ($[-34.1,-29.7]$). With seed 0 its training accuracy is 0.692, so the loss is underfitting rather than overfitting. Its calibration error after temperature scaling is also higher, with 95\% intervals of the paired difference in ECE of $[0.021,0.051]$ and $[0.027,0.055]$, and the ascending Beta gate at a target of 95\% decides 260 to 379 items per 1,000, against 998 to 999 for the original head, while breaking its promise in 8\% to 11\% of the partitions, against none. The word scores do not explain even the additive model itself. On the 770 Portuguese report items with at least five units, their mean Spearman correlation with the exact Shapley value of $y_{jk^*}$ in the same model is 0.024, below the 99th percentile of a within-item permutation null (0.035), and a median 69\% of the summed evidence falls outside the word tokens, on special tokens and the final separator. On the original head, occlusion reaches 0.567 against its own exact Shapley value and SLALOM, the additive surrogate of \citet{leemann2024}, 0.840. By the criterion, the head is additive in positions, not a self-explanation by word. The preregistered fallback adds to the logit a residual $r_j$ from the original head, penalised by $\lambda\,\mathrm{mean}_j r_j^2$, and reports the additive share, the median over items of the variance across options carried by the additive sum relative to the sum and the residual together. With seed 0 the penalty works as a dial. At $\lambda=0.01$ the residual carries the decision (share 0.0003, accuracy 0.970 in Portuguese and 0.967 in English), at $\lambda=0.1$ the share is 0.62 with an accuracy of 0.967 and 0.961, and at $\lambda=1$ it is 0.996. Because $\lambda=1$ was singled out after seeing seed 0, its two further seeds are exploratory. Over the three seeds it keeps an additive share of 0.996 to 0.998 at a paired loss of 3.7 points in Portuguese ($[-4.7,-2.7]$) and 3.8 in English ($[-4.9,-2.8]$), with an ECE of 0.06 to 0.09 against 0.01 to 0.02 for the original head. Its word scores now track its own exact Shapley value above the permutation null (mean Spearman 0.226 against 0.035), but far below occlusion and SLALOM on the original head, and 58\% of the evidence still falls outside the words. By the criterion it is a measured cost, not a competitive head, and it is additive in positions rather than by word. For this problem neither head is a useful route to exact explanations. The pure additive head costs about 30 points of accuracy and does not explain itself, the penalised residual costs about 4 points and worse calibration and still does not explain itself by word, and for short inputs the exact game of the original head gives exact attributions without any loss.

\section{Repairing the re-qualification of the loop}\label{app:repair}

The second ablation of the drift loop, summarised in \Cref{sec:drift}, is reported here in full.

\paragraph{Repairing the re-qualification.} A second ablation, designed after the first and therefore exploratory, separates three causes on the same streams, one change at a time. Re-qualifying only on the held-out labels that decided the promotion, which the new model was not trained on, repairs D1: the share of post-change windows below the promise falls by 9.2 points ($[-15.8,-2.6]$) and the cost by 0.016. It does not help in D3. The fixed-sequence scan in place of the ascending one repairs D3, lowering that share by 21.7 points ($[-29.8,-13.5]$) at the price of 124 more labels per 1,000 items, and weighting each audited item by the inverse of the audit rate, so that the calibration sample represents the stream, lowers it by 17.9 points ($[-25.2,-10.6]$) with a lower cost. Neither changes D1 detectably. The causes therefore differ by drift. Under new scenarios the threshold fails because it is chosen on items the model has just been trained on, and under a label change it fails because the ascending scan and an unrepresentative sample pick a threshold the stream does not support, the failure that \Cref{prop:scan} describes. The three changes together keep the gain in D1 ($-9.5$ points) and lose it in D3 ($-0.4$, $[-6.1,5.3]$). A plausible reason, not tested here, is that the held-out sample is small, the weights shrink its effective size further, and no threshold then qualifies, so the loop keeps the old one. The pair without weights, fixed-sequence scan on held-out labels, was not run.

\paragraph{Real streams.} Ten year-to-year transitions of News Category headlines, five orders each, were run with all nine arms. The manipulation check keeps five transitions, those starting in 2013, 2015, 2016, 2018 and 2019, whose frozen model loses 2.6 to 36.6 points of accuracy across the year boundary. In the other five the frozen model loses less than 2 points or improves, and they are reported but left out of the summary, as preregistered. Over the five kept transitions, treated as clusters, the loop is cheaper than the periodic arm by 0.27 per item on average, with a 95\% interval of $[-0.58,0.04]$ that contains zero, so the preregistered summary detects no difference. Transition by transition the direction is the same in all ten, kept or not, and the loop has the lowest mean cost of the nine arms in nine of them. The exception is the transition with the largest drop, from 2013 to 2014, where Learn then Test per window is cheapest (0.86 against 0.95 for the loop and 1.64 for the periodic arm). On the share of windows below the promise the transitions disagree (interval $[-0.90,0.35]$). As supporting evidence, the real streams agree with the synthetic reading. Where the change is mild, as in most year transitions of news headlines, the loop is the cheapest arm, and where it is large and abrupt, re-qualifying the threshold wins.

\section{Gate at all targets, split-half evaluation}\label{app:tables}

The split-half evaluation draws 1,000 random partitions of the 2,000 items into a calibration half and a report half (seed 20261021), fits the temperature and the threshold on the first and measures accuracy among accepted items on the second. Its rates include report-side noise and serve to attribute effects in \Cref{sec:ablation}. The promise itself is judged against the plug-in population in \Cref{tab:gate}.

\begin{table}[!htbp]
\centering
\caption{Broken promises per split on the report half and decisions per 1,000, for the point rule, the ascending Beta rule ($q=0.05$) and Learn then Test ($\delta=0.05$), at targets of 95\%, 99\% and 99.5\%, 1,000 splits. ``none'': no qualifying threshold in any split, which keeps the promise by abstaining. In bold, per system and target, the rule with the fewest broken promises and, among ties, the most decisions.}
\label{tab:gate-all}
\scriptsize
\setlength{\tabcolsep}{2.5pt}
\begin{tabular}{llccccccccc}
\toprule
 & & \multicolumn{3}{c}{95\%} & \multicolumn{3}{c}{99\%} & \multicolumn{3}{c}{99.5\%} \\
\cmidrule(lr){3-5}\cmidrule(lr){6-8}\cmidrule(lr){9-11}
Lang. & System & point & Beta & LTT & point & Beta & LTT & point & Beta & LTT \\
\midrule
pt & Jev 1.13 & 0.50 / 995 & \textbf{0.10 / 976} & \textbf{0.10 / 976} & 0.50 / 859 & \textbf{0.03 / 314} & 0.03 / 251 & 0.46 / 536 & \textbf{0.02 / 15} & \textbf{0.02 / 15} \\
pt & Laya multilingual & 0.55 / 443 & 0.10 / 311 & \textbf{0.07 / 212} & 0.64 / 142 & \textbf{none} & \textbf{none} & 0.36 / 58 & \textbf{none} & \textbf{none} \\
pt & Laya English & 0.52 / 327 & 0.09 / 182 & \textbf{0.02 / 89} & 0.60 / 105 & \textbf{none} & \textbf{none} & 0.47 / 82 & \textbf{none} & \textbf{none} \\
pt & Laya typed-decisions & 0.52 / 397 & 0.09 / 279 & \textbf{0.05 / 226} & 0.47 / 160 & \textbf{none} & \textbf{none} & 0.49 / 139 & \textbf{none} & \textbf{none} \\
pt & Laya multilingual, FT & \textbf{0.00 / 999} & \textbf{0.00 / 999} & \textbf{0.00 / 999} & 0.49 / 955 & 0.07 / 883 & \textbf{0.06 / 846} & 0.48 / 898 & \textbf{0.04 / 709} & \textbf{0.04 / 709} \\
pt & Laya typed-dec., FT & \textbf{0.00 / 999} & 0.00 / 997 & 0.00 / 996 & 0.50 / 913 & \textbf{0.07 / 834} & 0.07 / 820 & 0.47 / 854 & \textbf{0.05 / 688} & \textbf{0.05 / 688} \\
\midrule
en & Jev 1.13 & \textbf{0.00 / 999} & 0.00 / 996 & 0.00 / 996 & 0.47 / 899 & \textbf{0.07 / 401} & \textbf{0.07 / 401} & 0.45 / 547 & \textbf{0.01 / 7} & \textbf{0.01 / 7} \\
en & Laya English & 0.50 / 835 & 0.10 / 748 & \textbf{0.05 / 402} & 0.47 / 511 & 0.08 / 34 & \textbf{0.05 / 19} & 0.56 / 300 & \textbf{none} & \textbf{none} \\
en & Laya typed-decisions & 0.48 / 868 & \textbf{0.11 / 831} & 0.11 / 829 & 0.48 / 632 & 0.05 / 277 & \textbf{0.03 / 129} & 0.52 / 549 & \textbf{0.00 / 2} & \textbf{0.00 / 2} \\
en & Laya multilingual & 0.51 / 639 & 0.10 / 584 & \textbf{0.09 / 472} & 0.46 / 348 & 0.02 / 7 & \textbf{0.01 / 5} & 0.28 / 101 & \textbf{none} & \textbf{none} \\
\bottomrule

\end{tabular}
\end{table}

\section{Reference table by language}\label{app:languages}

\Cref{tab:languages} gives, for each locale of MASSIVE 1.1, the accuracy of the three deciders on the 1,000 parallel report items and the violation rate of the promise of 95\% by the ascending Beta rule, calibrated on the en-US calibration items or on the calibration items of the locale (Laya only, since Jev answered the report items alone in the new locales). The pt-PT and en-US answers of Jev are those of the noise run of \Cref{sec:noise}. Intervals, non-inferiority to en-US, ECE, decisions and the other rules are in the released file \texttt{analise\_multilingue.json}, and the same table as CSV in \texttt{tabela\_linguas.csv}.

\begin{table}[!htbp]
\centering
\caption{Languages without fine-tuning (and Laya fine-tuned in Portuguese), report items. Script: the Unicode script of most letters in the locale's utterances. Accuracy and violation rate of the Beta rule at 95\% (2,000 draws) calibrated in en-US or in the locale. Laya-ft is the multilingual checkpoint fine-tuned on pt-PT. In bold, per row, the highest accuracy and the lowest violation rate of each group, where they differ.}
\label{tab:languages}
\scriptsize
\setlength{\tabcolsep}{3pt}
\begin{tabular}{llccccccccc}
\toprule
 & & \multicolumn{3}{c}{Accuracy} & \multicolumn{3}{c}{Violation, calibrated in en-US} & \multicolumn{2}{c}{Violation, in the locale} \\
\cmidrule(lr){3-5}\cmidrule(lr){6-8}\cmidrule(lr){9-10}
Locale & Script & Jev & Laya & Laya-ft & Jev & Laya & Laya-ft & Laya & Laya-ft \\
\midrule
af-ZA & Latin & \textbf{0.934} & 0.595 & 0.832 & \textbf{0.86} & 1.00 & 1.00 & 0.51 & \textbf{0.40} \\
am-ET & Ethiopic & \textbf{0.905} & 0.344 & 0.571 & 1.00 & 1.00 & 1.00 & \textbf{0.00} & 0.15 \\
ar-SA & Arabic & \textbf{0.917} & 0.637 & 0.878 & \textbf{0.98} & 1.00 & 1.00 & \textbf{0.01} & 0.08 \\
az-AZ & Latin & \textbf{0.946} & 0.624 & 0.844 & \textbf{0.73} & 1.00 & 1.00 & \textbf{0.16} & 0.25 \\
bn-BD & Bengali & \textbf{0.938} & 0.555 & 0.788 & \textbf{0.83} & 1.00 & 1.00 & \textbf{0.03} & 0.19 \\
ca-ES & Latin & \textbf{0.941} & 0.601 & 0.896 & \textbf{0.77} & 1.00 & 1.00 & \textbf{0.00} & 0.08 \\
cy-GB & Latin & \textbf{0.889} & 0.420 & 0.647 & 1.00 & 1.00 & 1.00 & \textbf{0.00} & 0.84 \\
da-DK & Latin & \textbf{0.946} & 0.716 & 0.932 & \textbf{0.73} & 1.00 & 1.00 & \textbf{0.00} & 0.06 \\
de-DE & Latin & \textbf{0.942} & 0.726 & 0.933 & 0.86 & \textbf{0.82} & 1.00 & \textbf{0.02} & 0.08 \\
el-GR & Greek & \textbf{0.944} & 0.609 & 0.877 & \textbf{0.77} & 1.00 & 1.00 & \textbf{0.32} & 0.86 \\
en-US & Latin & 0.961 & 0.816 & \textbf{0.973} & \textbf{0.00} & 0.01 & \textbf{0.00} & 0.01 & \textbf{0.00} \\
es-ES & Latin & 0.949 & 0.749 & \textbf{0.952} & 0.31 & 0.44 & \textbf{0.00} & 0.00 & 0.00 \\
fa-IR & Arabic & \textbf{0.945} & 0.687 & 0.905 & \textbf{0.77} & 1.00 & 1.00 & \textbf{0.00} & 0.44 \\
fi-FI & Latin & \textbf{0.935} & 0.636 & 0.785 & \textbf{0.86} & 1.00 & 1.00 & 0.56 & \textbf{0.06} \\
fr-FR & Latin & \textbf{0.958} & 0.741 & 0.953 & \textbf{0.00} & 0.84 & \textbf{0.00} & 0.00 & 0.00 \\
he-IL & Hebrew & \textbf{0.944} & 0.626 & 0.847 & \textbf{0.67} & 1.00 & 1.00 & 0.68 & \textbf{0.24} \\
hi-IN & Devanagari & \textbf{0.953} & 0.707 & 0.903 & \textbf{0.00} & 1.00 & 1.00 & 0.12 & \textbf{0.00} \\
hu-HU & Latin & \textbf{0.943} & 0.618 & 0.855 & \textbf{0.77} & 1.00 & 1.00 & 0.08 & \textbf{0.07} \\
hy-AM & Armenian & \textbf{0.933} & 0.440 & 0.718 & \textbf{0.87} & 1.00 & 1.00 & \textbf{0.00} & 0.01 \\
id-ID & Latin & \textbf{0.944} & 0.735 & 0.936 & \textbf{0.77} & 1.00 & 1.00 & 0.24 & \textbf{0.01} \\
is-IS & Latin & \textbf{0.921} & 0.503 & 0.729 & \textbf{0.97} & 1.00 & 1.00 & 0.03 & \textbf{0.00} \\
it-IT & Latin & 0.948 & 0.735 & \textbf{0.952} & 0.77 & 0.84 & \textbf{0.00} & 0.02 & \textbf{0.00} \\
ja-JP & Hiragana & 0.951 & 0.802 & \textbf{0.960} & 0.00 & 0.00 & 0.00 & 0.00 & 0.00 \\
jv-ID & Latin & \textbf{0.867} & 0.523 & 0.756 & 1.00 & 1.00 & 1.00 & 0.28 & \textbf{0.14} \\
ka-GE & Georgian & \textbf{0.916} & 0.384 & 0.635 & \textbf{0.98} & 1.00 & 1.00 & \textbf{0.00} & 0.06 \\
km-KH & Khmer & \textbf{0.881} & 0.378 & 0.610 & 1.00 & 1.00 & 1.00 & \textbf{0.00} & 0.40 \\
kn-IN & Kannada & \textbf{0.938} & 0.417 & 0.693 & \textbf{0.89} & 1.00 & 1.00 & \textbf{0.00} & 0.04 \\
ko-KR & Hangul & \textbf{0.954} & 0.716 & 0.938 & \textbf{0.00} & 0.84 & 1.00 & 0.00 & 0.00 \\
lv-LV & Latin & \textbf{0.922} & 0.549 & 0.798 & \textbf{0.93} & 1.00 & 1.00 & \textbf{0.30} & 0.62 \\
ml-IN & Malayalam & \textbf{0.936} & 0.421 & 0.726 & \textbf{0.86} & 1.00 & 1.00 & \textbf{0.00} & 0.09 \\
mn-MN & Cyrillic & \textbf{0.924} & 0.446 & 0.628 & \textbf{0.89} & 1.00 & 1.00 & \textbf{0.01} & 0.68 \\
ms-MY & Latin & \textbf{0.927} & 0.694 & 0.901 & 0.86 & \textbf{0.84} & 1.00 & \textbf{0.00} & 0.02 \\
my-MM & Myanmar & \textbf{0.929} & 0.376 & 0.625 & \textbf{0.86} & 1.00 & 1.00 & \textbf{0.00} & 0.48 \\
nb-NO & Latin & \textbf{0.947} & 0.701 & 0.912 & \textbf{0.67} & 0.84 & 1.00 & \textbf{0.01} & 0.10 \\
nl-NL & Latin & \textbf{0.946} & 0.714 & 0.936 & \textbf{0.83} & 0.98 & 0.99 & 0.66 & \textbf{0.13} \\
pl-PL & Latin & \textbf{0.949} & 0.720 & 0.923 & \textbf{0.77} & 0.97 & 1.00 & 0.41 & \textbf{0.25} \\
pt-PT & Latin & 0.950 & 0.732 & \textbf{0.973} & \textbf{0.00} & 0.88 & \textbf{0.00} & 0.19 & \textbf{0.00} \\
ro-RO & Latin & \textbf{0.940} & 0.713 & 0.922 & \textbf{0.83} & 1.00 & 1.00 & 0.05 & \textbf{0.02} \\
ru-RU & Cyrillic & \textbf{0.954} & 0.710 & 0.937 & \textbf{0.00} & 1.00 & 1.00 & 0.26 & \textbf{0.18} \\
sl-SL & Latin & \textbf{0.931} & 0.594 & 0.861 & \textbf{0.87} & 1.00 & 1.00 & 0.51 & \textbf{0.11} \\
sq-AL & Latin & \textbf{0.915} & 0.601 & 0.818 & \textbf{0.97} & 1.00 & 1.00 & 0.49 & \textbf{0.16} \\
sv-SE & Latin & \textbf{0.952} & 0.695 & 0.923 & \textbf{0.00} & 0.84 & 1.00 & \textbf{0.01} & 0.13 \\
sw-KE & Latin & \textbf{0.873} & 0.450 & 0.642 & 1.00 & 1.00 & 1.00 & \textbf{0.00} & 0.28 \\
ta-IN & Tamil & \textbf{0.942} & 0.504 & 0.795 & \textbf{0.83} & 1.00 & 1.00 & \textbf{0.00} & 0.01 \\
te-IN & Telugu & \textbf{0.940} & 0.513 & 0.706 & \textbf{0.86} & 1.00 & 1.00 & \textbf{0.00} & 0.34 \\
th-TH & Thai & \textbf{0.936} & 0.681 & 0.919 & \textbf{0.83} & 1.00 & 1.00 & 0.91 & \textbf{0.00} \\
tl-PH & Latin & \textbf{0.939} & 0.652 & 0.881 & \textbf{0.83} & 1.00 & 1.00 & 0.51 & \textbf{0.00} \\
tr-TR & Latin & \textbf{0.945} & 0.698 & 0.907 & \textbf{0.83} & 0.99 & 1.00 & \textbf{0.01} & 0.32 \\
ur-PK & Arabic & \textbf{0.938} & 0.635 & 0.863 & \textbf{0.86} & 0.97 & 1.00 & 0.01 & 0.01 \\
vi-VN & Latin & \textbf{0.957} & 0.690 & 0.880 & \textbf{0.00} & 1.00 & 1.00 & 0.51 & \textbf{0.03} \\
zh-CN & Han & 0.946 & 0.778 & \textbf{0.956} & 0.83 & 0.76 & \textbf{0.00} & 0.06 & \textbf{0.00} \\
zh-TW & Han & \textbf{0.927} & 0.715 & 0.926 & 0.93 & \textbf{0.79} & 1.00 & \textbf{0.18} & 0.25 \\
\bottomrule

\end{tabular}
\end{table}

\section{Report cards of the battery}\label{app:battery}

\Cref{tab:battery-cards} gives one block per domain of \Cref{sec:battery}, with the same layout for all, MASSIVE pt-PT first as the reference measured with the same code. Each promise cell reads violation rate over decisions per 1,000, against the plug-in population with 2,000 draws, so a cell such as 0.00/0 is a promise kept by deciding nothing. The released and fine-tuned Laya are calibrated by temperature on each draw, and Jev is used as returned. The label curves are in \Cref{fig:battery}, and every value with its interval is in the released file \texttt{analise\_bateria.json}.

\begin{table}[!htbp]
\centering
\caption{Report cards of the battery, report items. Accuracy, ECE (Jev as returned, Laya after temperature on the calibration half), violation rate over decisions per 1,000 of the point rule, the ascending Beta rule and Learn then Test at 0.90 and 0.95, the Beta promise at 0.95 calibrated on clean text and checked under typos and under the mixed profile at 10\%, and accuracy with an empty state. In bold, within each block, the highest accuracy, the lowest ECE and, in each promise column, the decider that keeps the promise with the most decisions.}
\label{tab:battery-cards}
\scriptsize
\setlength{\tabcolsep}{2.2pt}
\begin{tabular}{@{}lcccccccccccc@{}}
\toprule
 & & & \multicolumn{3}{c}{Promise at 0.90} & \multicolumn{3}{c}{Promise at 0.95} & \multicolumn{2}{c}{Beta at 0.95 under} & Empty \\
\cmidrule(lr){4-6}\cmidrule(lr){7-9}\cmidrule(lr){10-11}
Decider & Acc. & ECE & point & Beta & LTT & point & Beta & LTT & typos & mixed & state \\
\midrule
\multicolumn{12}{@{}l}{\textit{MASSIVE pt-PT (reference)}, 5 options, fine-tunes on 8,000 labels} \\
Jev 1.13 & 0.947 & 0.020 & \textbf{0.00/999} & \textbf{0.00/999} & \textbf{0.00/999} & 0.41/993 & 0.03/975 & 0.03/974 & 0.48/968 & 0.16/967 & 0.281 \\
Laya & 0.732 & 0.043 & 0.48/635 & 0.05/561 & 0.04/525 & 0.52/438 & 0.05/321 & 0.03/229 & 0.54/319 & 0.98/303 & 0.175 \\
Laya-ft, seed 0 & 0.959 & \textbf{0.014} & \textbf{0.00/999} & \textbf{0.00/999} & \textbf{0.00/999} & \textbf{0.00/999} & 0.00/994 & 0.00/993 & 0.95/983 & 0.97/983 & 0.395 \\
Laya-ft, seed 1 & \textbf{0.963} & 0.019 & \textbf{0.00/999} & \textbf{0.00/999} & \textbf{0.00/999} & \textbf{0.00/999} & \textbf{0.00/997} & \textbf{0.00/997} & 1.00/995 & 0.99/992 & 0.408 \\
Laya-ft, seed 2 & 0.958 & 0.015 & \textbf{0.00/999} & \textbf{0.00/999} & \textbf{0.00/999} & \textbf{0.00/999} & 0.00/996 & 0.00/996 & 0.99/996 & 1.00/996 & 0.405 \\
\midrule
\multicolumn{12}{@{}l}{\textit{News Category}, 5 options, fine-tunes on 8,000 labels} \\
Jev 1.13 & 0.817 & 0.076 & 0.47/834 & 0.04/783 & \textbf{0.04/782} & 0.31/650 & \textbf{0.02/536} & \textbf{0.01/529} & 0.20/564 & 0.62/559 & 0.231 \\
Laya & 0.585 & 0.048 & 0.55/269 & 0.05/149 & 0.01/25 & 0.55/121 & 0.03/14 & 0.00/6 & \textbf{0.04/3} & \textbf{0.04/4} & 0.174 \\
Laya-ft, seed 0 & \textbf{0.833} & 0.038 & 0.50/830 & 0.04/778 & 0.04/775 & 0.51/665 & 0.05/558 & 0.03/407 & 0.92/595 & 0.99/541 & 0.345 \\
Laya-ft, seed 1 & \textbf{0.833} & \textbf{0.029} & 0.50/829 & \textbf{0.04/785} & 0.04/771 & 0.52/648 & 0.05/510 & 0.03/331 & 0.73/517 & 0.60/430 & 0.372 \\
Laya-ft, seed 2 & 0.827 & 0.039 & 0.51/824 & 0.05/766 & 0.04/747 & 0.47/635 & 0.04/517 & 0.03/327 & 0.79/538 & 0.98/476 & 0.318 \\
\midrule
\multicolumn{12}{@{}l}{\textit{Financial tweets}, 3 options, fine-tunes on 8,000 labels} \\
Jev 1.13 & 0.728 & 0.171 & 0.45/379 & 0.03/204 & 0.02/202 & 0.07/25 & 0.00/0 & 0.00/0 & 0.00/0 & 0.00/0 & 0.645 \\
Laya & 0.642 & 0.047 & 0.57/32 & 0.03/1 & 0.01/1 & 0.30/9 & 0.00/0 & 0.00/0 & 0.00/0 & 0.00/0 & 0.645 \\
Laya-ft, seed 0 & 0.925 & 0.039 & \textbf{0.00/999} & 0.00/992 & 0.00/980 & 0.52/893 & 0.04/818 & 0.03/532 & 0.21/691 & 1.00/623 & 0.645 \\
Laya-ft, seed 1 & \textbf{0.929} & \textbf{0.020} & \textbf{0.00/999} & \textbf{0.00/997} & \textbf{0.00/995} & 0.48/921 & 0.04/842 & \textbf{0.03/582} & 0.87/744 & 1.00/652 & 0.645 \\
Laya-ft, seed 2 & 0.924 & 0.034 & \textbf{0.00/999} & 0.00/989 & 0.00/858 & 0.48/904 & \textbf{0.04/860} & 0.03/524 & 0.78/758 & 1.00/745 & 0.645 \\
\midrule
\multicolumn{12}{@{}l}{\textit{LIAR2}, 6 options, fine-tunes on 8,000 labels} \\
Jev 1.13 & 0.319 & 0.162 & 0.00/0 & 0.00/0 & 0.00/0 & 0.00/0 & 0.00/0 & 0.00/0 & 0.00/0 & 0.00/0 & 0.285 \\
Laya & 0.282 & 0.030 & 0.00/0 & 0.00/0 & 0.00/0 & 0.00/0 & 0.00/0 & 0.00/0 & 0.00/0 & 0.00/0 & 0.156 \\
Laya-ft, seed 0 & 0.320 & 0.034 & 0.01/0 & 0.00/0 & 0.00/0 & 0.00/0 & 0.00/0 & 0.00/0 & 0.00/0 & 0.00/0 & 0.285 \\
Laya-ft, seed 1 & 0.309 & 0.043 & 0.01/0 & 0.00/0 & 0.00/0 & 0.00/0 & 0.00/0 & 0.00/0 & 0.00/0 & 0.00/0 & 0.102 \\
Laya-ft, seed 2 & \textbf{0.342} & \textbf{0.017} & 0.00/0 & 0.00/0 & 0.00/0 & 0.00/0 & 0.00/0 & 0.00/0 & 0.00/0 & 0.00/0 & 0.285 \\
\midrule
\multicolumn{12}{@{}l}{\textit{Fake.Br (pt)}, 2 options, fine-tunes on 5,200 labels} \\
Jev 1.13 & 0.716 & 0.087 & 0.40/406 & 0.03/289 & 0.01/96 & 0.49/179 & 0.03/18 & 0.00/6 & 0.42/36 & 0.42/27 & 0.507 \\
Laya & 0.567 & 0.027 & 0.05/1 & 0.00/0 & 0.00/0 & 0.01/0 & 0.00/0 & 0.00/0 & 0.00/0 & 0.00/0 & 0.507 \\
Laya-ft, seed 0 & \textbf{0.983} & 0.007 & \textbf{0.00/999} & \textbf{0.00/999} & \textbf{0.00/999} & \textbf{0.00/999} & \textbf{0.00/999} & \textbf{0.00/999} & 0.00/996 & 1.00/966 & 0.493 \\
Laya-ft, seed 1 & 0.981 & 0.008 & \textbf{0.00/999} & \textbf{0.00/999} & \textbf{0.00/999} & \textbf{0.00/999} & \textbf{0.00/999} & \textbf{0.00/999} & \textbf{0.00/998} & 1.00/984 & 0.493 \\
Laya-ft, seed 2 & 0.980 & \textbf{0.006} & \textbf{0.00/999} & \textbf{0.00/999} & \textbf{0.00/999} & \textbf{0.00/999} & \textbf{0.00/999} & \textbf{0.00/999} & \textbf{0.00/998} & 1.00/988 & 0.493 \\
\bottomrule
 
\end{tabular}
\end{table}

\end{document}